\documentclass{article} 
\usepackage{iclr2027_conference,times}

\usepackage[utf8]{inputenc} 
\usepackage[T1]{fontenc}    
\usepackage{hyperref}       
\usepackage{url}            
\usepackage{booktabs}       
\usepackage{amsfonts}       
\usepackage{microtype}      
\usepackage{multirow}
\usepackage{tabularx}
\usepackage{algorithm}
\usepackage{algpseudocode}
\usepackage{float}
\usepackage{amsmath}
\usepackage{amssymb}
\usepackage{amsthm}
\usepackage{graphicx}
\usepackage[most]{tcolorbox}
\usepackage{bbm}
\usepackage{longtable}
\usepackage{pdflscape}

\newtheorem{theorem}{Theorem}

\iclrfinalcopy

\newtcolorbox{takeaway}{
  enhanced,
  colback=blue!3,
  colframe=blue!60!black,
  boxrule=0pt,
  leftrule=2.5pt,
  arc=0pt,
  left=7pt,
  right=7pt,
  top=5pt,
  bottom=5pt,
  before skip=8pt,
  after skip=8pt
}

\title{Unmask the State: When Does State Adaptation Matter for Masked Diffusion Language Models}

\author{
  \textbf{Injin Kong} \quad
  \textbf{Sunghwan Choi} \quad
  \textbf{Yohan Jo}\thanks{Corresponding author.} \\
  Graduate School of Data Science, Seoul National University \\
  \texttt{mtkong77@snu.ac.kr, b1lly13@snu.ac.kr, yohan.jo@snu.ac.kr}\\
}

\begin{document}

\maketitle

\begin{abstract}
Masked diffusion language models (MDMs) admit flexible generation orders, making the unmasking strategy an inference decision. Existing methods vary in how they prioritize positions, control parallelism, restrict selection regions, revise predictions, or plan future denoising, yet it remains unclear when these choices should change during generation. We study this question through strategy reversals, where an alternative action becomes preferable to a fixed choice. We organize MDM inference into five axes---score, cardinality, region, commitment, and planning---and define adaptation opportunity as the one-step utility advantage of the best candidate action over a validation-selected fixed action. This view shows that adaptation value depends on both the frequency and magnitude of such reversals. Across three MDMs and ten tasks, adaptation opportunities are highly heterogeneous, with some regimes exhibiting concentrated and predictable one-step gains. This motivates selective adaptation: lightweight detectors calibrated on validation prompts identify high-opportunity states, capturing, for example, 56.9\% of the candidate-set oracle opportunity by adapting only the top 10\% of states on LLaDA-8B constrained JSON filling. Our transition-level results suggest that state adaptation is most useful when applied selectively rather than uniformly.
\end{abstract}

\section{Introduction}
\label{intro}

Masked diffusion language models (MDMs) have emerged as an alternative to autoregressive generation \citep{austin2021structured, lou2023discrete, nie2025large, sahoo2024simple, shi2024simplified}, iteratively unmasking tokens from partially masked sequences without a predefined generation order. This flexibility leaves open which predictions should be revealed at each step, making the unmasking strategy an inference decision that can affect generation quality \citep{kim2025train,peng2025pathplanning,hong2026improving,zhong-etal-2026-parallelism}.

A growing body of work explores unmasking rules: which predictions to prioritize \citep{ghazvininejad2019maskpredict,kim2025klass,benhamu2025accelerated,zhou2026dos,ringel2026demask}, how many and where to reveal \citep{chang2022maskgit,wu2026fastdllm,luxembourg2026plan,luo2026dsb,zhang2026swordsman}, whether revealed predictions remain fixed \citep{wang2025remasking,hong2026widein,zhai2026core,shu2026dcd,wang2026vsb}, and whether the decision accounts for future denoising \citep{peng2025pathplanning,hong2026improving,yang2026improving,lee2025lookum,huang2026medal,shen2026ordertoken,liu2025think}. This literature reveals a broad strategy space, but leaves open when the preferred choice should change with the denoising state.

We refer to state-dependent changes as state adaptation. We first ask whether adaptation opportunities exist and where they concentrate across states. We then separate practical exploitability into two questions: whether observable trajectory signals can propose useful alternative actions, and whether high-opportunity states can be identified before deviating from the fixed action.

We organize the strategy space into five axes---score, cardinality, region, commitment, and planning. We study the first four axes while holding planning fixed. For each axis, we measure how much the best state-dependent action can improve over a fixed action selected on validation. We then analyze how often the fixed action becomes suboptimal across states, and by how much. This allows us to distinguish uniformly small gains from gains concentrated in a subset of states.

The results support this selective view. Across three models and ten tasks, adaptation opportunities vary substantially across tasks and axes. Detector performance is modest on average, but some regimes exhibit concentrated and predictable gains. For example, on LLaDA-8B constrained JSON filling, a region detector achieves $0.854$ AUROC and captures $84.3\%$ of one-step oracle opportunity in the top $20\%$ of states. We use opportunity detection to gate deviations from the fixed action. End-to-end experiments demonstrate improvements over coverage-matched random adaptation in settings where intervention timing matters. Our contributions are summarized below.

\begin{takeaway}

\textbf{Unified strategy space.}
We organize MDM unmasking strategies into five axes: \\ score, cardinality, region, commitment, and planning.
(Section~\ref{taxonomy})

\textbf{Adaptation opportunity.}
We formalize when state-dependent actions can improve over a strong fixed action.
(Section~\ref{theorem})

\textbf{Selective adaptation.}
We separate deciding when to adapt from selecting which alternative action to take.
(Section~\ref{method})

\textbf{Empirical characterization.}
Across three models and ten tasks, we identify where adaptation opportunities exist, where they are predictable, and when they can be exploited.
(Section~\ref{experiments})

\end{takeaway}

\section{Related Work}
\label{related_work}

\paragraph{Masked Diffusion Language Models and Unmasking Strategies.}
MDMs~\citep{sahoo2024simple} iteratively recover tokens from partially masked sequences, enabling bidirectional conditioning and flexible generation orders. Models such as SMDM~\citep{nie2025scaling}, LLaDA~\citep{nie2025large}, and Dream~\citep{ye2025dream} demonstrate the scalability of this paradigm. Because MDMs do not prescribe a unique generation order, inference requires decisions about how masked positions are revealed. Existing methods prioritize positions through confidence, stability, entropy, or dependency-aware scores \citep{wu2026fastdllm,kim2025klass,benhamu2025accelerated,zhou2026dos,ringel2026demask}; control parallelism through fixed or adaptive cardinality \citep{wu2026fastdllm,kim2025klass,benhamu2025accelerated,luo2026dsb,zhang2026swordsman}; and adapt or constrain generation regions through blocks, restricted windows, or dilated patterns \citep{arriola2025block,seo2025fast,luxembourg2026plan,luo2026dsb,zhang2026swordsman,shu2026dcd,wang2026vsb}. Others control whether predictions remain committed or revisable \citep{wang2025remasking,hong2026widein,zhai2026core,shu2026dcd,wang2026vsb}, or explicitly search or account for future denoising consequences \citep{peng2025pathplanning,yang2026improving,lee2025lookum,cao2026soar,shen2026ordertoken,huang2026medal,liu2025think}. Together, these approaches span five recurring decisions: score, cardinality, region, commitment, and planning. We ask whether the preferred choices along these axes should vary with the current denoising state.

\paragraph{Toward a Unified View of MDM Inference.}
Recent work has begun to formalize this policy space: Path Planning (P2)~\citep{peng2025pathplanning} casts alternative unmasking orders within a common planning framework, DUEL~\citep{turok2026duel} unifies deterministic rules for selecting which positions to reveal, and OeMDM/LoMDM~\citep{hong2026unifying} learn context-dependent generation orders during training. Other work studies adaptive generation order or scheduling, including adaptive token ordering~\citep{kim2025train}, learned unmasking policies~\citep{hong2026improving,jazbec2026learning}, and ground-truth-guided ordering in Where-to-Unmask~\citep{asano2026wheretounmaskgroundtruthguidedunmaskingorder}. Recent empirical work further shows that parallelism and generation order vary across tasks and reasoning stages~\citep{zhong-etal-2026-parallelism}. We take a complementary view by factorizing unmasking into five interpretable axes and studying when deviations from a fixed choice have measurable utility, where those opportunities concentrate, and whether they can be identified from observable trajectory signals.


\section{Unmasking Strategies: Taxonomy and Parameterization}
\label{taxonomy}

We organize MDM unmasking strategies around five recurring inference decisions: which positions to prioritize (score), where selection is allowed (region), how many positions to reveal (cardinality), whether revealed predictions can be revised (commitment), and whether the current decision accounts for future denoising (planning). Table~\ref{tab:taxonomy} illustrates how representative methods instantiate different choices along these axes. This factorization separates decisions that existing algorithms often couple. We focus on the first four transition-level axes and hold planning fixed, since planning evaluates current decisions through future denoising rather than the instantaneous state transition.

To provide a common representation of different unmasking strategies, we parameterize these four axes over the current masked state. Let $\mathcal{M}_t=\{i:x_{t,i}=[\mathrm{MASK}]\}$ denote the positions still masked at step $t$, with $m_t=|\mathcal{M}_t|$ and $i_1<\cdots<i_{m_t}$.

\paragraph{Score: Which positions should be prioritized?}
The score axis determines which masked positions are preferred for revelation. We represent this preference by a mixture weight $\boldsymbol{\alpha}_t\in\Delta^{J-1}$, where $J$ is the number of candidate scoring rules. Each rule provides a normalized score vector $\mathbf{s}^{(j)}_t\in[0,1]^{m_t}$, and the resulting priority scores are $\mathbf{s}_t=\sum_{j=1}^{J}\alpha_{t,j}\mathbf{s}^{(j)}_t$. The candidate rules may capture confidence, margin, entropy, KL-based stability, or dependency-aware signals, with simplex vertices corresponding to individual rules and interior points to their mixtures.

\paragraph{Region: Where can positions be selected from?}
The region axis determines which masked positions are eligible for selection. We represent it by a binary mask $\mathbf{r}_t\in\{0,1\}^{m_t}$ over the currently masked positions. An entry $r_{t,\ell}=1$ makes position $i_\ell$ selectable, defining the candidate region $\mathcal{R}_t=\{i_\ell\in\mathcal{M}_t:r_{t,\ell}=1\}$. By varying $\mathbf{r}_t$, this representation covers the full masked sequence, blocks, local neighborhoods, dilated patterns, and arbitrary subsets.

\paragraph{Cardinality: How many positions should be revealed?}
The cardinality axis determines how many eligible positions are revealed at the current step. We represent it by a reveal fraction $\kappa_t\in[0,1]$. Given the candidate region $\mathcal{R}_t$, it determines the number of revealed positions as $k_t=\lceil\kappa_t|\mathcal{R}_t|\rceil$, and the reveal set $\mathcal{N}_t\subseteq\mathcal{R}_t$ contains the $k_t$ highest-scoring positions under $\mathbf{s}_t$. Varying $\kappa_t$ covers single-token, fixed-fraction, and adaptive levels of parallelism.

\paragraph{Commitment: Can revealed predictions be revised?}
The commitment axis determines whether a revealed prediction becomes permanent or remains revisable. We represent it by $\mathbf{c}_t=(c_{t,i})_{i\in\mathcal{N}_t}\in\{0,1\}^{k_t}$ over the newly revealed positions $\mathcal{N}_t$. For each $i\in\mathcal{N}_t$, $c_{t,i}=1$ makes the prediction irreversible, whereas $c_{t,i}=0$ keeps it visible as context but allows it to be remasked and predicted again later. The rule for when revisable positions are reconsidered, and which ones, is fixed separately.

\paragraph{Unified decision space.}
Together, the four transition-level axes define
\begin{equation}
a_t=\bigl(\boldsymbol{\alpha}_t,\mathbf{r}_t,\kappa_t,\mathbf{c}_t\bigr)
\in
\Delta^{J-1}\times\{0,1\}^{m_t}\times[0,1]\times\{0,1\}^{k_t}.
\label{eq:decision_space}
\end{equation}

Within this parameterization, existing unmasking algorithms can be viewed as fixing or restricting components of $a_t$. The framework covers all reveal-and-commit policies under fixed denoising and revision dynamics (Theorem~\ref{thm:policy_completeness}). We thus ask when fixed configurations become suboptimal and whether such states can be detected.

\begin{figure}[t]
\centering
\includegraphics[width=0.95\linewidth]{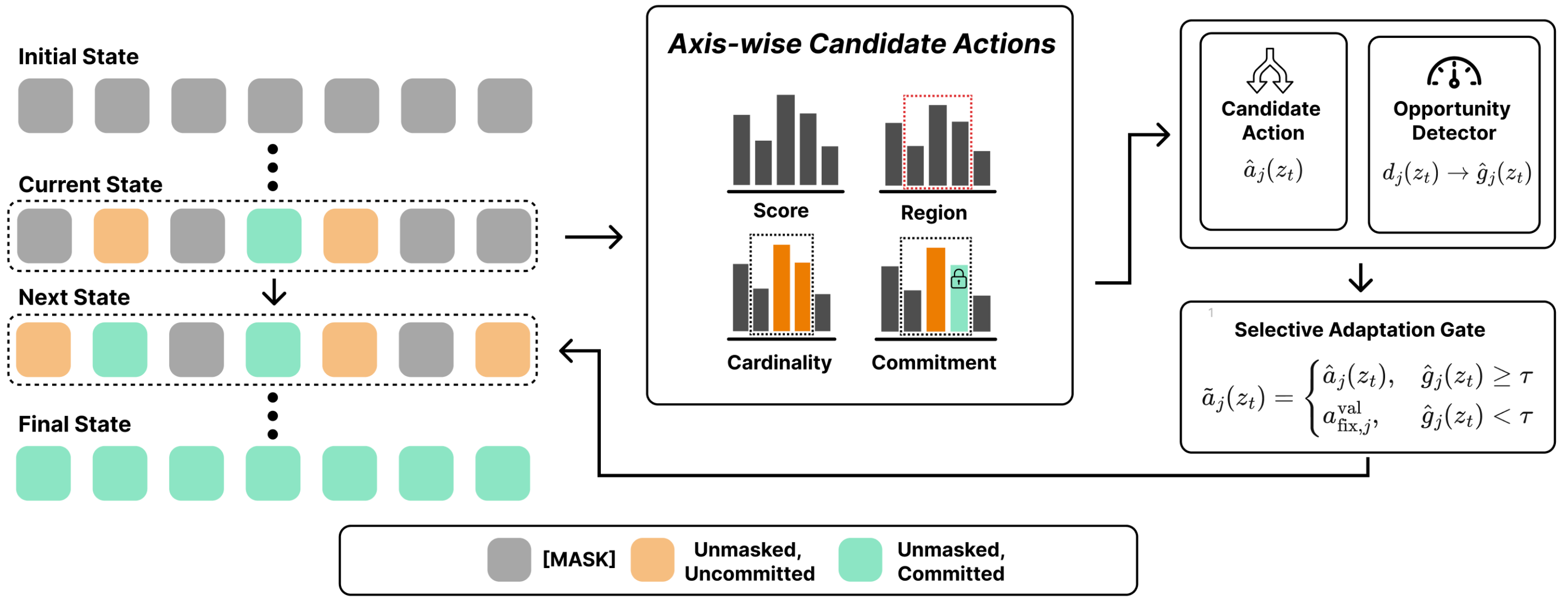}
\caption{\textbf{Selective state adaptation.} An axis-wise diagnostic proposes an alternative action, while a validation-calibrated opportunity detector decides whether to deviate from the fixed action.}
\label{fig:overview}
\end{figure}

\begin{table}[t]
\centering
\caption{\textbf{Representative methods under our five-axis taxonomy.} Entries are abbreviated; full terms and an extended mapping are provided in Appendix~\ref{app:taxonomy}. ``--'' denotes no explicit mechanism beyond the underlying sampler.}
\label{tab:taxonomy}
\scriptsize
\setlength{\tabcolsep}{2.5pt}
\begin{tabular}{lccccc}
\toprule
\textbf{Method} & \textbf{Score} & \textbf{Card.} & \textbf{Region} &
\textbf{Commit.} & \textbf{Planning} \\
\midrule
LLaDA~\citep{nie2025large}
    & Conf. & Scheduled & Full & Irrev. & -- \\
Fast-dLLM~\citep{wu2026fastdllm}
    & Conf. & Thresh.\ adapt. & Block & Irrev. & -- \\
KLASS~\citep{kim2025klass}
    & KL stab.\ + conf. & Multi-token & Full & Irrev. & -- \\
DUS~\citep{luxembourg2026plan}
    & Joint entropy & Grouped & Dilated & Irrev. & -- \\
ReMDM~\citep{wang2025remasking}
    & Conf.\ / remask & Sched.\ dep. & Full & Remask. & -- \\
WINO~\citep{hong2026widein}
    & Draft verif. & Parallel draft & Active & Remask. & Draft--verify \\
Info-Gain~\citep{yang2026improving}
    & Certainty + IG & Cand.-set dep. & Full & Irrev. & Look-ahead \\
\bottomrule
\end{tabular}
\end{table}

\section{Adaptation Opportunity and Strategy Reversals}
\label{theorem}

\subsection{State--Action Utility and Adaptation Opportunity}
\label{state_dependent_optimality}

Let $z$ denote the current denoising state and let $\mathcal A$ denote an admissible candidate set of unmasking actions. For an action $a\in\mathcal A$, we define its one-step state--action utility as
\begin{equation}
Q(z,a)
=
\mathbb E\!\left[
U(X_T)
\mid
z_t=z,\,
a_t=a,\,
a_{t+1:T}=a_{\mathrm{ref}}
\right],
\label{eq:state_action_utility}
\end{equation}
where $U(X_T)$ is terminal task utility and $a_{\mathrm{ref}}$ is a fixed continuation policy. Thus, $Q(z,a)$ isolates the effect of changing the current unmasking decision while holding subsequent decoding fixed.

Under a distribution of denoising states $\mathcal D$, let
$a_{\mathrm{fix}}^\star\in\arg\max_{a\in\mathcal A}
\mathbb E_{z\sim\mathcal D}[Q(z,a)]$
denote the best fixed candidate action. The value available to a state-dependent policy over this fixed action is
\begin{equation}
\Delta_{\mathrm{adapt}}
=
\mathbb E_{z\sim\mathcal D}
\!\left[
\max_{a\in\mathcal A}Q(z,a)
-
Q(z,a_{\mathrm{fix}}^\star)
\right].
\label{eq:adaptation_gap}
\end{equation}
We refer to $\Delta_{\mathrm{adapt}}$ as the adaptation opportunity. It measures candidate-set advantage from state-conditioned actions; it does not assume the maximizing action is identifiable from state information.

The corresponding state-wise opportunity is
\begin{equation}
g(z)
=
\max_{a\in\mathcal A}Q(z,a)
-
Q(z,a_{\mathrm{fix}}^\star)
\ge 0,
\qquad
\Delta_{\mathrm{adapt}}
=
\mathbb E[g(z)].
\label{eq:state_opportunity}
\end{equation}
The mean gap alone does not reveal how this value is distributed across states: the same $\Delta_{\mathrm{adapt}}$ may arise from small improvements on many states or large improvements concentrated on a small subset.

\subsection{Axis-Wise Opportunity}
\label{state_wise_opportunity}

For each transition-level axis $j\in\{\text{score},\text{region},\text{cardinality},\text{commitment}\}$, we define an action set $\mathcal A_j$ and hold other axes at the reference configuration. The opportunity is $g_j(z)=\max_{a\in\mathcal A_j}Q(z,a)-Q(z,a_{\mathrm{fix},j}^\star)$, where $a_{\mathrm{fix},j}^\star$ is the best fixed action for axis $j$. Our experiments evaluate these opportunities separately rather than assuming multiple axes should adapt jointly.

In experiments, we avoid selecting the fixed action on held-out states. Instead, we choose
$\widehat a_{\mathrm{fix},j}^{\,\mathrm{val}}$
using validation prompts only and evaluate
\begin{equation}
\widehat g_j(z)
=
\max_{a\in\mathcal A_j}Q(z,a)
-
Q\!\left(z,\widehat a_{\mathrm{fix},j}^{\,\mathrm{val}}\right)
\label{eq:crossfit_axis_opportunity}
\end{equation}
on disjoint held-out states. This held-out quantity is the principal empirical target in Sections~\ref{method}--\ref{experiments}; its finite-rollout estimator uses the rollout cross-fitting procedure described in Appendix~\ref{app:evaluation_protocol}.

\subsection{Reversal Decomposition}
\label{reversal_decomposition}

State adaptation is useful only on states where the globally preferred fixed action is locally suboptimal. The following result separates adaptation value into how often such reversals occur and how large their utility advantage is.

\begin{theorem}[Reversal Decomposition of Adaptation Value]
\label{thm:strategy_reversal}
Let $a_{\mathrm{fix}}^\star$ be the best fixed action and
$\mathcal R=\{z:g(z)>0\}$. Then
\begin{equation}
\Delta_{\mathrm{adapt}}
=
\Pr(\mathcal R)\,
\mathbb E[g(z)\mid z\in\mathcal R].
\label{eq:frequency_margin_decomposition}
\end{equation}
Thus, adaptation value is the reversal frequency times its average
available utility margin. In particular, if some $b\in\mathcal A$
beats $a_{\mathrm{fix}}^\star$ by at least $m>0$ on a fraction
$p>0$ of states, then $\Delta_{\mathrm{adapt}}\ge pm$.

For $\mathcal A=\{a,b\}$, letting
$D_{a,b}(z)=Q(z,a)-Q(z,b)$ gives
\begin{equation}
\Delta_{a,b}
=
\frac{\mathbb E|D_{a,b}(z)|
-\left|\mathbb E D_{a,b}(z)\right|}{2}.
\label{eq:pairwise_adaptation_gap}
\end{equation}
Equivalently, if $a$ is the better fixed action,
$\Delta_{a,b}
=
\Pr[D_{a,b}(z)<0]\,
\mathbb E[-D_{a,b}(z)\mid D_{a,b}(z)<0]$.
\end{theorem}

Theorem~\ref{thm:strategy_reversal} provides the theoretical basis for our selective view. A small global adaptation gap can arise because reversals are rare, because their margins are small, or both. Conversely, a modest average gap can hide a small set of states with substantial local opportunity. This motivates examining not only mean opportunity, but also its frequency, magnitude, and concentration across states.

The theorem characterizes when state adaptation can matter, but not which observable properties of the current state reveal those opportunities before acting. We treat predictability as a separate empirical question. Ranking reliability, positional drift, context transfer, and revision pressure provide four axis-specific hypotheses, summarized in Section~\ref{state_diagnostics}. Their detailed theoretical motivations are deferred to Appendix~\ref{app:axis_mechanisms}; they explain possible mechanisms of strategy reversal rather than guarantee recovery of the state-wise optimal action.

\section{Detecting and Exploiting Adaptation Opportunities}
\label{method}

Theorem~\ref{thm:strategy_reversal} characterizes how adaptation value arises from state-dependent utility reversals, but it does not identify those states from observable trajectory information. A practical selective policy must therefore solve two distinct problems. First, action selection asks which alternative action to take when deviating from the fixed action. Second, opportunity detection asks whether the current state contains enough predictable opportunity to justify that deviation. We evaluate these components separately: an axis-wise diagnostic proposes a candidate action, while a validation-calibrated detector decides whether to use it instead of the fixed action.

\subsection{Axis-Wise Diagnostics and Action Proposals}
\label{action_proposals}
\label{state_diagnostics}

For each transition-level axis $j$, a training-free diagnostic $\mathcal D_j$ maps observable trajectory statistics to a candidate action $\widehat a_j(z)$, while all remaining axes stay at the reference configuration. We evaluate each diagnostic independently for every task--model--axis setting rather than assuming that a single mechanism should jointly adapt all four decisions.

Each diagnostic is motivated by an axis-specific mechanism of strategy reversal analyzed in Appendix~\ref{app:axis_mechanisms}. For score, we estimate ranking reliability and redundancy among scoring rules, asking which rule provides the most trustworthy ordering. For region, we measure positional drift in the priority landscape, capturing whether high-priority locations remain spatially stable as denoising proceeds. For cardinality, we use a directed context-transfer proxy that balances the benefit of revealing a token now against context it could gain by waiting. For commitment, we compare the persistent-context value of a revealed prediction with its revision pressure under later posterior changes.

These quantities serve two roles. They define the axis-wise candidate actions evaluated in our action-selection experiments and provide low-cost observable signals that may indicate when the relative utility of competing actions changes. They are not assumed to recover the state-wise optimum universally; whether they are predictive is evaluated separately on held-out states. Exact diagnostic definitions and optimization procedures are given in Appendix~\ref{app:axis_diagnostics}. All diagnostics use quantities available along the ordinary denoising trajectory and require no additional denoiser forward pass.

\subsection{Opportunity Detection and Selective Adaptation}
\label{opportunity_detection}

Action selection alone is insufficient: a useful alternative in a high-opportunity state may be harmful or irrelevant elsewhere. We therefore estimate whether the current state warrants deviation from the fixed action. Let $d_j(z)$ denote the observable scalar diagnostic for axis $j$. Using validation prompts only, we partition the empirical distribution of $d_j$ into quantile bins and assign each bin the mean validation opportunity of its states. For a held-out state,
\begin{equation}
\widetilde g_j(z)
=
\frac{1}{|\mathcal V_b|}
\sum_{z'\in\mathcal V_b} g_j(z'),
\qquad
d_j(z)\in b,
\label{eq:opportunity_detector}
\end{equation}
where $\mathcal V_b$ is the validation set in bin $b$. No held-out rollout utility is used to calibrate $\widetilde g_j$.

Given the diagnostic action $\widehat a_j(z)=\mathcal D_j(z)$ and a threshold $\tau$ fixed on validation data, the selective policy is
\begin{equation}
\pi^{\mathrm{sel}}_{j,\tau}(z)
=
\begin{cases}
\widehat a_j(z), & \widetilde g_j(z)\ge\tau,\\
\widehat a_{\mathrm{fix},j}^{\,\mathrm{val}}, & \text{otherwise}.
\end{cases}
\label{eq:selective_policy}
\end{equation}
Thus, the axis diagnostic determines \emph{what} alternative to take, while the opportunity detector determines \emph{whether} to deviate from the validation-selected fixed action.

We evaluate opportunity predictability with AUROC for $g_j(z)>0$, Spearman correlation with continuous $g_j(z)$, positive-state precision/recall, and opportunity capture across coverage levels. We rank held-out states by $\widetilde g_j(z)$ and activate adaptation on the top
$c\in\{5,10,20,50,100\}\%$. If $S_{j,c}$ denotes the selected set, the fraction of total candidate-set oracle opportunity captured at coverage $c$ is
\begin{equation}
C_{\mathrm{det},j}(c)
=
\frac{
\sum_{z\in S_{j,c}} g_j(z)
}{
\sum_z g_j(z)
}.
\label{eq:captured_opportunity}
\end{equation}
We separately report realized one-step utility lift over the validation-selected fixed action, since identifying a high-opportunity state and selecting a useful alternative are distinct sources of error.

To separate detector quality from intrinsic opportunity concentration, we compare against
$C_j^\star(c)=
\sum_{z\in\operatorname{Top}_c(g_j)}g_j(z)/
\sum_z g_j(z)$
and report capture efficiency
$C_{\mathrm{det},j}(c)/C_j^\star(c)$.

\section{Experiments}
\label{experiments}

We evaluate five questions that follow from the selective formulation: \emph{(i)} how much adaptation opportunity exists, \emph{(ii)} how concentrated that opportunity is across states, \emph{(iii)} whether the axis diagnostic can choose a useful alternative action when adaptation is warranted, \emph{(iv)} whether opportunity can be predicted before acting, and \emph{(v)} how much one-step utility can selective adaptation recover at limited coverage. End-to-end multi-step generation is treated separately in Section~\ref{exp:e4}.

We evaluate LLaDA-8B-Instruct~\citep{nie2025large}, LLaDA-1.5~\citep{zhu2025llada}, and Dream-7B~\citep{ye2025dream} on a fixed ten-task suite: CSV Missing Cells, HumanEval, Multi-Span Cloze, GSM8K, Carry RTL, TD07 Sparse Mask, JSON Mode Eval, Constrained JSON Fill, Unique List Commit, and HTML Close Tags. The suite is fixed before inspecting selective-adaptation results. Full checkpoint identifiers and task details are provided in Appendix~\ref{app:experimental_details}.

\paragraph{Protocol.}
For each task, we sample eight spaced states per evaluation prompt and estimate candidate state--action utility with four continuation rollouts. The intervention horizon is one step and temperature is $0$. Candidate actions share rollout seeds within each fold. To mitigate maximization bias in candidate-set oracle quantities, we use two-fold rollout cross-fitting: actions selected using rollouts ${0,1}$ are evaluated using ${2,3}$, and vice versa, before averaging the estimates. Validation prompts choose the best fixed action, fit the stage-only baseline, calibrate diagnostic thresholds, and calibrate the opportunity detector. Opportunity-ranking and coverage-dependent analyses use disjoint held-out prompts. Throughout, ``oracle'' denotes the cross-fitted candidate-set oracle over the evaluated actions under Eq.~\eqref{eq:state_action_utility}, not an unconstrained optimal decoder. Dataset sizes, state sampling, utility estimation, and rollout cross-fitting details are provided in Appendix~\ref{app:evaluation_protocol}.

\paragraph{Metrics.}
For opportunity detection we report AUROC for $g_j(z)>0$, Spearman correlation with $g_j(z)$, positive-state precision/recall, captured opportunity $C_{\mathrm{det},j}(c)$, and realized selective lift over the validation-selected fixed action. Coverage is evaluated at $c\in\{5,10,20,50,100\}\%$. We report point estimates in the main tables and prompt-cluster bootstrap 95\% confidence intervals in Appendix~\ref{app:complete_results}, with detector robustness and coverage controls in Tables~\ref{tab:app-detector-robustness} and~\ref{tab:app-coverage-controls}. The intervals use 1,000 prompt-level bootstrap replicates; random-coverage controls use 10,000 draws.

\subsection{How Much Adaptation Opportunity Exists?}
\label{exp:value}

\begin{figure}[t]
\centering
\caption{\textbf{Cross-fitted candidate-set adaptation opportunity.}
Each marker reports the candidate-set oracle gap for one model--task pair,
using the hindsight-selected best fixed action on the same evaluation states.
Color denotes the model, and marker shape denotes the strongest adaptation axis.}
\label{fig:e1-all}
\includegraphics[width=\linewidth]{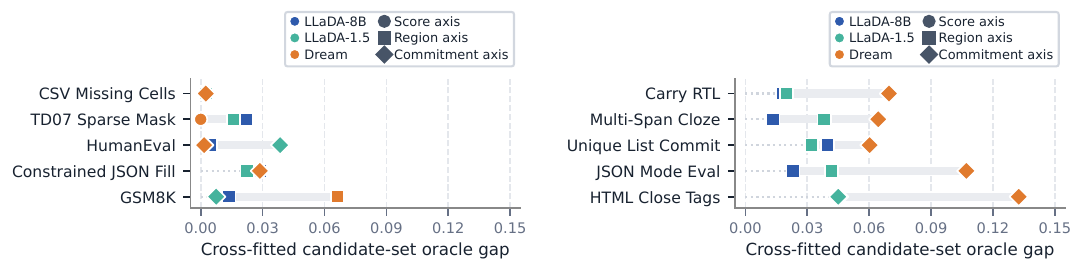}
\end{figure}

Figure~\ref{fig:e1-all} reports the cross-fitted candidate-set oracle gap for the strongest axis of each model--task pair, relative to the hindsight-selected best fixed action. The result is heterogeneity rather than uniformly large gains. LLaDA-8B gaps range from $0.0025$ to $0.0482$, LLaDA-1.5 from $0.0025$ to $0.0450$, and Dream from $0$ to $0.1325$. Several tasks admit little room over the fixed action, while structured tasks can expose larger one-step opportunities. The strongest axis also changes across models. Axis-wise results, including confidence intervals and bidirectional reversal mass, appear in Appendix Table~\ref{tab:app-e1-full}. We examine how held-out validation-fixed opportunity is distributed across states.

\subsection{How Concentrated Is Adaptation Opportunity?}
\label{exp:concentration}

The reversal decomposition suggests that a modest average gap may arise from rare, high-margin states. We therefore distinguish two complementary quantities. The \emph{positive-opportunity rate} is the held-out fraction $\Pr[g_j(z)>0]$ relative to the validation-selected fixed action. The \emph{bidirectional reversal mass} is a utility-weighted pairwise quantity,
$B_j=\max_{a,b}\min\{\mathbb E[D_{a,b}(z)_+],\mathbb E[(-D_{a,b}(z))_+]\}$.
For example, LLaDA-1.5 Carry RTL/region has an $11.4\%$ positive-opportunity rate but bidirectional reversal mass $0.0200$; these values are not expected to coincide. Table~\ref{tab:reversal-anatomy} reports both quantities, and complete axis-wise results for all three models are provided in Appendix Table~\ref{tab:app-e1-full}.

\begin{table}[!htbp]
\centering

\caption{\textbf{Opportunity frequency and bidirectional reversal mass on LLaDA-1.5.} Positive states use 640 held-out states (384 for JSON Mode Eval) and a validation-selected fixed action. Bidirectional reversal mass uses 800 pooled states (544 for JSON Mode Eval) and measures utility-weighted pairwise reversals, not percentages.}
\label{tab:reversal-anatomy}
\small
\setlength{\tabcolsep}{5pt}
\begin{tabular}{llrr}
\toprule
Task & Axis & Positive states & Bidir.\ mass \\
\midrule
CSV Missing Cells      & region     & 0.0\%  & 0.0025 \\
Multi-Span Cloze       & region     & 9.2\%  & 0.0401 \\
Constrained JSON Fill  & region     & 13.6\% & 0.0225 \\
TD07 Sparse Mask       & region     & 4.5\%  & 0.0238 \\
JSON Mode Eval         & region     & 7.0\%  & 0.0418 \\
Carry RTL              & region     & 11.4\% & 0.0200 \\
HTML Close Tags        & commitment & 7.8\%  & 0.0325 \\
Unique List Commit     & region     & 12.8\% & 0.0420 \\
\bottomrule
\end{tabular}
\end{table}

We separate intrinsic opportunity concentration from detector quality. $C^\star_c$ ranks states by the true held-out $g_j(z)$, while $C^{\mathrm{det}}_c$ ranks them by the validation-calibrated detector. Table~\ref{tab:opportunity-concentration} shows that opportunity can be highly concentrated in some regimes, while detector capture varies substantially.

\begin{table}[t]
\centering
\caption{\textbf{Opportunity concentration versus detectability.} $C^\star_c$ denotes candidate-set oracle opportunity captured by top-$c$ states, while $C^{\mathrm{det}}_c$ denotes opportunity captured by detector-ranked states.}
\label{tab:opportunity-concentration}
\scriptsize
\setlength{\tabcolsep}{2.2pt}
\begin{tabular}{lllrrrrrrr}
\toprule
Model & Task & Axis & Pos. & $C^\star_5$ & $C^\star_{10}$ & $C^\star_{20}$ & $C^{\rm det}_5$ & $C^{\rm det}_{10}$ & $C^{\rm det}_{20}$ \\
\midrule
LLaDA-8B  & Constrained JSON Fill & region & 8.0\% & 62.7\% & 100.0\% & 100.0\% & 35.3\% & 56.9\% & 84.3\% \\
Dream     & Carry RTL             & cardinality & 4.2\% & 100.0\% & 100.0\% & 100.0\% & 42.9\% & 64.3\% & 71.4\% \\
Dream     & HTML Close Tags       & region & 4.7\% & 100.0\% & 100.0\% & 100.0\% & 16.7\% & 36.7\% & 66.7\% \\
LLaDA-1.5 & Carry RTL             & region & 11.4\% & 53.9\% & 89.9\% & 100.0\% & 4.5\% & 9.0\% & 31.5\% \\
\bottomrule
\end{tabular}
\end{table}

\subsection{Can Axis Diagnostics Choose Useful Deviations?}
\label{exp:action-selection}

We first evaluate action selection independently of opportunity detection. We use an exploratory held-out viability screen: a setting passes when its regret is lower than the validation-selected stage-only baseline and it recovers positive cross-fitted candidate-set oracle gap. Because this screen uses held-out outcomes, it characterizes regimes rather than serving as an independent confirmatory test. Seven task--model--axis combinations pass: two for LLaDA-8B, two for LLaDA-1.5, and three for Dream, with no task--axis pair passing across all three models. The clearest cross-model replication is HTML Close Tags/region for LLaDA-8B and Dream, supporting regime-specific action proposals rather than a universal diagnostic mechanism. Complete results are reported in Appendix Tables~\ref{tab:app-e2-stats},~\ref{tab:app-e2-full}, and~\ref{tab:app-e3-full}.

\subsection{Can Opportunity Be Predicted Before Acting?}
\label{exp:predict}

We next evaluate the opportunity detector on held-out states. Model-level mean AUROC is approximately $0.545$ for LLaDA-8B, $0.541$ for LLaDA-1.5, and $0.571$ for Dream, and only $3$, $1$, and $5$ task--axis settings, respectively, have positive $10\%$ selective lift. Opportunity predictability is therefore substantially more limited than the mere existence of an oracle gap.

Several regimes are clear. Table~\ref{tab:opportunity-detector} shows successes and failures. LLaDA-8B Constrained JSON Fill/region reaches $0.854$ AUROC and captures $56.9\%$ of oracle opportunity at $10\%$ coverage, while Dream Carry RTL/cardinality captures $64.3\%$ with $0.802$ AUROC. Other settings are near chance. Notably, Dream JSON Mode Eval/score attains $0.891$ AUROC but captures no oracle utility mass in the top $10\%$ and yields no lift, showing that binary detection and utility-relevant ranking are distinct. Results and bootstrap intervals appear in Appendix Tables~\ref{tab:app-opportunity-detection} and~\ref{tab:app-detector-robustness}.

\begin{table}[t]
\centering
\caption{\textbf{Held-out opportunity detection.} $C_{10}$ is detector-captured candidate-set oracle opportunity at $10\%$ coverage; $L_{10}$ is realized one-step lift over the validation-selected fixed action. Prompt-cluster bootstrap 95\% confidence intervals are reported in Appendix~\ref{app:complete_results}.}
\label{tab:opportunity-detector}
\scriptsize
\setlength{\tabcolsep}{3.0pt}
\begin{tabular}{lllrrrrr}
\toprule
Model & Task & Axis & AUROC & Spearman & Pos. states & $C_{10}$ & $L_{10}$ \\
\midrule
LLaDA-8B  & Constrained JSON Fill & region      & .854 & .381 & 8.0\% & 56.9\% & +.0453 \\
Dream     & Carry RTL             & cardinality & .802 & .368 & 4.2\% & 64.3\% & +.0089 \\
Dream     & HTML Close Tags       & region      & .804 & .261 & 4.7\% & 36.7\% & +.0094 \\
LLaDA-1.5 & Carry RTL             & region      & .689 & .242 & 11.4\% & 9.0\%  & +.0036 \\
\midrule
Dream     & JSON Mode Eval        & score       & .891 & .136 & 0.5\% & 0.0\%  & +.0000 \\
LLaDA-1.5 & JSON Mode Eval        & region      & .542 & .051 & 7.0\% & 14.1\% & +.0000 \\
LLaDA-8B  & HTML Close Tags       & region      & .550 & .035 & 2.8\% & 11.1\% & +.0031 \\
\bottomrule
\end{tabular}
\end{table}

\subsection{How Much One-Step Utility Can Selective Adaptation Recover?}
\label{exp:selective-coverage}

Table~\ref{tab:selective-coverage} reports coverage-dependent selective utility for three representative successful regimes. For LLaDA-8B Constrained JSON Fill/region, the top $5\%$, $10\%$, and $20\%$ of detector-ranked states capture $35.3\%$, $56.9\%$, and $84.3\%$ of the available one-step candidate-set oracle opportunity. The realized one-step lift saturates by $20\%$ coverage. Dream Carry RTL/cardinality shows a similar pattern, whereas Dream HTML Close Tags/region requires broader coverage. Complete coverage-dependent selective-utility results are provided in Appendix Table~\ref{tab:app-opportunity-full}.

\begin{table}[!htbp]
\centering
\caption{\textbf{Coverage-dependent selective utility for representative successful regimes.} Each entry reports the fraction of total candidate-set oracle opportunity captured / realized average one-step utility lift over the validation-selected fixed action.}
\label{tab:selective-coverage}
\scriptsize
\setlength{\tabcolsep}{3.3pt}
\begin{tabular}{rccc}
\toprule
Coverage
& LLaDA-8B: Constrained JSON / region
& Dream: Carry / cardinality
& Dream: HTML / region \\
\midrule
5\%   & 35.3\% / +.0281 & 42.9\% / +.0057 & 16.7\% / +.0016 \\
10\%  & 56.9\% / +.0453 & 64.3\% / +.0089 & 36.7\% / +.0094 \\
20\%  & 84.3\% / +.0594 & 71.4\% / +.0099 & 66.7\% / +.0188 \\
50\%  & 88.2\% / +.0594 & 82.1\% / +.0099 & 90.0\% / +.0297 \\
100\% & 100\% / +.0594  & 100\% / +.0099  & 100\% / +.0297 \\
\bottomrule
\end{tabular}
\end{table}

We compare detector gating with random-state coverage and an oracle gate. Random coverage tests whether the detector is doing more than selecting an arbitrary fraction of states. The oracle gate ranks states by true held-out $g_j(z)$ but applies the same diagnostic action, separating gating error from action-selection error. Table~\ref{tab:coverage-controls} reports both controls; bootstrap intervals and the corresponding concentration/random-coverage controls are provided in Appendix Table~\ref{tab:app-coverage-controls}.

\begin{table}[!htbp]
\centering
\caption{\textbf{Coverage controls at $10\%$ coverage.} Random $C_{10}$ averages opportunity capture over 10,000 uniform state selections. Oracle-gate $C_{10}$ ranks states by held-out candidate-set opportunity, while oracle-gate lift applies the same diagnostic action only to those oracle-selected states.}
\label{tab:coverage-controls}
\scriptsize
\setlength{\tabcolsep}{2.5pt}
\begin{tabular}{lllrrrrr}
\toprule
Model & Task & Axis & Random $C_{10}$ & Detector $C_{10}$ & Oracle-gate $C_{10}$ & Detector lift & Oracle-gate diagnostic lift \\
\midrule
LLaDA-8B & Constrained JSON Fill & region & 10.0\% & 56.9\% & 100.0\% & +0.0453 & +0.0625 \\
Dream    & Carry RTL             & cardinality & 10.0\% & 64.3\% & 100.0\% & +0.0089 & +0.0099 \\
Dream    & HTML Close Tags       & region & 10.1\% & 36.7\% & 100.0\% & +0.0094 & +0.0375 \\
\bottomrule
\end{tabular}
\end{table}

The current evidence therefore supports a conditional statement: when opportunity is predictable and the action diagnostic is useful, selective one-step adaptation can recover substantial local utility at limited coverage. Many settings do not satisfy those conditions.

\subsection{End-to-End Selective Decoding}
\label{exp:e4}

The preceding analyses measure one-step utility rather than
terminal generation quality. We therefore evaluate selective
decoding on a disjoint final-test split against fixed,
always-diagnostic, and coverage-matched random adaptation,
alongside reference and stage-only baselines.
Table~\ref{tab:e4-selective-decoding} presents one setting per
model, while results across all ten tasks and baselines
are reported in Appendix Table~\ref{tab:app-mini-e4-full}.

\begin{table}[htbp]
\centering
\caption{\textbf{End-to-end selective decoding.}
Terminal utility on representative settings.
$\Delta_F$ and $\Delta_R$ denote Selective$-$Fixed and
Selective$-$Random, respectively.
Random matches selective activation counts per prompt.
Full results are in Appendix Table~\ref{tab:app-mini-e4-full}.}
\label{tab:e4-selective-decoding}
\scriptsize
\setlength{\tabcolsep}{2.5pt}
\begin{tabular}{lllrrrrrrr}
\toprule
Model & Task & Axis & Fixed & Always & Rand. &
Sel. & $\Delta_F$ & $\Delta_R$ & Cov. \\
\midrule
LLaDA-8B & Unique List Commit & region
& 0.365 & 0.160 & 0.395 & \textbf{0.580}
& +0.215 & \textbf{+0.185} & 54.0\% \\

LLaDA-1.5 & JSON Mode Eval & region
& 0.423 & 0.384 & 0.461 & \textbf{0.645}
& +0.222 & \textbf{+0.184} & 20.0\% \\

Dream & Carry RTL & commitment
& 0.467 & 0.367 & 0.500 & \textbf{0.636}
& +0.169 & \textbf{+0.136} & 45.0\% \\
\bottomrule
\end{tabular}
\end{table}

In all three examples, unconditional adaptation degrades
performance, whereas selective adaptation outperforms both
Fixed and coverage-matched Random. Across the ten-task suite,
Selective improves task-macro utility from 0.478 to 0.523
(LLaDA-8B), 0.317 to 0.391 (LLaDA-1.5), and 0.428 to 0.497
(Dream), yielding gains of +4.5, +7.4, and +7.0 points.

More importantly, in 12 of 30 model--task settings, Always
underperforms Fixed while Selective exceeds Fixed, Always,
and Random; all three selective contrasts have positive
paired 95\% bootstrap intervals in 10 of these cases.
Dream/GSM8K instead reflects an alternative-rule gain at
100\% coverage, not state-aware timing.

\section{Conclusion}
\label{conclusion}

We study masked-diffusion unmasking as a selective decision problem, using a five-axis taxonomy and a reversal decomposition to show that adaptation value depends on both how often the preferred action changes and how large those local gains are. Empirically, we find that adaptation opportunity is highly heterogeneous across tasks, models, and decision axes: fixed actions remain competitive in many settings, while useful gains concentrate in specific regimes. We further show that the existence of oracle opportunity does not imply that it can be predicted or exploited, motivating a separation between action selection and opportunity detection. These results support selective, regime-dependent adaptation at the transition level, where a fixed action is retained by default and adaptation is activated only when predictable opportunity is present.

\bibliographystyle{iclr2027_conference}
\bibliography{ref}

@inproceedings{shi2024simplified,
  title={Simplified and Generalized Masked Diffusion for Discrete Data},
  author={Shi, Jiaxin and Han, Kehang and Wang, Zhe and Doucet, Arnaud and Titsias, Michalis K.},
  booktitle={Advances in Neural Information Processing Systems},
  year={2024}
}

@inproceedings{
austin2021structured,
title={Structured Denoising Diffusion Models in Discrete State-Spaces},
author={Jacob Austin and Daniel D. Johnson and Jonathan Ho and Daniel Tarlow and Rianne van den Berg},
booktitle={Advances in Neural Information Processing Systems},
editor={A. Beygelzimer and Y. Dauphin and P. Liang and J. Wortman Vaughan},
year={2021},
url={https://openreview.net/forum?id=h7-XixPCAL}
}

@INPROCEEDINGS{chang2022maskgit,
  author={Chang, Huiwen and Zhang, Han and Jiang, Lu and Liu, Ce and Freeman, William T.},
  booktitle={2022 IEEE/CVF Conference on Computer Vision and Pattern Recognition (CVPR)}, 
  title={MaskGIT: Masked Generative Image Transformer}, 
  year={2022},
  volume={},
  number={},
  pages={11305-11315},
  doi={10.1109/CVPR52688.2022.01103}}

@inproceedings{ghazvininejad2019maskpredict,
    title = "Mask-Predict: Parallel Decoding of Conditional Masked Language Models",
    author = "Ghazvininejad, Marjan  and
      Levy, Omer  and
      Liu, Yinhan  and
      Zettlemoyer, Luke",
    editor = "Inui, Kentaro  and
      Jiang, Jing  and
      Ng, Vincent  and
      Wan, Xiaojun",
    booktitle = "Proceedings of the 2019 Conference on Empirical Methods in Natural Language Processing and the 9th International Joint Conference on Natural Language Processing (EMNLP-IJCNLP)",
    month = nov,
    year = "2019",
    address = "Hong Kong, China",
    publisher = "Association for Computational Linguistics",
    url = "https://aclanthology.org/D19-1633/",
    doi = "10.18653/v1/D19-1633",
    pages = "6112--6121"
}

@misc{lou2023discrete,
      title={Discrete Diffusion Modeling by Estimating the Ratios of the Data Distribution}, 
      author={Aaron Lou and Chenlin Meng and Stefano Ermon},
      year={2024},
      eprint={2310.16834},
      archivePrefix={arXiv},
      primaryClass={stat.ML},
      url={https://arxiv.org/abs/2310.16834}, 
}

@inproceedings{hong2026improving,
title={Improving Discrete Diffusion Unmasking Policies Beyond Explicit Reference Policies},
author={Chunsan Hong and Seonho An and Min-Soo Kim and Jong Chul Ye},
booktitle={The Fourteenth International Conference on Learning Representations},
year={2026},
url={https://openreview.net/forum?id=on6cb46OhD}
}

@inproceedings{
    sahoo2024simple,
    title={Simple and Effective Masked Diffusion Language Models},
    author={Subham Sekhar Sahoo and Marianne Arriola and Aaron Gokaslan and Edgar Mariano Marroquin and Alexander M Rush and Yair Schiff and Justin T Chiu and Volodymyr Kuleshov},
    booktitle={The Thirty-eighth Annual Conference on Neural Information Processing Systems},
    year={2024},
    url={https://openreview.net/forum?id=L4uaAR4ArM}
    }

@inproceedings{
nie2025scaling,
title={Scaling up Masked Diffusion Models on Text},
author={Shen Nie and Fengqi Zhu and Chao Du and Tianyu Pang and Qian Liu and Guangtao Zeng and Min Lin and Chongxuan Li},
booktitle={The Thirteenth International Conference on Learning Representations},
year={2025},
url={https://openreview.net/forum?id=WNvvwK0tut}
}

@inproceedings{nie2025large,
 author = {Nie, Shen and Zhu, Fengqi and You, Zebin and Zhang, Xiaolu and Ou, Jingyang and Hu, Jun and Zhou, Jun and Lin, Yankai and Wen, Ji-Rong and LI, Chongxuan},
 booktitle = {Advances in Neural Information Processing Systems},
 doi = {10.52202/085713-1689},
 editor = {D. Belgrave and C. Zhang and H. Lin and R. Pascanu and P. Koniusz and M. Ghassemi and N. Chen},
 pages = {50608--50646},
 publisher = {Curran Associates, Inc.},
 title = {Large Language Diffusion Models},
 url = {https://proceedings.neurips.cc/paper_files/paper/2025/file/48b383b24230e0e6e649d9c98dae4d8c-Paper-Conference.pdf},
 volume = {38},
 year = {2025}
}

@article{ye2025dream,
  title={Dream 7B: Diffusion Large Language Models},
  author={Ye, Jiacheng and Xie, Zhihui and Zheng, Lin and Gao, Jiahui and Wu, Zirui and Jiang, Xin and Li, Zhenguo and Kong, Lingpeng},
  journal={arXiv preprint arXiv:2508.15487},
  year={2025}
}

@inproceedings{
wu2026fastdllm,
title={Fast-d{LLM}: Training-free Acceleration of Diffusion {LLM} by Enabling {KV} Cache and Parallel Decoding},
author={Chengyue Wu and Hao Zhang and Shuchen Xue and Zhijian Liu and Shizhe Diao and Ligeng Zhu and Ping Luo and Song Han and Enze Xie},
booktitle={The Fourteenth International Conference on Learning Representations},
year={2026},
url={https://openreview.net/forum?id=3Z3Is6hnOT}
}

@inproceedings{
kim2025klass,
title={{KLASS}: {KL}-Guided Fast Inference in Masked Diffusion Models},
author={Seo Hyun Kim and Sunwoo Hong and Hojung Jung and Youngrok Park and Se-Young Yun},
booktitle={The Thirty-ninth Annual Conference on Neural Information Processing Systems},
year={2025},
url={https://openreview.net/forum?id=gOG9Zoyn4R}
}

@inproceedings{
luxembourg2026plan,
title={Plan for Speed: Dilated Scheduling for Masked Diffusion Language Models},
author={Omer Luxembourg and Haim H. Permuter and Eliya Nachmani},
booktitle={Forty-third International Conference on Machine Learning},
year={2026},
url={https://openreview.net/forum?id=DbzAsMRNhy}
}

@inproceedings{
wang2025remasking,
title={Remasking Discrete Diffusion Models with Inference-Time Scaling},
author={Guanghan Wang and Yair Schiff and Subham Sekhar Sahoo and Volodymyr Kuleshov},
booktitle={The Thirty-ninth Annual Conference on Neural Information Processing Systems},
year={2025},
url={https://openreview.net/forum?id=IJryQAOy0p}
}

@inproceedings{
hong2026widein,
title={Wide-In, Narrow-Out: Revokable Decoding for Efficient and Effective {DLLM}s},
author={Feng Hong and Geng Yu and Yushi Ye and Haicheng Huang and Huangjie Zheng and Ya Zhang and Yanfeng Wang and Jiangchao Yao},
booktitle={The Fourteenth International Conference on Learning Representations},
year={2026},
url={https://openreview.net/forum?id=XtLQHlNLxy}
}

@inproceedings{
yang2026improving,
title={Improving Sampling for Masked Diffusion Models via Information Gain},
author={Kaisen Yang and Jayden Teoh and Kaicheng Yang and Yitong Zhang and Alex Lamb},
booktitle={Forty-third International Conference on Machine Learning},
year={2026},
url={https://openreview.net/forum?id=1HPTFpwktA}
}

@misc{peng2025pathplanning,
      title={Path Planning for Masked Diffusion Model Sampling}, 
      author={Fred Zhangzhi Peng and Zachary Bezemek and Sawan Patel and Jarrid Rector-Brooks and Sherwood Yao and Avishek Joey Bose and Alexander Tong and Pranam Chatterjee},
      year={2026},
      eprint={2502.03540},
      archivePrefix={arXiv},
      primaryClass={cs.LG},
      url={https://arxiv.org/abs/2502.03540}, 
}

@inproceedings{
turok2026duel,
title={{DUEL}: Exact Likelihood for Masked Diffusion via Deterministic Unmasking},
author={Gilad Turok and Yair Schiff and Christopher De Sa and Volodymyr Kuleshov},
booktitle={ICML 2026 Workshop on Foundations of Deep Generative Models: Understanding Memorization, Generalization, and Reasoning},
year={2026},
url={https://openreview.net/forum?id=FQA0rTzTVd}
}

@article{hong2026unifying,
  title={Unifying Masked Diffusion Models with Various Generation Orders and Beyond},
  author={Hong, Chunsan and Lee, Sanghyun and Ye, Jong Chul},
  journal={arXiv preprint arXiv:2602.02112},
  year={2026}
}

@inproceedings{
arriola2025block,
title={Block Diffusion: Interpolating Between Autoregressive and Diffusion Language Models},
author={Marianne Arriola and Subham Sekhar Sahoo and Aaron Gokaslan and Zhihan Yang and Zhixuan Qi and Jiaqi Han and Justin T Chiu and Volodymyr Kuleshov},
booktitle={The Thirteenth International Conference on Learning Representations},
year={2025},
url={https://openreview.net/forum?id=tyEyYT267x}
}

@inproceedings{
seo2025fast,
title={Fast and Fluent Diffusion Language Models via Convolutional Decoding and Rejective Fine-tuning},
author={Yeongbin Seo and Dongha Lee and Jaehyung Kim and Jinyoung Yeo},
booktitle={The Thirty-ninth Annual Conference on Neural Information Processing Systems},
year={2025},
url={https://openreview.net/forum?id=HvIRFV0J90}
}

@misc{benhamu2025accelerated,
      title={Accelerated Sampling from Masked Diffusion Models via Entropy Bounded Unmasking}, 
      author={Heli Ben-Hamu and Itai Gat and Daniel Severo and Niklas Nolte and Brian Karrer},
      year={2025},
      eprint={2505.24857},
      archivePrefix={arXiv},
      primaryClass={cs.LG},
      url={https://arxiv.org/abs/2505.24857}, 
}

@inproceedings{zhou2026dos,
    title = "{DOS}: Dependency-Oriented Sampler for Masked Diffusion Language Models",
    author = "Zhou, Xueyu  and
      Hu, Yangrong  and
      Huang, Jian",
    editor = "Liakata, Maria  and
      Moreira, Viviane P.  and
      Zhang, Jiajun  and
      Jurgens, David",
    booktitle = "Findings of the {A}ssociation for {C}omputational {L}inguistics: {ACL} 2026",
    month = jul,
    year = "2026",
    address = "San Diego, California, United States",
    publisher = "Association for Computational Linguistics",
    url = "https://aclanthology.org/2026.findings-acl.861/",
    doi = "10.18653/v1/2026.findings-acl.861",
    pages = "17404--17419",
    ISBN = "979-8-89176-395-1"
}

@misc{ringel2026demask,
      title={Dependency-Guided Parallel Decoding in Discrete Diffusion Language Models}, 
      author={Liran Ringel and Ameen Ali and Yaniv Romano},
      year={2026},
      eprint={2604.02560},
      archivePrefix={arXiv},
      primaryClass={cs.CL},
      url={https://arxiv.org/abs/2604.02560}, 
}

@misc{luo2026dsb,
      title={DSB: Dynamic Sliding Block Scheduling for Diffusion LLMs}, 
      author={Lizhuo Luo and Shenggui Li and Yonggang Wen and Tianwei Zhang},
      year={2026},
      eprint={2602.05992},
      archivePrefix={arXiv},
      primaryClass={cs.CL},
      url={https://arxiv.org/abs/2602.05992}, 
}

@misc{zhang2026swordsman,
      title={Swordsman: Entropy-Driven Adaptive Block Partition for Efficient Diffusion Language Models}, 
      author={Yu Zhang and Xinchen Li and Jialei Zhou and Hongnan Ma and Zhongwei Wan and Yiwei Shi and Duoqian Miao and Qi Zhang and Longbing Cao},
      year={2026},
      eprint={2602.04399},
      archivePrefix={arXiv},
      primaryClass={cs.CL},
      url={https://arxiv.org/abs/2602.04399}, 
}

@misc{shu2026dcd,
      title={Deferred Commitment Decoding for Diffusion Language Models}, 
      author={Yingte Shu and Yuchuan Tian and Chao Xu and Yunhe Wang and Hanting Chen},
      year={2026},
      eprint={2601.02076},
      archivePrefix={arXiv},
      primaryClass={cs.CL},
      url={https://arxiv.org/abs/2601.02076}, 
}

@misc{wang2026vsb,
      title={When to Commit? Towards Variable-Size Self-Contained Blocks for Discrete Diffusion Language Models}, 
      author={Danny Wang and Ruihong Qiu and Zi Huang},
      year={2026},
      eprint={2604.23994},
      archivePrefix={arXiv},
      primaryClass={cs.LG},
      url={https://arxiv.org/abs/2604.23994}, 
}

@inproceedings{
zhai2026core,
title={{CORE}: Context-Robust Remasking for Diffusion Language Models},
author={Kevin Zhai and Sabbir Mollah and Zhenyi Wang and Mubarak Shah},
booktitle={Forty-third International Conference on Machine Learning},
year={2026},
url={https://openreview.net/forum?id=bmKHxLWkz9}
}

@misc{lee2025lookum,
      title={Lookahead Unmasking Elicits Accurate Decoding in Diffusion Language Models}, 
      author={Sanghyun Lee and Seungryong Kim and Jongho Park and Dongmin Park},
      year={2025},
      eprint={2511.05563},
      archivePrefix={arXiv},
      primaryClass={cs.LG},
      url={https://arxiv.org/abs/2511.05563}, 
}

@misc{cao2026soar,
      title={Search or Accelerate: Confidence-Switched Position Beam Search for Diffusion Language Models}, 
      author={Mingyu Cao and Alvaro H. C. Correia and Christos Louizos and Shiwei Liu and Lu Yin},
      year={2026},
      eprint={2602.10953},
      archivePrefix={arXiv},
      primaryClass={cs.CL},
      url={https://arxiv.org/abs/2602.10953}, 
}

@misc{shen2026ordertoken,
      title={Improving Diffusion Language Model Decoding through Joint Search in Generation Order and Token Space}, 
      author={Yangyi Shen and Tianjian Feng and Jiaqi Han and Wen Wang and Tianlang Chen and Chunhua Shen and Jure Leskovec and Stefano Ermon},
      year={2026},
      eprint={2601.20339},
      archivePrefix={arXiv},
      primaryClass={cs.CL},
      url={https://arxiv.org/abs/2601.20339}, 
}

@inproceedings{huang2026medal,
    title = "Diffusion Language Model Inference with {M}onte {C}arlo Tree Search",
    author = "Huang, Zheng  and
      Ramnath, Kiran  and
      Chen, Yueyan  and
      Feng, Aosong  and
      Woo, Sangmin  and
      Srinivasan, Balasubramaniam  and
      Xu, Zhichao  and
      Zhou, Kang  and
      Wang, Shuai  and
      Ding, Haibo  and
      Cheong, Lin Lee",
    editor = "Demberg, Vera  and
      Inui, Kentaro  and
      Marquez, Llu{\'i}s",
    booktitle = "Findings of the {A}ssociation for {C}omputational {L}inguistics: {EACL} 2026",
    month = mar,
    year = "2026",
    address = "Rabat, Morocco",
    publisher = "Association for Computational Linguistics",
    url = "https://aclanthology.org/2026.findings-eacl.180/",
    doi = "10.18653/v1/2026.findings-eacl.180",
    pages = "3493--3512",
    ISBN = "979-8-89176-386-9"
}

@inproceedings{
liu2025think,
title={Think while You Generate: Discrete Diffusion with Planned Denoising},
author={Sulin Liu and Juno Nam and Andrew Campbell and Hannes Stark and Yilun Xu and Tommi Jaakkola and Rafael Gomez-Bombarelli},
booktitle={The Thirteenth International Conference on Learning Representations},
year={2025},
url={https://openreview.net/forum?id=MJNywBdSDy}
}

@inproceedings{
kim2025train,
title={Train for the Worst, Plan for the Best: Understanding Token Ordering in Masked Diffusions},
author={Jaeyeon Kim and Kulin Shah and Vasilis Kontonis and Sham M. Kakade and Sitan Chen},
booktitle={Forty-second International Conference on Machine Learning},
year={2025},
url={https://openreview.net/forum?id=DjJmre5IkP}
}

@article{zhu2025llada,
  title={LLaDA 1.5: Variance-Reduced Preference Optimization for Large Language Diffusion Models},
  author={Zhu, Fengqi and Wang, Rongzhen and Nie, Shen and Zhang, Xiaolu and Wu, Chunwei and Hu, Jun and Zhou, Jun and Chen, Jianfei and Lin, Yankai and Wen, Ji-Rong and others},
  journal={arXiv preprint arXiv:2505.19223},
  year={2025}
}

@inproceedings{zhong-etal-2026-parallelism,
    title = "Parallelism and Generation Order in Masked Diffusion Language Models: Limits Today, Potential Tomorrow",
    author = "Zhong, Yangyang  and
      Gu, Yanmei  and
      Zang, Zhengqing  and
      Li, Xiaomeng  and
      Ding, Yuqi  and
      Jia, Xibei  and
      Shen, Yuting  and
      Lan, Zhenzhong  and
      Zhu, Liwang  and
      Liu, Weiping  and
      Zhou, Junlin  and
      Liu, Haisheng  and
      Yu, Zhong Xin  and
      Luo, Pengxin  and
      Qi, Donglian  and
      Yan, Yunfeng  and
      Zhao, Junbo",
    editor = "Liakata, Maria  and
      Moreira, Viviane P.  and
      Zhang, Jiajun  and
      Jurgens, David",
    booktitle = "Findings of the {A}ssociation for {C}omputational {L}inguistics: {ACL} 2026",
    month = jul,
    year = "2026",
    address = "San Diego, California, United States",
    publisher = "Association for Computational Linguistics",
    url = "https://aclanthology.org/2026.findings-acl.357/",
    doi = "10.18653/v1/2026.findings-acl.357",
    pages = "7178--7218",
    ISBN = "979-8-89176-395-1"
}

@article{DBLP:journals/corr/abs-2602-00286,
  publtype={informal},
  author={Shaorong Zhang and Longxuan Yu and Rob Brekelmans and Luhan Tang and Salman Asif and Greg Ver Steeg},
  title={Generation Order and Parallel Decoding in Masked Diffusion Models: An Information-Theoretic Perspective},
  year={2026},
  month={February},
  cdate={1769904000000},
  journal={CoRR},
  volume={abs/2602.00286},
  url={https://doi.org/10.48550/arXiv.2602.00286}
}

@misc{chen2025optimalinferenceschedulesmasked,
      title={Optimal Inference Schedules for Masked Diffusion Models}, 
      author={Sitan Chen and Kevin Cong and Jerry Li},
      year={2025},
      eprint={2511.04647},
      archivePrefix={arXiv},
      primaryClass={cs.LG},
      url={https://arxiv.org/abs/2511.04647}, 
}

@inproceedings{
jazbec2026learning,
title={Learning Unmasking Policies for Diffusion Language Models},
author={Metod Jazbec and Theo X. Olausson and Louis B{\'e}thune and Pierre Ablin and Michael Kirchhof and Joao Monteiro and Victor Guilherme Turrisi da Costa and Jason Ramapuram and marco cuturi},
booktitle={Forty-third International Conference on Machine Learning},
year={2026},
url={https://openreview.net/forum?id=F9NDKf5oPy}
}

@misc{asano2026wheretounmaskgroundtruthguidedunmaskingorder,
      title={Where-to-Unmask: Ground-Truth-Guided Unmasking Order Learning for Masked Diffusion Language Models}, 
      author={Hikaru Asano and Tadashi Kozuno and Kuniaki Saito and Yukino Baba},
      year={2026},
      eprint={2602.09501},
      archivePrefix={arXiv},
      primaryClass={cs.CL},
      url={https://arxiv.org/abs/2602.09501}, 
}


\clearpage

\appendix

\section{Expressive Scope and Mapping of Unmasking Strategies}
\label{app:taxonomy}

\subsection{Expressive Scope and Completeness}
\label{app:taxonomy_completeness}

We establish the representational completeness of the
four transition-level axes and extend it to
history-dependent unmasking policies.

\begin{theorem}[Completeness of Reveal-and-Commit Policies]
\label{thm:policy_completeness}
Fix a denoiser, token-value sampler, and revision operator.
For a state $z$ with masked positions $M(z)$, define
\[
\mathcal U(z)
=
\bigcup_{N\subseteq M(z)}
\left(\{N\}\times\{0,1\}^{|N|}\right).
\]
Let $\mathcal A(z)$ be the four-axis parameter space
and $F_z:\mathcal A(z)\to\mathcal U(z)$ its induced
reveal-and-commit map. Assume arbitrary candidate
regions and at least one scoring rule. Then:

\begin{enumerate}
    \item \textbf{Action completeness.}
    The map $F_z$ is surjective:
    \[
    F_z(\mathcal A(z))=\mathcal U(z).
    \]

    \item \textbf{Policy completeness.}
    For any stochastic, history-dependent policy
    $\pi_t(\cdot\mid h_t)$ over $\mathcal U(z_t)$,
    there exists a policy $\widetilde\pi_t$ over
    $\mathcal A(z_t)$ such that
    \[
    (F_{z_t})_{\#}\widetilde\pi_t
    (\cdot\mid h_t)
    =
    \pi_t(\cdot\mid h_t),
    \]
    where $(F_{z_t})_{\#}$ denotes the pushforward
    distribution. Under shared initial-state and
    transition kernels,
    \[
    \mathbb P_{\widetilde\pi}
    (z_{0:T},u_{0:T-1})
    =
    \mathbb P_{\pi}
    (z_{0:T},u_{0:T-1}).
    \]
\end{enumerate}
\end{theorem}

\begin{proof}
We prove the two statements in sequence.

\paragraph{(i) Action completeness.}
Fix an arbitrary state $z$ and action
$u=(N,c)\in\mathcal U(z)$.
Let $\alpha_0$ be any valid score mixture.
Construct
\[
\psi_z(N,c)
=
\left(
\alpha_0,\,
\mathbf 1_N,\,
1,\,
c
\right),
\]
where $\mathbf 1_N$ is the indicator of $N$
over the masked positions $M(z)$.

By the region definition,
\[
R(\psi_z(N,c))=N.
\]
Since $\kappa=1$, the reveal cardinality is
\[
k=\lceil\kappa|R|\rceil=|N|.
\]
Consequently, the top-$k$ operator selects
every position in $N$, independently of
the score mixture:
\[
\operatorname{TopK}_{|N|}
(N,s_{\alpha_0})=N.
\]
The commitment component is exactly $c$.
Therefore,
\[
F_z(\psi_z(N,c))=(N,c)=u.
\]
The construction also holds for
$N=\varnothing$, with $k=0$ and an empty
commitment vector.

Since $u$ was arbitrary, every element of
$\mathcal U(z)$ has a preimage under $F_z$.
Hence,
\[
F_z(\mathcal A(z))=\mathcal U(z).
\]

\paragraph{(ii) Policy completeness.}
Let $\pi_t(u\mid h_t)$ be an arbitrary
history-dependent stochastic policy,
where $h_t$ contains all information available
to the policy at decision time.

Using the canonical representation
$\psi_z$ constructed above, define
\[
\widetilde\pi_t(da\mid h_t)
=
\sum_{u\in\mathcal U(z_t)}
\pi_t(u\mid h_t)\,
\delta_{\psi_{z_t}(u)}(da),
\]
where $\delta_a$ is the Dirac measure at $a$.

For any $u\in\mathcal U(z_t)$, the induced
action probability satisfies
\begin{align*}
&\Pr_{\widetilde\pi_t}
\left(F_{z_t}(a_t)=u\mid h_t\right)\\
&=
\sum_{v\in\mathcal U(z_t)}
\pi_t(v\mid h_t)\,
\mathbf 1
\left\{
F_{z_t}(\psi_{z_t}(v))=u
\right\}\\
&=
\sum_{v\in\mathcal U(z_t)}
\pi_t(v\mid h_t)\,
\mathbf 1\{v=u\}\\
&=\pi_t(u\mid h_t),
\end{align*}
where the second equality follows from
$F_z\circ\psi_z=\mathrm{id}_{\mathcal U(z)}$.
Thus,
\[
(F_{z_t})_{\#}\widetilde\pi_t
(\cdot\mid h_t)
=
\pi_t(\cdot\mid h_t).
\]

It remains to establish trajectory equivalence.
Let $K_t(z_{t+1}\mid h_t,u_t)$ denote the
shared conditional transition kernel, including
the fixed denoiser, token sampling, and revision
dynamics.

For either policy, the joint conditional law
of the action and next state is
\[
\Pr(u_t,z_{t+1}\mid h_t)
=
\pi_t(u_t\mid h_t)
K_t(z_{t+1}\mid h_t,u_t).
\]
The equality of induced action distributions
therefore implies equality of these joint
conditional laws at every history.

Starting from the same initial-state
distribution, induction over
$t=0,\ldots,T-1$ yields
\[
\mathbb P_{\widetilde\pi}
(z_{0:T},u_{0:T-1})
=
\mathbb P_{\pi}
(z_{0:T},u_{0:T-1}),
\]
establishing policy completeness.
\end{proof}

\begin{takeaway}
\textbf{Expressive completeness.}
The five-axis framework can represent any reveal-and-commit policy under fixed denoising and revision dynamics. The four transition-level axes cover all admissible decisions, while planning determines how these decisions are selected across generation states.
\end{takeaway}

\subsection{Extended Mapping of Existing Strategies}

We provide an extended mapping of existing masked diffusion inference methods under the five axes introduced in Section~\ref{taxonomy}. The purpose of this mapping is not to reproduce the terminology used by each original work, but to characterize the inference decisions made by each method through a common set of questions: \textit{which} positions are prioritized (Score), \textit{how many} are selected (Cardinality), \textit{where} they are selected from (Region), \textit{whether} selected predictions can later be revised (Commitment), and \textit{whether} the decision explicitly evaluates future or alternative denoising trajectories (Planning).

Table~\ref{tab:extended_taxonomy} focuses on methods that directly modify the unmasking policy or decoding trajectory. We therefore omit inference optimizations that only change the computational implementation, such as KV-cache reuse or quantization, without changing the underlying unmasking decision. An em dash (--) indicates that the method does not introduce an explicit mechanism along the corresponding axis beyond its underlying sampler. Methods marked with $\dagger$ require an additional learned component beyond the pretrained denoiser.

\begin{table}[p]
\centering
\small
\renewcommand{\arraystretch}{1.35}
\caption{\textbf{Extended mapping of masked diffusion inference methods under our five-axis taxonomy.} Entries reflect our interpretation under the proposed taxonomy rather than terminology necessarily used by the original authors. ``--'' denotes no explicit additional mechanism, and $\dagger$ denotes methods requiring an additional learned component.}
\label{tab:extended_taxonomy}
\begin{tabularx}{\linewidth}{@{}l *{5}{>{\centering\arraybackslash}X}@{}}
\toprule
\textbf{Method}
& \textbf{Score}
& \textbf{Cardinality}
& \textbf{Region}
& \textbf{Commitment}
& \textbf{Planning} \\
\midrule

LLaDA\citep{nie2025large}
& Confidence
& Scheduled
& Full
& Irreversible
& -- \\

Fast-dLLM\citep{wu2026fastdllm}
& Confidence
& Threshold-adaptive
& Block
& Irreversible
& -- \\

KLASS\citep{kim2025klass}
& KL stability + confidence
& Adaptive multi-token
& Full
& Irreversible
& -- \\

EB-Sampler\citep{benhamu2025accelerated}
& Entropy bound
& Adaptive
& Full
& Irreversible
& -- \\

DOS\citep{zhou2026dos}
& Attention dependency
& Base sampler
& Full
& Irreversible
& -- \\

DEMASK$^\dagger$\citep{ringel2026demask}
& Learned pairwise dependency
& Dependency-bounded
& Full
& Irreversible
& -- \\

DUS\citep{luxembourg2026plan}
& Joint-entropy criterion
& Group-based
& Dilated groups
& Irreversible
& -- \\

DSB\citep{luo2026dsb}
& State-dependent
& Dynamic block size
& Sliding block
& Irreversible
& -- \\

Swordsman\citep{zhang2026swordsman}
& Entropy shift / confidence
& Adaptive threshold
& Adaptive blocks
& Irreversible
& -- \\

DCD\citep{shu2026dcd}
& Confidence
& Confidence-adaptive
& Sliding window
& Irreversible
& -- \\

VSB\citep{wang2026vsb}
& Future-context divergence
& Variable block size
& Self-contained blocks
& Irreversible
& Future-aware criterion \\

ReMDM\citep{wang2025remasking}
& Confidence / remask schedule
& Schedule-dependent
& Full
& Remaskable
& -- \\

WINO\citep{hong2026widein}
& Draft verification
& Parallel draft
& Active region
& Remaskable
& Draft--verify \\

CoRe\citep{zhai2026core}
& Context robustness
& Revision-budget dependent
& Full
& Remaskable
& -- \\

Info-Gain Sampler\citep{yang2026improving}
& Certainty + information gain
& Candidate-set dependent
& Full
& Irreversible
& One-step look-ahead \\

LookUM\citep{lee2025lookum}
& Path uncertainty
& Candidate-set budget
& Full
& Irreversible
& Path look-ahead \\

SOAR\citep{cao2026soar}
& Confidence
& State-adaptive
& Position beam
& Search-delayed
& Beam search \\

Order-Token Search\citep{shen2026ordertoken}
& Action likelihood
& Search-dependent
& Full
& Search-delayed
& Joint trajectory search \\

MEDAL\citep{huang2026medal}
& Confidence / search value
& Search-dependent
& Full
& Search-delayed
& MCTS \\

DDPD$^\dagger$\citep{liu2025think}
& Learned planner
& Planner-controlled
& Planner-selected
& Revisable
& Learned planning \\

\bottomrule
\end{tabularx}
\end{table}

The mapping highlights that recent approaches explore substantially different parts of the unmasking strategy space. Methods such as Fast-dLLM, KLASS, and EB-Sampler primarily alter token prioritization and the number of tokens decoded in parallel. DUS, DSB, Swordsman, DCD, and VSB introduce structured restrictions on where decoding proceeds, often jointly adapting the effective decoding granularity. ReMDM, WINO, and CoRe instead relax irreversible commitment by allowing previously decoded predictions to be reconsidered. Finally, Info-Gain Sampler, LookUM, SOAR, Order-Token Search, MEDAL, and DDPD explicitly expand the decision procedure beyond a purely local greedy choice through look-ahead, search, or learned planning.

These methods need not occupy only one axis: many combine several decisions simultaneously. The taxonomy is intended precisely to disentangle such coupled design choices and expose the larger configuration space that arises from their composition.

\section{Mechanistic Motivation for Axis-Wise Diagnostics}
\label{app:axis_mechanisms}

This appendix provides axis-specific analyses that motivate the observable diagnostics used in our experiments. Each analysis characterizes a local mechanism by which the preferred action can vary with the denoising state. These results motivate the corresponding diagnostics, but do not imply that a single four-axis adaptive controller is universally optimal or reliable across tasks and models.

The opportunity in Eq.~\eqref{eq:state_opportunity} is only useful if we can understand what causes a preference to switch. Each axis exposes a different state-dependent trade-off: ranking reliability for score, positional stability for region, context provision versus reception for cardinality, and stable context versus revisability for commitment. We analyze one axis at a time while holding the others fixed. The goal is not to claim a universally optimal adaptive controller, but to characterize state variables whose changes can move the preferred action between regimes. Throughout, $Z$ denotes information available before the decision and $X_i$ the target token at position $i$.

\paragraph{Score: Uncertainty-Aware Aggregation.}
\label{para:score_dynamics}

The score axis asks which signals most reliably identify positions that are safe to reveal next. As denoising progresses, a rule that ranks positions well in one state may become less informative in another, while combining rules that make nearly identical rankings adds little. The preferred mixture should therefore depend on both current ranking reliability and complementary information, which we formalize through pairwise ranking evidence.

For positions $(i,j)$, let $Y_t^{ij}\in\{-1,+1\}$ indicate their preferred ordering under state--action utility. For scoring rule $r$, the signed difference $Y_t^{ij}(s_t^{(r)}(i)-s_t^{(r)}(j))$ is positive when the rule agrees with this ordering. Collecting these across $J$ rules gives
$\mathbf{x}_t^{ij}=Y_t^{ij}(s_t^{(1)}(i)-s_t^{(1)}(j),\ldots,s_t^{(J)}(i)-s_t^{(J)}(j))^\top$.
Let $\boldsymbol{\mu}_t=\mathbb{E}[\mathbf{x}_t^{ij}]$ denote average signed ranking evidence and $\boldsymbol{\Sigma}_t\succ0$ its sub-Gaussian covariance proxy. For score mixture $\boldsymbol{\alpha}$, define
$\Psi_t(\boldsymbol{\alpha})=\boldsymbol{\alpha}^{\top}\boldsymbol{\mu}_t/\sqrt{\boldsymbol{\alpha}^{\top}\boldsymbol{\Sigma}_t\boldsymbol{\alpha}}$,
which measures ranking signal relative to uncertainty.

\begin{theorem}[Correlated-Noise Score Aggregation]
\label{thm:score_aggregation}
If $\boldsymbol{\alpha}^{\top}\boldsymbol{\mu}_t>0$, the misranking probability is bounded as shown on the left. When the nonnegativity constraint is inactive, the optimal mixture is shown on the right.

\begin{minipage}[t]{0.48\linewidth}
\vspace{0pt}
\begin{equation}
\Pr[\boldsymbol{\alpha}^{\top}\mathbf{x}_t^{ij}\leq0]
\leq
\exp\!\left(-\frac{1}{2}\Psi_t(\boldsymbol{\alpha})^2\right).
\label{eq:score_misranking_bound}
\end{equation}
\end{minipage}
\hfill
\begin{minipage}[t]{0.48\linewidth}
\vspace{0pt}
\begin{equation}
\boldsymbol{\alpha}_t^\star
=
\frac{\boldsymbol{\Sigma}_t^{-1}\boldsymbol{\mu}_t}
{\mathbf{1}^{\top}\boldsymbol{\Sigma}_t^{-1}\boldsymbol{\mu}_t}.
\label{eq:score_closed_form}
\end{equation}
\end{minipage}
\end{theorem}

Theorem~\ref{thm:score_aggregation} shows that the preferred score depends on ranking reliability, captured by $\boldsymbol{\mu}_t$, and uncertainty and redundancy, captured by $\boldsymbol{\Sigma}_t$. As these change, so can the preferred mixture.

\begin{takeaway}
\textbf{Takeaway: }The best scoring rule depends on how reliably it ranks the positions and on how much unique information it provides. The optimal mixture rewards the reliable scores ($\boldsymbol{\mu}_t$) while also discounting the scores that make similar mistakes ($\boldsymbol{\Sigma}_t^{-1}$). Combining multiple score only help when they are not redundant and also accurate.
\end{takeaway}

\paragraph{Region: Robustness to Positional Uncertainty.}
\label{para:region_dynamics}

Once positions are scored, the next question is how much to trust where high-priority positions lie. If they remain spatially stable as new context arrives, a targeted region can exploit that structure; if they shift, the same restriction can exclude better candidates. We therefore model region selection through the spatial stability of the priority landscape.

To isolate region from cardinality, let $b_t$ be the number of masked positions in the candidate region and
$\mathcal{F}_{b_t}=\{\mathbf{r}\in\{0,1\}^{m_t}:\mathbf{1}^{\top}\mathbf{r}=b_t\}$.
Let $\mathcal{E}_t$ contain nearby masked-position pairs between which score may be redistributed:
\begin{equation}
\mathcal{U}_t(\nu_t)
=
\left\{
\mathbf{s}_t
+
\sum_{(u,v)\in\mathcal{E}_t}
\xi_{uv}(\mathbf{e}_u-\mathbf{e}_v)
:
|\xi_{uv}|\leq\nu_t
\right\},
\label{eq:region_uncertainty_set}
\end{equation}
where $\mathbf{e}_u$ is the unit vector for position $u$, $\xi_{uv}$ the signed score redistribution between $u$ and $v$, and $\nu_t$ the allowed positional fluctuation. Each redistribution preserves the sum of scores. We measure fragmentation by
$\mathrm{Cut}_t(\mathbf{r})=\sum_{(u,v)\in\mathcal{E}_t}|r_u-r_v|$,
where $r_u,r_v\in\{0,1\}$ indicate region membership, so $|r_u-r_v|=1$ only across a region boundary.

\begin{theorem}[Local-Transport Region Regimes]
\label{thm:region_regime}
The robust region satisfies
\begin{equation}
\arg\max_{\mathbf{r}\in\mathcal{F}_{b_t}}
\min_{\widetilde{\mathbf{s}}\in\mathcal{U}_t(\nu_t)}
\widetilde{\mathbf{s}}^\top\mathbf{r}
=
\arg\max_{\mathbf{r}\in\mathcal{F}_{b_t}}
\left\{
\mathbf{s}_t^\top\mathbf{r}
-
\nu_t\mathrm{Cut}_t(\mathbf{r})
\right\}.
\label{eq:robust_region_equivalent}
\end{equation}
Moreover, if $0\leq\nu_1<\nu_2$, corresponding optima $\mathbf{r}_1$ and $\mathbf{r}_2$ satisfy
$\mathrm{Cut}_t(\mathbf{r}_2)\leq\mathrm{Cut}_t(\mathbf{r}_1)$.
\end{theorem}

Theorem~\ref{thm:region_regime} shows how positional uncertainty changes the preferred region. When $\nu_t$ is small, selection mainly follows current score mass; as it grows, boundary penalties favor less fragmented regions that are more robust to local score fluctuations.

\begin{takeaway}
\textbf{Takeaway: }When token priorities are stable ($\nu_t \approx 0$), the decoder can target scattered positions with high scores. As positional uncertainty increases $(\nu_t\uparrow)$, the boundary penalty also increases. Therefore, it should prefer less fragmented regions that remain reliable when scores shift locally.
\end{takeaway}

\paragraph{Cardinality: Context Transfer Across Rounds.}
\label{para:cardinality_dynamics}

Cardinality asks how many positions to reveal now rather than later. Revealing more lets those positions provide context sooner, but revealing them too early prevents them from first receiving context from the current round. Recent information-theoretic analyses similarly connect generation order and parallel decoding quality to inter-token dependence and total correlation \citep{DBLP:journals/corr/abs-2602-00286,chen2025optimalinferenceschedulesmasked}.

Consider adjacent rounds with preceding context $Z$, current batch $\mathcal{B}$, next batch $\mathcal{B}'$, and a position $x$ whose reveal time may change. Let
$\mathsf{S}_{-}=(\mathcal{B},\mathcal{B}'\cup\{x\})$ postpone $x$, while
$\mathsf{S}_{+}=(\mathcal{B}\cup\{x\},\mathcal{B}')$ reveals it earlier.
For schedule $\mathsf{S}$, define
$\mathcal{L}(\mathsf{S})=D_{\mathrm{KL}}(P\Vert Q_{\mathsf{S}})$,
where $P$ is the target conditional distribution and $Q_{\mathsf{S}}$ the parallel factorization induced by $\mathsf{S}$.

\begin{theorem}[Context-Transfer Cardinality Regimes]
\label{thm:cardinality_regime}
For round-$h$ batch $\mathcal{B}_h$ with preceding context $Z_h$, the loss decomposes as shown on the left, where
$\operatorname{TC}(X_{\mathcal{B}_h}\mid Z_h)
=
D_{\mathrm{KL}}\!\left(
P(X_{\mathcal{B}_h}\mid Z_h)
\middle\Vert
\prod_{i\in\mathcal{B}_h}P(X_i\mid Z_h)
\right)$.
For the adjacent schedules above, their loss difference is shown on the right.

\begin{minipage}[t]{0.36\linewidth}
\vspace{0pt}
\begin{equation}
\mathcal{L}(\mathsf{S})
=
\sum_h
\operatorname{TC}
(X_{\mathcal{B}_h}\mid Z_h),
\label{eq:cardinality_tc_decomposition}
\end{equation}
\end{minipage}
\hfill
\begin{minipage}[t]{0.60\linewidth}
\vspace{0pt}
\begin{equation}
\begin{aligned}
\mathcal{L}(\mathsf{S}_{-})
-
\mathcal{L}(\mathsf{S}_{+})
&=
\sum_{j\in\mathcal{B}'}
I(X_j;X_x\mid Z,X_{\mathcal{B}})
\\
&\quad-
I(X_x;X_{\mathcal{B}}\mid Z).
\end{aligned}
\label{eq:cardinality_context_balance}
\end{equation}
\end{minipage}
\end{theorem}

In Eq.~\eqref{eq:cardinality_context_balance}, the first term measures how much revealing $x$ early helps the next batch, while the second measures how much $x$ gains by waiting. Thus, revealing more is beneficial when the context provided outweighs the benefit of waiting, and revealing fewer is preferable otherwise.

\begin{takeaway}
\textbf{Takeaway: }A token should be revealed early when the context it provides to the later tokens outweighs the context it would gain by waiting. That is, the preferred number of tokens to reveal depends on the direction and strength of inter-token dependencies.
\end{takeaway}

\paragraph{Commitment: Context Value versus Revision.}
\label{para:commitment_dynamics}

Commitment asks whether revealed predictions should remain fixed or revisable. Fixing a prediction provides stable context, but may become costly if later context would change that prediction.

For a revealed position $i$, let $\mathbf{p}_{t,i}$ be its current predictive distribution and
$a_i=\arg\max_y p_{t,i}(y)$ its most probable token. As more context arrives, let $\mathbf{q}$ be a possible future distribution, with $\rho\geq0$ bounding its deviation from $\mathbf{p}_{t,i}$. To quantify the benefit of revision, we measure how much another token can overtake $a_i$ and define
\[
V_{t,i}(\rho)
=
\max_{\mathbf{q}:\,
D_{\mathrm{KL}}(\mathbf{p}_{t,i}\Vert\mathbf{q})\leq\rho}
\log\!\left(\frac{\max_y q_y}{q_{a_i}}\right).
\]

The following theorem gives the threshold at which revision becomes valuable.

\begin{theorem}[Context--Revision Commitment Regimes]
\label{thm:commitment_regime}
Let $p_{(1)}$ and $p_{(2)}$ be the largest and second-largest probabilities in $\mathbf{p}_{t,i}$. The minimum posterior change required for the two to become equally probable is
\begin{equation}
\gamma_{t,i}
=
p_{(1)}
\log
\frac{2p_{(1)}}{p_{(1)}+p_{(2)}}
+
p_{(2)}
\log
\frac{2p_{(2)}}{p_{(1)}+p_{(2)}}.
\label{eq:commitment_flip_radius}
\end{equation}
Moreover, $V_{t,i}(\rho)=0$ if and only if $\rho\leq\gamma_{t,i}$, and
$V_{t,i}(\rho)$ is non-decreasing in $\rho$.
\end{theorem}

Theorem~\ref{thm:commitment_regime} shows that $\gamma_{t,i}$ is a robustness threshold: if $\rho\leq\gamma_{t,i}$, posterior change cannot overturn the prediction and revision has no value. Otherwise, we compare $V_{t,i}(\rho)$ with $\beta_{t,i}\geq0$, the downstream context value of keeping it fixed, committing when $\beta_{t,i}\geq V_{t,i}(\rho)$ and retaining revision otherwise.

\begin{takeaway}
\textbf{Takeaway: }A prediction should remain revisable only when future posterior changes can overturn its current top token. Predictions with a large top-two margin are more robust and can be committed more safely.
\end{takeaway}

Together, the four analyses identify candidate mechanisms for strategy reversal: ranking reliability and redundancy for score, positional stability for region, directed context transfer for cardinality, and posterior stability versus context value for commitment. These results are structural characterizations, not guarantees that a single observable proxy will be predictive across all tasks and models. Section~\ref{state_diagnostics} instantiates trajectory-level diagnostics for these quantities, and Section~\ref{opportunity_detection} separately asks whether the resulting local opportunities can be recognized on held-out states.


\section{Axis-Wise Diagnostic Action Proposals}
\label{app:axis_diagnostics}

The main paper treats action selection as an empirical component rather than a theorem-backed guarantee. For completeness, this appendix gives the exact training-free action proposals evaluated for each axis. In every experiment, only one axis is changed and all others remain fixed to the reference configuration.

For each axis, we instantiate one training-free proposal motivated by Appendix~\ref{app:axis_mechanisms}. When evaluating one axis, all other action components are held fixed to the reference configuration.

\paragraph{Score.}
Theorem~\ref{thm:score_aggregation} requires signed ranking evidence. We obtain it retrospectively from $(\mathbf S_{t-1},\delta_t)$: if position $i$ moves less than $j$ after new context arrives, $i$ would have been safer to prioritize at the preceding state. These comparisons yield a signed-evidence mean $\widehat{\boldsymbol\mu}_t$ and covariance $\widehat{\boldsymbol\Sigma}_t$, measuring respectively discriminative ranking evidence and its redundancy across scoring rules. We then choose
\begin{equation}
\hat{\boldsymbol{\alpha}}_t
\in
\arg\max_{\boldsymbol{\alpha}\in\Delta^{J-1}}
\frac{
\boldsymbol{\alpha}^{\top}\widehat{\boldsymbol{\mu}}_t
}{
\sqrt{
\boldsymbol{\alpha}^{\top}
\widehat{\boldsymbol{\Sigma}}_t
\boldsymbol{\alpha}
}
},
\label{eq:method_score}
\end{equation}
and form $\hat{\mathbf s}_t=\sum_j\hat\alpha_{t,j}\mathbf s_t^{(j)}$. Exact details are given in Appendix~\ref{app:score_diagnostic}.

\paragraph{Region.}
Using the selected mixture $\hat{\boldsymbol{\alpha}}_t$, we compare its current and preceding priority landscapes and estimate a transport radius $\widehat\nu_t$ that measures spatial priority drift while holding the score mixture fixed. Applying Theorem~\ref{thm:region_regime}, we select
\begin{equation}
\hat{\mathbf r}_t
\in
\arg\max_{\mathbf r\in\mathcal F_{b_t}}
\left\{
\hat{\mathbf s}_t^\top\mathbf r
-
\widehat\nu_t\,\mathrm{Cut}_t(\mathbf r)
\right\}.
\label{eq:method_region}
\end{equation}
The resulting support $\widehat{\mathcal R}_t=\{i:\hat r_{t,i}=1\}$ contains $b_t$ candidates. The budget is fixed independently of the diagnostic, so region determines \emph{where} candidates are drawn while cardinality determines how many are revealed. The transport estimator and dynamic program are given in Appendix~\ref{app:region_diagnostic}.

\paragraph{Cardinality.}
Theorem~\ref{thm:cardinality_regime} balances context provided by revealing a position against context forgone by revealing it too early. From source confidence, temporal response, and structural coupling, we construct the directed context-transfer proxy $\widehat D_t(i\!\rightarrow\!j)=\bar p_{t,i}\widehat\tau_{t,j}\phi_t(i\!\rightarrow\!j)$, where $\widehat\tau_{t,j}$ calibrates the trajectory-level sensitivity of target $j$. Ordering $\widehat{\mathcal R}_t$ by $\hat{\mathbf s}_t$, we use $\widehat D_t$ to estimate the marginal context balance $\widehat\Gamma_t$ of each additional candidate. With $k_t^{\min}=\lceil m_t/h_t\rceil$ denoting the minimum batch size required to finish within the remaining horizon, we select
\begin{equation}
\hat k_t
\in
\arg\max_{k_t^{\min}\leq k\leq b_t}
\widehat V_t(k),
\qquad
\hat\kappa_t=\frac{\hat k_t}{b_t},
\label{eq:method_cardinality}
\end{equation}
where $\widehat V_t(k)$ accumulates the marginal context-transfer balance beyond $k_t^{\min}$. Thus the candidate action satisfies the horizon constraint first and widens the reveal set only when the estimated context balance favors doing so. Definitions of $\widehat\tau_t$, $\widehat\Gamma_t$, and $\widehat V_t$ are given in Appendix~\ref{app:cardinality_diagnostic}.

\paragraph{Commitment.}
Commitment reuses $\widehat D_t$ to estimate the value of keeping a revealed prediction as persistent context. For each selected $i\in\mathcal N_t$, we denote this value by $\widehat\beta_{t,i}$ and estimate its historical posterior displacement $\widehat\rho_{t,i}$ from the recent trajectory. Relative to the decision-boundary radius $\gamma_{t,i}$ of Eq.~\eqref{eq:commitment_flip_radius}, the resulting revision pressure is $\widehat\eta_{t,i}=[\widehat\rho_{t,i}-\gamma_{t,i}]_+$. Following Theorem~\ref{thm:commitment_regime}, the diagnostic proposes commitment iff
\begin{equation}
\hat c_{t,i}
=
\mathbb I
\left[
\widehat\beta_{t,i}
\geq
\widehat\eta_{t,i}
\right].
\label{eq:method_commitment}
\end{equation}
Otherwise the position remains eligible for ReMDM's remasking operator~\citep{wang2025remasking}. Exact definitions are given in Appendix~\ref{app:commitment_diagnostic}.

\section{Implementation Details}
\label{app:implementation}

This appendix collects the solver and read-out details deferred from
Section~\ref{method}.

\paragraph{Score mixture (Eq.~\eqref{eq:method_score}).}
The objective is invariant to positive rescaling of $\boldsymbol{\alpha}$, so
maximizing it over $\Delta^{J-1}$ is equivalent to the convex program
\[
\min_{\boldsymbol{\alpha}\geq0}
\boldsymbol{\alpha}^{\top}\widehat{\boldsymbol{\Sigma}}_t\boldsymbol{\alpha}
\quad\text{subject to}\quad
\boldsymbol{\alpha}^{\top}\widehat{\boldsymbol{\mu}}_t=1,
\]
which is Eq.~\eqref{eq:app_score_qp} of
Appendix~\ref{app:proof_score_aggregation} with the ideal quantities replaced
by their estimates. When
$\widehat{\boldsymbol{\Sigma}}_t^{-1}\widehat{\boldsymbol{\mu}}_t$ is
entrywise positive the nonnegativity constraint is inactive and we use the
closed form
$\hat{\boldsymbol{\alpha}}_t\propto
\widehat{\boldsymbol{\Sigma}}_t^{-1}\widehat{\boldsymbol{\mu}}_t$;
otherwise we solve the $J$-dimensional quadratic program directly. Since $J$
is a small constant (the number of base scoring rules), both branches cost
$O(J^3)$.

We sample $|\Pi_t|$ position pairs uniformly without replacement from
$\mathcal{P}_t$ rather than enumerating all $\binom{|\mathcal{P}_t|}{2}$ of
them, which bounds the cost of forming
$(\widehat{\boldsymbol{\mu}}_t,\widehat{\boldsymbol{\Sigma}}_t)$ at
$O(|\Pi_t|J^2)$. The ridge term $\epsilon I$ keeps
$\widehat{\boldsymbol{\Sigma}}_t$ invertible and regularizes it when
$|\Pi_t|$ is comparable to $J$.

\paragraph{Region program (Eq.~\eqref{eq:method_region}).}
On the one-dimensional locality graph, $\mathrm{Cut}_t(\mathbf r)=
\sum_{i}w_i|r_{i+1}-r_i|$ is the weighted total variation of $\mathbf r$
along the sequence, so the objective decomposes over adjacent positions.
Scanning positions left to right with state (number of positions selected so
far, membership of the current position) and transition cost $w_i$ on each
membership change solves Eq.~\eqref{eq:method_region} exactly by dynamic
programming in $O(m_tb_t)$ time and $O(b_t)$ memory. No relaxation or
rounding is involved. For higher-dimensional or non-chain locality graphs the
cardinality-constrained problem is NP-hard in general, and a Lagrangian
relaxation of the budget constraint would be required.

\paragraph{Attention read-out.}
The structural coupling $\phi_t(i\!\rightarrow\!j)$ averages the attention
weight from query position $j$ to key position $i$ over a fixed set of layers
and heads, selected once and held constant across all experiments. These
weights are produced by the forward pass the decoder already performs, so no
additional evaluation of the denoiser is required; the remaining overhead
comes from retaining and reducing the attention maps.

\paragraph{Divergence directions.}
The two temporal divergences in Section~\ref{state_diagnostics} deliberately
run in opposite directions. For context transfer we use
$\delta_{t,i}=D_{\mathrm{KL}}(\mathbf p_{t,i}\Vert\mathbf p_{t-1,i})$, with
the new posterior first, because
$\mathbb{E}[D_{\mathrm{KL}}(p(x\mid y,z)\Vert p(x\mid z))]=I(X;Y\mid Z)$ is
exactly the identity that makes it a mutual-information proxy for
Eq.~\eqref{eq:cardinality_context_balance}. For commitment we use
$D_{\mathrm{KL}}(\mathbf p_{\tau-1,i}\Vert\mathbf p_{\tau,i})$, with the old
posterior first, because the KL ball defining $V_{t,i}(\rho)$ in
Theorem~\ref{thm:commitment_regime} is centered
at the current posterior and measures displacement away from it. Using a
single direction for both would misalign one of the two with its theorem.

\paragraph{Revision operator.}
Positions left revisable ($c_{t,i}=0$) are exposed to the remasking
operator of ReMDM~\citep{wang2025remasking}, which stochastically remasks
revealed tokens according to its schedule; a position with $c_{t,i}=1$ is
permanently excluded from remasking. The commitment diagnostic therefore
controls only \emph{eligibility} for revision; the revision dynamics follow
the underlying operator unchanged.

\paragraph{History window.}
$\mathcal H_{t,i}$ is the set of the most recent transitions at which
position $i$ was masked at both endpoints, truncated to a fixed window
length. Taking the maximum rather than the mean over the window makes
$\widehat\rho_{t,i}$ a conservative estimate of posterior movement, which
biases the decoder toward leaving predictions revisable.

\section{Selective Decoding Procedure}
\label{app:algorithm}

Algorithm~\ref{alg:state_adaptive_unmasking} summarizes the generic procedure used for a single transition-level axis. The remaining axes stay at their reference configuration. The axis diagnostic proposes an alternative action from the current trajectory, while the validation-calibrated opportunity detector decides whether that proposal should replace the best fixed action.

\begin{algorithm}[H]
\caption{Axis-Wise Selective State Adaptation}
\label{alg:state_adaptive_unmasking}
\begin{algorithmic}[1]
\Require Initial state $z_0$, horizon $T$, axis $j$, reference configuration $\pi_{\mathrm{ref}}$, validation-selected fixed action $a_{\mathrm{fix},j}^\star$, axis diagnostic $\mathcal D_j$, opportunity detector $\widehat g_j$, threshold $\tau$
\Ensure Generated sequence $X_T$

\State Initialize trajectory history
\For{$t=0,\ldots,T-1$}
    \State Run the MDM on $z_t$ and collect the trajectory statistics required by axis $j$
    \State Set all non-$j$ action components to their reference values
    \If{$t=0$ and the diagnostic requires temporal history}
        \State Set the axis-$j$ action to $a_{\mathrm{fix},j}^\star$
    \Else
        \State Compute the observable axis diagnostic $d_j(z_t)$
        \State Propose a state-dependent candidate action $\widehat a_j(z_t)=\mathcal D_j(z_t)$
        \State Estimate opportunity $\widehat g_j(z_t)$ from the validation-calibrated detector
        \If{$\widehat g_j(z_t)\geq\tau$}
            \State Use $\widehat a_j(z_t)$ on axis $j$
        \Else
            \State Use $a_{\mathrm{fix},j}^\star$ on axis $j$
        \EndIf
    \EndIf
    \State Apply the assembled action to obtain $z_{t+1}$
    \State Store current scores, posteriors, reveal decisions, and attention statistics
\EndFor
\State \Return $X_T$
\end{algorithmic}
\end{algorithm}

\section{Proofs}
\label{app:proof}

\subsection{Proof of Theorem~\ref{thm:strategy_reversal}}
\label{app:proof_strategy_reversal}

Let
$a_{\mathrm{fix}}^\star
\in
\arg\max_a
\mathbb{E}_{z\sim\mathcal D}[Q(z,a)]$
denote the best fixed action. By the definition of state-wise opportunity,
\[
g(z)
=
\max_a Q(z,a)
-
Q(z,a_{\mathrm{fix}}^\star)
\ge 0,
\]
and therefore
\begin{equation}
\Delta_{\mathrm{adapt}}
=
\mathbb{E}_{z\sim\mathcal D}[g(z)].
\label{eq:proof_adaptation_opportunity}
\end{equation}

Let $\mathcal R=\{z:g(z)>0\}$ denote the reversal set. Since
$g(z)=0$ for $z\notin\mathcal R$,
\begin{align}
\Delta_{\mathrm{adapt}}
&=
\mathbb{E}
\left[
g(z)\mathbf{1}\{z\in\mathcal R\}
\right]
\nonumber\\
&=
\Pr_{z\sim\mathcal D}(\mathcal R)\,
\mathbb{E}_{z\sim\mathcal D}
\left[
g(z)\mid z\in\mathcal R
\right].
\end{align}
If $\Pr(\mathcal R)=0$, then $g(z)=0$ almost surely and
$\Delta_{\mathrm{adapt}}=0$; otherwise the conditional expectation is well defined. This proves Eq.~\eqref{eq:frequency_margin_decomposition}.

For the lower bound, suppose that for some $b\in\mathcal A$ and $m,p>0$,
\[
\Pr_{z\sim\mathcal D}
\left[
Q(z,b)-Q(z,a_{\mathrm{fix}}^\star)\ge m
\right]
\ge p.
\]
On this event,
\[
g(z)
=
\max_a Q(z,a)-Q(z,a_{\mathrm{fix}}^\star)
\ge m.
\]
Hence,
\[
\Delta_{\mathrm{adapt}}
=
\mathbb{E}[g(z)]
\ge
m\,
\Pr_{z\sim\mathcal D}
\left[
Q(z,b)-Q(z,a_{\mathrm{fix}}^\star)\ge m
\right]
\ge pm,
\]
which proves the stated lower bound.

For the two-action case $\mathcal A=\{a,b\}$, define
$A(z)=Q(z,a)$, $B(z)=Q(z,b)$, and
$D(z)=A(z)-B(z)$. Using
$\max\{x,y\}=(x+y+|x-y|)/2$,
\[
\mathbb{E}[\max\{A,B\}]
=
\frac{
\mathbb{E}A+\mathbb{E}B+\mathbb{E}|D|
}{2}.
\]
Likewise, the best fixed utility is
\[
\max\{\mathbb{E}A,\mathbb{E}B\}
=
\frac{
\mathbb{E}A+\mathbb{E}B+|\mathbb{E}D|
}{2}.
\]
Subtracting gives
\[
\Delta_{a,b}
=
\frac{
\mathbb{E}|D|-|\mathbb{E}D|
}{2}.
\]
Substituting
$D(z)=D_{a,b}(z)=Q(z,a)-Q(z,b)$
proves Eq.~\eqref{eq:pairwise_adaptation_gap}.

Finally, suppose $\mathbb{E}[D]\ge0$, so that $a$ is the better fixed action. Then
\begin{align}
\Delta_{a,b}
&=
\frac{
\mathbb{E}|D|-\mathbb{E}D
}{2}
\nonumber\\
&=
\mathbb{E}[(-D)_+].
\end{align}
Since $(-D)_+=-D$ when $D<0$ and is zero otherwise,
\[
\Delta_{a,b}
=
\Pr[D<0]\,
\mathbb{E}[-D\mid D<0].
\]
Thus, in the pairwise case, adaptation value is exactly the frequency with which the globally inferior action becomes locally preferable times its average utility margin on those states. The case $\mathbb{E}[D]\le0$ follows symmetrically with $a$ and $b$ exchanged.
\hfill$\square$

\subsection{Proof of Theorem~\ref{thm:score_aggregation}}
\label{app:proof_score_aggregation}

We prove the theorem in four steps. We first derive the pairwise
misranking bound, then establish its distributionally robust
interpretation, characterize the optimal mixture, and finally derive the
incremental value of a new correlated scoring rule.

For readability, we suppress the time index $t$ and write
$\mathbf{x}=\mathbf{x}_t^{ij}$,
$\boldsymbol{\mu}=\boldsymbol{\mu}_t$, and
$\boldsymbol{\Sigma}=\boldsymbol{\Sigma}_t$.

\paragraph{Step 1: Pairwise misranking probability.}

By assumption, the centered signed evidence
$\mathbf{x}-\boldsymbol{\mu}$ is sub-Gaussian with covariance proxy
$\boldsymbol{\Sigma}$. Thus, for every
$\mathbf{u}\in\mathbb{R}^J$,
\begin{equation}
\log
\mathbb{E}
\left[
\exp
\left(
\mathbf{u}^{\top}
(\mathbf{x}-\boldsymbol{\mu})
\right)
\right]
\leq
\frac{1}{2}
\mathbf{u}^{\top}
\boldsymbol{\Sigma}
\mathbf{u}.
\label{eq:app_score_subgaussian}
\end{equation}

Fix any nonzero mixture $\boldsymbol{\alpha}\geq0$ such that
$\boldsymbol{\alpha}^{\top}\boldsymbol{\mu}>0$.
The aggregate misranks a quality-preferred pair precisely when
\[
\boldsymbol{\alpha}^{\top}\mathbf{x}\leq0.
\]
Equivalently,
\[
\boldsymbol{\alpha}^{\top}
(\mathbf{x}-\boldsymbol{\mu})
\leq
-
\boldsymbol{\alpha}^{\top}\boldsymbol{\mu}.
\]

For any $\lambda>0$, Markov's inequality gives
\begin{align}
\Pr
\left[
\boldsymbol{\alpha}^{\top}\mathbf{x}\leq0
\right]
&=
\Pr
\left[
-\lambda
\boldsymbol{\alpha}^{\top}
(\mathbf{x}-\boldsymbol{\mu})
\geq
\lambda
\boldsymbol{\alpha}^{\top}\boldsymbol{\mu}
\right]
\\
&\leq
\exp
\left(
-\lambda
\boldsymbol{\alpha}^{\top}\boldsymbol{\mu}
\right)
\mathbb{E}
\left[
\exp
\left(
-\lambda
\boldsymbol{\alpha}^{\top}
(\mathbf{x}-\boldsymbol{\mu})
\right)
\right].
\end{align}

Applying Eq.~\eqref{eq:app_score_subgaussian} with
$\mathbf{u}=-\lambda\boldsymbol{\alpha}$ yields
\[
\Pr
\left[
\boldsymbol{\alpha}^{\top}\mathbf{x}\leq0
\right]
\leq
\exp
\left[
-\lambda
\boldsymbol{\alpha}^{\top}\boldsymbol{\mu}
+
\frac{\lambda^2}{2}
\boldsymbol{\alpha}^{\top}
\boldsymbol{\Sigma}
\boldsymbol{\alpha}
\right].
\]

The right-hand side is minimized at
\[
\lambda^\star
=
\frac{
\boldsymbol{\alpha}^{\top}\boldsymbol{\mu}
}{
\boldsymbol{\alpha}^{\top}
\boldsymbol{\Sigma}
\boldsymbol{\alpha}
}.
\]
Substituting $\lambda^\star$ gives
\[
\Pr
\left[
\boldsymbol{\alpha}^{\top}\mathbf{x}\leq0
\right]
\leq
\exp
\left(
-
\frac{
(\boldsymbol{\alpha}^{\top}\boldsymbol{\mu})^2
}{
2
\boldsymbol{\alpha}^{\top}
\boldsymbol{\Sigma}
\boldsymbol{\alpha}
}
\right).
\]

Using the definition of
$\Psi(\boldsymbol{\alpha})$ proves
Eq.~\eqref{eq:score_misranking_bound}.

\paragraph{Step 2: Robust interpretation of the ranking certificate.}

Consider an ellipsoidal uncertainty set around the mean signed evidence,
\[
\mathcal{E}(\rho)
=
\left\{
\widetilde{\boldsymbol{\mu}}
=
\boldsymbol{\mu}+\mathbf{u}
:
\mathbf{u}^{\top}
\boldsymbol{\Sigma}^{-1}
\mathbf{u}
\leq
\rho^2
\right\}.
\]

For a fixed mixture $\boldsymbol{\alpha}$, its worst-case expected signed
margin is
\begin{align}
\min_{\widetilde{\boldsymbol{\mu}}\in\mathcal{E}(\rho)}
\boldsymbol{\alpha}^{\top}
\widetilde{\boldsymbol{\mu}}
&=
\boldsymbol{\alpha}^{\top}\boldsymbol{\mu}
+
\min_{
\mathbf{u}^{\top}
\boldsymbol{\Sigma}^{-1}\mathbf{u}
\leq\rho^2
}
\boldsymbol{\alpha}^{\top}\mathbf{u}.
\end{align}

Set
$\mathbf{v}=\boldsymbol{\Sigma}^{-1/2}\mathbf{u}$.
Then $\|\mathbf{v}\|_2\leq\rho$ and
\[
\boldsymbol{\alpha}^{\top}\mathbf{u}
=
(\boldsymbol{\Sigma}^{1/2}\boldsymbol{\alpha})^\top
\mathbf{v}.
\]
By Cauchy--Schwarz,
\[
\min_{\|\mathbf{v}\|_2\leq\rho}
(\boldsymbol{\Sigma}^{1/2}\boldsymbol{\alpha})^\top
\mathbf{v}
=
-
\rho
\sqrt{
\boldsymbol{\alpha}^{\top}
\boldsymbol{\Sigma}
\boldsymbol{\alpha}
}.
\]

Therefore,
\begin{equation}
\min_{\widetilde{\boldsymbol{\mu}}\in\mathcal{E}(\rho)}
\boldsymbol{\alpha}^{\top}
\widetilde{\boldsymbol{\mu}}
=
\boldsymbol{\alpha}^{\top}\boldsymbol{\mu}
-
\rho
\sqrt{
\boldsymbol{\alpha}^{\top}
\boldsymbol{\Sigma}
\boldsymbol{\alpha}
}.
\label{eq:app_score_robust_margin}
\end{equation}

This worst-case margin remains positive exactly when
\[
\rho
<
\frac{
\boldsymbol{\alpha}^{\top}\boldsymbol{\mu}
}{
\sqrt{
\boldsymbol{\alpha}^{\top}
\boldsymbol{\Sigma}
\boldsymbol{\alpha}
}
}
=
\Psi(\boldsymbol{\alpha}).
\]

Hence, $\Psi(\boldsymbol{\alpha})$ is precisely the largest
ellipsoidal perturbation radius under which the expected quality-aligned
ranking margin remains positive.

\paragraph{Step 3: Optimal score aggregation.}

Because
$\Psi(\boldsymbol{\alpha})$
is invariant to positive rescaling of $\boldsymbol{\alpha}$, maximizing it
over nonnegative nonzero vectors is equivalent to the convex problem
\begin{equation}
\min_{\boldsymbol{\alpha}\geq0}
\boldsymbol{\alpha}^{\top}
\boldsymbol{\Sigma}
\boldsymbol{\alpha}
\qquad
\text{subject to}
\qquad
\boldsymbol{\alpha}^{\top}\boldsymbol{\mu}=1.
\label{eq:app_score_qp}
\end{equation}

We first consider the case in which the nonnegativity constraint is
inactive. By generalized Cauchy--Schwarz,
\begin{align}
\left(
\boldsymbol{\alpha}^{\top}\boldsymbol{\mu}
\right)^2
&=
\left[
(\boldsymbol{\Sigma}^{1/2}\boldsymbol{\alpha})^\top
(\boldsymbol{\Sigma}^{-1/2}\boldsymbol{\mu})
\right]^2
\\
&\leq
\left(
\boldsymbol{\alpha}^{\top}
\boldsymbol{\Sigma}
\boldsymbol{\alpha}
\right)
\left(
\boldsymbol{\mu}^{\top}
\boldsymbol{\Sigma}^{-1}
\boldsymbol{\mu}
\right).
\end{align}

Consequently,
\[
\Psi(\boldsymbol{\alpha})^2
\leq
\boldsymbol{\mu}^{\top}
\boldsymbol{\Sigma}^{-1}
\boldsymbol{\mu}.
\]

Equality in Cauchy--Schwarz holds if and only if
\[
\boldsymbol{\Sigma}^{1/2}\boldsymbol{\alpha}
\propto
\boldsymbol{\Sigma}^{-1/2}\boldsymbol{\mu},
\]
or equivalently,
\[
\boldsymbol{\alpha}
\propto
\boldsymbol{\Sigma}^{-1}\boldsymbol{\mu}.
\]

Therefore, whenever
$\boldsymbol{\Sigma}^{-1}\boldsymbol{\mu}$
is entrywise positive, the nonnegativity constraint is inactive and the
simplex-normalized solution is
\[
\boldsymbol{\alpha}^\star
=
\frac{
\boldsymbol{\Sigma}^{-1}\boldsymbol{\mu}
}{
\mathbf{1}^{\top}
\boldsymbol{\Sigma}^{-1}\boldsymbol{\mu}
}.
\]

The corresponding optimal squared ranking certificate is
\begin{equation}
\mathcal{R}^{\star 2}
=
\boldsymbol{\mu}^{\top}
\boldsymbol{\Sigma}^{-1}
\boldsymbol{\mu}.
\label{eq:app_score_optimal_certificate}
\end{equation}

If some entries of
$\boldsymbol{\Sigma}^{-1}\boldsymbol{\mu}$
are non-positive, the optimum lies on a face of the nonnegative orthant.
In that case, Eq.~\eqref{eq:app_score_qp} remains a convex quadratic
program, and the same argument applies to the active subset of scoring
rules selected by its KKT conditions.

\paragraph{Step 4 (supplementary): incremental value of a correlated scoring
rule.}

This step is not part of the statement of
Theorem~\ref{thm:score_aggregation}; it quantifies what a candidate scoring
rule adds to an existing family.

Consider an existing set of scoring rules with mean evidence
$\boldsymbol{\mu}$ and covariance proxy $\boldsymbol{\Sigma}$.
Add a new scoring rule with mean signed evidence $\mu_0$, uncertainty
$\sigma_0^2$, and covariance vector $\mathbf{c}$ with the existing rules.
The augmented quantities are
\[
\widetilde{\boldsymbol{\mu}}
=
\begin{bmatrix}
\boldsymbol{\mu}\\
\mu_0
\end{bmatrix},
\qquad
\widetilde{\boldsymbol{\Sigma}}
=
\begin{bmatrix}
\boldsymbol{\Sigma} & \mathbf{c}\\
\mathbf{c}^{\top} & \sigma_0^2
\end{bmatrix}.
\]

Assume the augmented matrix is positive definite and that both the original
and augmented optima are interior. Define the Schur complement
\[
S
=
\sigma_0^2
-
\mathbf{c}^{\top}
\boldsymbol{\Sigma}^{-1}
\mathbf{c}
>0.
\]

The block inverse formula gives
\[
\widetilde{\boldsymbol{\Sigma}}^{-1}
=
\begin{bmatrix}
\boldsymbol{\Sigma}^{-1}
+
\boldsymbol{\Sigma}^{-1}
\mathbf{c}
S^{-1}
\mathbf{c}^{\top}
\boldsymbol{\Sigma}^{-1}
&
-
\boldsymbol{\Sigma}^{-1}\mathbf{c}S^{-1}
\\
-
S^{-1}\mathbf{c}^{\top}\boldsymbol{\Sigma}^{-1}
&
S^{-1}
\end{bmatrix}.
\]

Using Eq.~\eqref{eq:app_score_optimal_certificate},
the augmented optimal squared certificate is
\[
\widetilde{\mathcal{R}}^{\star 2}
=
\widetilde{\boldsymbol{\mu}}^{\top}
\widetilde{\boldsymbol{\Sigma}}^{-1}
\widetilde{\boldsymbol{\mu}}.
\]

Expanding the quadratic form yields
\begin{align}
\widetilde{\mathcal{R}}^{\star 2}
&=
\boldsymbol{\mu}^{\top}
\boldsymbol{\Sigma}^{-1}
\boldsymbol{\mu}
+
\frac{
\left(
\mu_0
-
\mathbf{c}^{\top}
\boldsymbol{\Sigma}^{-1}
\boldsymbol{\mu}
\right)^2
}{
\sigma_0^2
-
\mathbf{c}^{\top}
\boldsymbol{\Sigma}^{-1}
\mathbf{c}
}.
\end{align}

Therefore,
\[
\widetilde{\mathcal{R}}^{\star 2}
-
\mathcal{R}^{\star 2}
=
\frac{
\left(
\mu_0
-
\mathbf{c}^{\top}
\boldsymbol{\Sigma}^{-1}
\boldsymbol{\mu}
\right)^2
}{
\sigma_0^2
-
\mathbf{c}^{\top}
\boldsymbol{\Sigma}^{-1}
\mathbf{c}
}.
\]

The numerator is the squared residual quality-aligned discrimination of
the new score after accounting for the existing scores, while the
denominator is its residual uncertainty after the same projection.
Thus, a correlated scoring rule contributes only to the extent that it
contains discriminative evidence not already represented by the existing
score family.

This completes the proof.
\qed

\subsection{Proof of Theorem~\ref{thm:region_regime}}
\label{app:proof_region_regime}

We prove the robust reformulation and the monotonicity of optimal boundary
complexity, and then characterize the resulting finite regime structure.

For clarity, we suppress the time index $t$ and write
$\mathbf{s}=\mathbf{s}_t$, $m=m_t$, $b=b_t$,
$\mathcal{E}=\mathcal{E}_t$, and $\mathbf{B}=\mathbf{B}_t$.
The feasible region family is
\[
\mathcal{F}_{b}
=
\left\{
\mathbf{r}\in\{0,1\}^{m}
:
\mathbf{1}^{\top}\mathbf{r}=b
\right\}.
\]
For each edge $e=(u,v)\in\mathcal{E}$, let $\mathbf{b}_e$ denote the
corresponding column of an oriented incidence matrix $\mathbf{B}$, with
orientation chosen arbitrarily. Then
$\mathbf{b}_e^\top\mathbf{r}=r_u-r_v$ up to sign, and hence
\[
\left|\mathbf{b}_e^\top\mathbf{r}\right|
=
|r_u-r_v|.
\]
The boundary complexity can therefore be written as
\[
\mathrm{Cut}(\mathbf{r})
=
\sum_{e\in\mathcal{E}}
\left|
\mathbf{b}_e^\top\mathbf{r}
\right|.
\]

\paragraph{Step 1: Local transport preserves total score.}

Every column of an incidence matrix contains one $+1$ and one $-1$, so
\[
\mathbf{1}^{\top}\mathbf{B}
=
\mathbf{0}^{\top}.
\]
Therefore, for any admissible transport vector $\boldsymbol{\xi}$,
\[
\mathbf{1}^{\top}
(\mathbf{s}+\mathbf{B}\boldsymbol{\xi})
=
\mathbf{1}^{\top}\mathbf{s}.
\]
Thus, the uncertainty set
\[
\mathcal{U}(\nu)
=
\left\{
\mathbf{s}+\mathbf{B}\boldsymbol{\xi}
:
|\xi_e|\leq\nu,\ \forall e\in\mathcal{E}
\right\}
\]
redistributes score across locally connected positions without changing
the total score.

\paragraph{Step 2: Exact robust reformulation.}

Fix any feasible region $\mathbf{r}\in\mathcal{F}_{b}$. Its worst-case
score under local redistribution is
\begin{align}
\min_{\widetilde{\mathbf{s}}\in\mathcal{U}(\nu)}
\widetilde{\mathbf{s}}^\top\mathbf{r}
&=
\min_{|\xi_e|\leq\nu}
(\mathbf{s}+\mathbf{B}\boldsymbol{\xi})^\top\mathbf{r}
\\
&=
\mathbf{s}^\top\mathbf{r}
+
\min_{|\xi_e|\leq\nu}
\boldsymbol{\xi}^\top\mathbf{B}^\top\mathbf{r}.
\end{align}
Because the constraints on $\boldsymbol{\xi}$ are independent across
edges, the inner problem separates:
\begin{align}
\min_{|\xi_e|\leq\nu}
\boldsymbol{\xi}^\top\mathbf{B}^\top\mathbf{r}
&=
\sum_{e\in\mathcal{E}}
\min_{|\xi_e|\leq\nu}
\xi_e\,\mathbf{b}_e^\top\mathbf{r}
\\
&=
-\nu
\sum_{e\in\mathcal{E}}
\left|
\mathbf{b}_e^\top\mathbf{r}
\right|
\\
&=
-\nu\,\mathrm{Cut}(\mathbf{r}),
\end{align}
where the second equality follows from the scalar identity
\[
\min_{|\xi|\leq\rho}\xi a
=
-\rho|a|.
\]
Hence,
\begin{equation}
\min_{\widetilde{\mathbf{s}}\in\mathcal{U}(\nu)}
\widetilde{\mathbf{s}}^\top\mathbf{r}
=
\mathbf{s}^\top\mathbf{r}
-
\nu\,\mathrm{Cut}(\mathbf{r}).
\label{eq:app_region_robust_value}
\end{equation}

Maximizing Eq.~\eqref{eq:app_region_robust_value} over
$\mathbf{r}\in\mathcal{F}_{b}$ gives
\[
\arg\max_{\mathbf{r}\in\mathcal{F}_{b}}
\min_{\widetilde{\mathbf{s}}\in\mathcal{U}(\nu)}
\widetilde{\mathbf{s}}^\top\mathbf{r}
=
\arg\max_{\mathbf{r}\in\mathcal{F}_{b}}
\left\{
\mathbf{s}^\top\mathbf{r}
-
\nu\,\mathrm{Cut}(\mathbf{r})
\right\},
\]
which proves Eq.~\eqref{eq:robust_region_equivalent}.

\paragraph{Step 3: Boundary complexity decreases with positional uncertainty.}

Let $0\leq\nu_1<\nu_2$, and choose arbitrary optimal regions
\[
\mathbf{r}_1
\in
\arg\max_{\mathbf{r}\in\mathcal{F}_{b}}
\left\{
\mathbf{s}^\top\mathbf{r}
-
\nu_1\mathrm{Cut}(\mathbf{r})
\right\}
\]
and
\[
\mathbf{r}_2
\in
\arg\max_{\mathbf{r}\in\mathcal{F}_{b}}
\left\{
\mathbf{s}^\top\mathbf{r}
-
\nu_2\mathrm{Cut}(\mathbf{r})
\right\}.
\]

Optimality of $\mathbf{r}_1$ at $\nu_1$ gives
\begin{equation}
\mathbf{s}^\top\mathbf{r}_1
-
\nu_1\mathrm{Cut}(\mathbf{r}_1)
\geq
\mathbf{s}^\top\mathbf{r}_2
-
\nu_1\mathrm{Cut}(\mathbf{r}_2),
\label{eq:app_region_opt_nu1}
\end{equation}
while optimality of $\mathbf{r}_2$ at $\nu_2$ gives
\begin{equation}
\mathbf{s}^\top\mathbf{r}_2
-
\nu_2\mathrm{Cut}(\mathbf{r}_2)
\geq
\mathbf{s}^\top\mathbf{r}_1
-
\nu_2\mathrm{Cut}(\mathbf{r}_1).
\label{eq:app_region_opt_nu2}
\end{equation}
Adding Eqs.~\eqref{eq:app_region_opt_nu1} and
\eqref{eq:app_region_opt_nu2} yields
\[
(\nu_2-\nu_1)
\left[
\mathrm{Cut}(\mathbf{r}_1)
-
\mathrm{Cut}(\mathbf{r}_2)
\right]
\geq0.
\]
Since $\nu_2>\nu_1$,
\[
\mathrm{Cut}(\mathbf{r}_2)
\leq
\mathrm{Cut}(\mathbf{r}_1),
\]
which proves the second statement of
Theorem~\ref{thm:region_regime}.
\qed

\paragraph{Additional regime structure.}

The same formulation also shows that the preferred region changes through
a finite set of regimes as positional uncertainty varies. For each
$\mathbf{r}\in\mathcal{F}_{b}$, define its score mass
$S(\mathbf{r})=\mathbf{s}^{\top}\mathbf{r}$ and the affine objective
\[
L_{\mathbf{r}}(\nu)
=
S(\mathbf{r})
-
\nu\,\mathrm{Cut}(\mathbf{r}).
\]
Since
$|\mathcal{F}_{b}|=\binom{m}{b}<\infty$, the optimal robust value is
\begin{equation}
V(\nu)
=
\max_{\mathbf{r}\in\mathcal{F}_{b}}
L_{\mathbf{r}}(\nu),
\label{eq:app_region_upper_envelope}
\end{equation}
the pointwise maximum of finitely many affine functions. Consequently,
$V(\nu)$ is convex and piecewise linear.

For two regions $\mathbf{r}$ and $\mathbf{r}'$ with different boundary
complexities, their objectives coincide only when
\[
S(\mathbf{r})
-
\nu\,\mathrm{Cut}(\mathbf{r})
=
S(\mathbf{r}')
-
\nu\,\mathrm{Cut}(\mathbf{r}'),
\]
which occurs at
\begin{equation}
\nu^\dagger(\mathbf{r},\mathbf{r}')
=
\frac{
S(\mathbf{r})-S(\mathbf{r}')
}{
\mathrm{Cut}(\mathbf{r})-\mathrm{Cut}(\mathbf{r}')
}.
\label{eq:app_region_switching_threshold}
\end{equation}
There are only finitely many pairs of feasible regions, and therefore
finitely many candidate switching thresholds. Between consecutive
thresholds, the ordering of the affine objectives cannot change.
Accordingly, whenever the optimizer is unique, the optimal region remains
constant throughout the interval. If multiple regions induce identical
optimal objectives, they may remain tied over an interval, but the set of
optimal objective regimes still changes only at finitely many thresholds.

This gives a direct interpretation of each transition. Suppose the
preferred region changes from $\mathbf{r}_1$ to a less fragmented region
$\mathbf{r}_2$ at a threshold $\nu^\dagger$, with
$\mathrm{Cut}(\mathbf{r}_2)<\mathrm{Cut}(\mathbf{r}_1)$. At the switching
point,
\[
S(\mathbf{r}_1)
-
\nu^\dagger\mathrm{Cut}(\mathbf{r}_1)
=
S(\mathbf{r}_2)
-
\nu^\dagger\mathrm{Cut}(\mathbf{r}_2),
\]
and therefore
\begin{equation}
\nu^\dagger
=
\frac{
S(\mathbf{r}_1)-S(\mathbf{r}_2)
}{
\mathrm{Cut}(\mathbf{r}_1)-\mathrm{Cut}(\mathbf{r}_2)
}.
\label{eq:app_region_switch_interpretation}
\end{equation}
Thus, a more fragmented region is preferred while its additional score
mass compensates for its greater boundary exposure; once positional
uncertainty becomes sufficiently large, a less fragmented region becomes
preferable.

\subsection{Proof of Theorem~\ref{thm:cardinality_regime}}
\label{app:proof_cardinality_regime}

We prove the theorem in three steps. We first establish the exact
decomposition of batch-factorization loss into conditional total
correlations, then derive the quality effect of moving one position across
a batch boundary, and finally obtain the cardinality expansion criterion.

Throughout the proof, $Z$ denotes the information available before the
first batch under consideration.

\paragraph{Step 1: Batch-factorization loss.}

Consider a batch schedule
\[
\mathsf{S}
=
(\mathcal{B}_1,\ldots,\mathcal{B}_H)
\]
over a collection of unresolved positions $U$, where the batches form a
partition of $U$. Let
\[
Z_h
=
(Z,X_{B_1},\ldots,X_{B_{h-1}})
\]
denote the information available before batch $\mathcal{B}_h$.

The reference joint conditional distribution admits the chain-rule
factorization
\[
P(X_U\mid Z)
=
\prod_{h=1}^{H}
P(X_{\mathcal{B}_h}\mid Z_h).
\]

A batch-parallel decoder instead factorizes the predictions within each
batch:
\[
Q_{\mathsf{S}}(X_U\mid Z)
=
\prod_{h=1}^{H}
\prod_{i\in \mathcal{B}_h}
P(X_i\mid Z_h).
\]

Therefore,
\begin{align}
\mathcal{L}(\mathsf{S})
&=
D_{\mathrm{KL}}
\left(
P(X_U\mid Z)
\Vert
Q_{\mathsf{S}}(X_U\mid Z)
\right)
\\
&=
\mathbb{E}_{P}
\left[
\log
\frac{
P(X_U\mid Z)
}{
Q_{\mathsf{S}}(X_U\mid Z)
}
\right]
\\
&=
\sum_{h=1}^{H}
\mathbb{E}_{P}
\left[
\log
\frac{
P(X_{\mathcal{B}_h}\mid Z_h)
}{
\prod_{i\in \mathcal{B}_h}P(X_i\mid Z_h)
}
\right].
\end{align}

By the definition of conditional total correlation,
\[
\operatorname{TC}(X_{\mathcal{B}_h}\mid Z_h)
=
D_{\mathrm{KL}}
\left(
P(X_{\mathcal{B}_h}\mid Z_h)
\middle\|
\prod_{i\in \mathcal{B}_h}P(X_i\mid Z_h)
\right),
\]
where the conditional divergence is averaged over $Z_h$.

Hence,
\[
\mathcal{L}(\mathsf{S})
=
\sum_{h=1}^{H}
\operatorname{TC}(X_{\mathcal{B}_h}\mid Z_h),
\]
which proves Eq.~\eqref{eq:cardinality_tc_decomposition}.

\paragraph{Step 2: Moving a position across a batch boundary.}

We now compare
\[
\mathsf{S}_{-}
=
(\mathcal{B},\{x\}\cup\mathcal{B}')
\]
with
\[
\mathsf{S}_{+}
=
(\mathcal{B}\cup\{x\},\mathcal{B}').
\]

The two schedules reveal exactly the same collection of positions after two
rounds. Therefore, all later terms in the factorization-loss decomposition
are identical and cancel. It suffices to compare the two rounds shown
above.

From Step~1,
\begin{align}
\mathcal{L}(\mathsf{S}_{-})
&=
\operatorname{TC}(X_{\mathcal{B}}\mid Z)
+
\operatorname{TC}
\left(
X_{\{x\}\cup\mathcal{B}'}
\mid
Z,X_{\mathcal{B}}
\right)
+
\mathrm{const},
\\
\mathcal{L}(\mathsf{S}_{+})
&=
\operatorname{TC}
\left(
X_{\mathcal{B}\cup\{x\}}
\mid
Z
\right)
+
\operatorname{TC}
\left(
X_{\mathcal{B}'}
\mid
Z,X_{\mathcal{B}},X_x
\right)
+
\mathrm{const}.
\end{align}

We first consider the change in the current batch. By definition,
\begin{align}
&
\operatorname{TC}
\left(
X_{\mathcal{B}\cup\{x\}}\mid Z
\right)
-
\operatorname{TC}(X_{\mathcal{B}}\mid Z)
\\
&=
H(X_x\mid Z)
-
H(X_x\mid Z,X_{\mathcal{B}})
\\
&=
I(X_x;X_{\mathcal{B}}\mid Z).
\label{eq:app_cardinality_incoming}
\end{align}

Thus, placing $x$ in the same batch as $\mathcal{B}$ incurs exactly the conditional
mutual information between $x$ and the current batch. This is the
information that $x$ could have gained by waiting one additional round.

Next, consider the following batch. Using the chain rule
\[
H(X_x,X_{\mathcal{B}'}\mid Z,X_{\mathcal{B}})
=
H(X_x\mid Z,X_{\mathcal{B}})
+
H(X_{\mathcal{B}'}\mid Z,X_{\mathcal{B}},X_x),
\]
we obtain
\begin{align}
&
\operatorname{TC}
\left(
X_{\{x\}\cup\mathcal{B}'}
\mid
Z,X_{\mathcal{B}}
\right)
-
\operatorname{TC}
\left(
X_{\mathcal{B}'}
\mid
Z,X_{\mathcal{B}},X_x
\right)
\\
&=
\sum_{j\in\mathcal{B}'}
\left[
H(X_j\mid Z,X_{\mathcal{B}})
-
H(X_j\mid Z,X_{\mathcal{B}},X_x)
\right]
\\
&=
\sum_{j\in\mathcal{B}'}
I(X_j;X_x\mid Z,X_{\mathcal{B}}).
\label{eq:app_cardinality_outgoing}
\end{align}

Hence, moving $x$ into the earlier batch reduces the next-round
factorization loss by exactly the aggregate information that $x$ provides
to the downstream positions in $\mathcal{B}'$.

\paragraph{Step 3: Exact cardinality trade-off.}

Combining
Eqs.~\eqref{eq:app_cardinality_incoming} and
\eqref{eq:app_cardinality_outgoing} gives
\begin{align}
\mathcal{L}(\mathsf{S}_{+})
-
\mathcal{L}(\mathsf{S}_{-})
&=
I(X_x;X_{\mathcal{B}}\mid Z)
-
\sum_{j\in\mathcal{B}'}
I(X_j;X_x\mid Z,X_{\mathcal{B}}).
\end{align}

Define
\[
P(x;\mathcal{B}\mid Z)
=
I(X_x;X_{\mathcal{B}}\mid Z)
\]
as the premature-revelation cost and
\[
G(x;\mathcal{B}'\mid Z,\mathcal{B})
=
\sum_{j\in\mathcal{B}'}
I(X_j;X_x\mid Z,X_{\mathcal{B}})
\]
as the downstream context gain.

Then
\[
\mathcal{L}(\mathsf{S}_{+})
-
\mathcal{L}(\mathsf{S}_{-})
=
P(x;\mathcal{B}\mid Z)
-
G(x;\mathcal{B}'\mid Z,\mathcal{B}).
\]

Since smaller KL factorization loss corresponds to higher local
log-quality,
\[
\mathcal{L}(\mathsf{S}_{+})
<
\mathcal{L}(\mathsf{S}_{-})
\]
if and only if
\[
G(x;\mathcal{B}'\mid Z,\mathcal{B})
>
P(x;\mathcal{B}\mid Z).
\]

Thus, increasing the current cardinality by advancing $x$ is
quality-improving exactly when the downstream context value supplied by
$x$ exceeds the information that $x$ itself would gain by waiting.

This proves Eq.~\eqref{eq:cardinality_context_balance}.
\qed

\subsection{Proof of Theorem~\ref{thm:commitment_regime}}
\label{app:proof_commitment_regime}

We prove the theorem in four steps. We first derive the exact KL distance
to a MAP decision boundary, then identify the closest competing token,
characterize the revision option value, and finally establish the
commitment threshold and its comparative statics.

For readability, we suppress the indices $t$ and $i$ throughout the proof.
Let
\[
\mathbf{p}
=
(p_1,\ldots,p_{|\mathcal V|})
\]
be the current predictive distribution, and let
\[
a
=
\arg\max_y p_y
\]
be its unique MAP prediction. Since model softmax probabilities are
strictly positive, we assume $p_y>0$ for all $y$.

\paragraph{Step 1: KL distance to the decision boundary against a fixed competitor.}

Fix any competing token $b\neq a$. Consider the optimization problem
\begin{equation}
\gamma_b
=
\min_{\mathbf{q}\in\Delta^{|\mathcal V|-1}}
D_{\mathrm{KL}}(\mathbf{p}\Vert\mathbf{q})
\quad
\text{subject to}
\quad
q_b\geq q_a.
\label{eq:app_commitment_pairwise_flip}
\end{equation}

Because $p_a>p_b$, the original distribution $\mathbf p$ does not satisfy
$q_b\geq q_a$. The feasible set is convex and the KL divergence
$D_{\mathrm{KL}}(\mathbf p\Vert\mathbf q)$ is strictly convex in
$\mathbf q$ over the positive simplex. Hence the minimizer is unique.

At the minimizer, the inequality constraint must be active:
\[
q_a=q_b.
\]
Otherwise, if $q_b>q_a$, one could move sufficiently small probability
mass from $b$ toward $a$, moving $\mathbf q$ toward $\mathbf p$ while
remaining feasible and strictly decreasing the KL divergence.

We therefore solve
\[
\min_{\mathbf q}
D_{\mathrm{KL}}(\mathbf p\Vert\mathbf q)
\quad\text{subject to}\quad
q_a=q_b,
\qquad
\sum_y q_y=1.
\]

Since the terms
$\sum_y p_y\log p_y$ do not depend on $\mathbf q$, this is equivalent to
minimizing
\[
-\sum_y p_y\log q_y.
\]
Introduce multipliers $\lambda$ and $\eta$ for the normalization and
equality constraints. The Lagrangian is
\[
\mathcal L(\mathbf q,\lambda,\eta)
=
-\sum_y p_y\log q_y
+
\lambda
\left(
\sum_y q_y-1
\right)
+
\eta(q_a-q_b).
\]

For $y\notin\{a,b\}$, stationarity gives
\[
-\frac{p_y}{q_y}+\lambda=0,
\]
so
\[
q_y=\frac{p_y}{\lambda}.
\]
For $a$ and $b$,
\[
q_a=\frac{p_a}{\lambda+\eta},
\qquad
q_b=\frac{p_b}{\lambda-\eta}.
\]

The constraint $q_a=q_b$ together with normalization yields
\[
q_a^\star
=
q_b^\star
=
\frac{p_a+p_b}{2},
\]
while
\[
q_y^\star=p_y
\qquad
\text{for }y\notin\{a,b\}.
\]

Thus, reaching the closest $a$--$b$ decision boundary only redistributes
probability mass between the current winner and the competing token; all
other probabilities remain unchanged.

Substituting $\mathbf q^\star$ into the KL divergence gives
\begin{equation}
\gamma_b
=
p_a
\log
\frac{2p_a}{p_a+p_b}
+
p_b
\log
\frac{2p_b}{p_a+p_b}.
\label{eq:app_commitment_pairwise_radius}
\end{equation}

\paragraph{Step 2: The nearest competitor is the runner-up.}

It remains to determine which $b\neq a$ gives the smallest value of
$\gamma_b$. For fixed $p_a$, define
\[
f(x)
=
p_a
\log
\frac{2p_a}{p_a+x}
+
x
\log
\frac{2x}{p_a+x},
\qquad
0<x<p_a.
\]

Differentiating with respect to $x$ gives
\[
f'(x)
=
\log
\frac{2x}{p_a+x}.
\]
Since $x<p_a$,
\[
\frac{2x}{p_a+x}<1,
\]
and therefore
\[
f'(x)<0.
\]

Hence, the KL distance to the pairwise decision boundary decreases
monotonically as the probability of the competing token increases.
The closest competing token is therefore the token with the largest
probability among all $b\neq a$, namely the runner-up.

Writing
\[
p_{(1)}=p_a,
\qquad
p_{(2)}=\max_{b\neq a}p_b,
\]
we obtain
\[
\gamma
=
p_{(1)}
\log
\frac{2p_{(1)}}{p_{(1)}+p_{(2)}}
+
p_{(2)}
\log
\frac{2p_{(2)}}{p_{(1)}+p_{(2)}}.
\]

This proves Eq.~\eqref{eq:commitment_flip_radius}.

\paragraph{Step 3: Characterization of the revision option value.}

Recall
\[
V(\rho)
=
\max_{
\mathbf q:
D_{\mathrm{KL}}(\mathbf p\Vert\mathbf q)\leq\rho
}
\log\frac{\max_yq_y}{q_a}.
\]

We first consider $\rho\leq\gamma$.
By definition, $\gamma$ is the minimum KL divergence required to reach
any distribution satisfying
\[
q_y\geq q_a
\]
for some $y\neq a$.

Therefore, for every
$\mathbf q$ satisfying
\[
D_{\mathrm{KL}}(\mathbf p\Vert\mathbf q)<\gamma,
\]
$a$ remains the unique MAP prediction. Hence,
\[
\log\bigl(\max_y q_y/q_a\bigr)=0.
\]

At $\rho=\gamma$, the uncertainty set first touches a MAP decision
boundary, where the current token and its nearest competitor tie. The
quantity $\log(\max_yq_y/q_a)$ remains zero at this boundary. Consequently,
\[
V(\rho)=0
\qquad
\text{for all }
\rho\leq\gamma.
\]

Now consider $\rho>\gamma$. Let $\mathbf q^\star$ be the nearest boundary
distribution derived in Step~1 for the runner-up token. At
$\mathbf q^\star$,
\[
q^\star_{(2)}=q^\star_a.
\]
Because the KL divergence is continuous in the interior of the simplex,
for any $\rho>\gamma$ there exists a sufficiently small perturbation of
$\mathbf q^\star$ that transfers mass from $a$ to the runner-up such that
\[
q_{(2)}>q_a
\]
while still satisfying
\[
D_{\mathrm{KL}}(\mathbf p\Vert\mathbf q)<\rho.
\]
For this distribution,
\[
\log\bigl(\max_yq_y/q_a\bigr)>0.
\]
Hence,
\[
V(\rho)>0
\qquad
\text{for every }\rho>\gamma.
\]

Combining the two directions gives
\[
V(\rho)=0
\quad\Longleftrightarrow\quad
\rho\leq\gamma.
\]

To establish monotonicity, note that for
$\rho_1\leq\rho_2$,
\[
\mathcal Q(\rho_1)
\subseteq
\mathcal Q(\rho_2).
\]
Maximizing the same objective over nested feasible sets yields
\[
V(\rho_1)
\leq
V(\rho_2).
\]
Thus, the revision option value is non-decreasing in posterior uncertainty.

\paragraph{Step 4: Context--revision switching threshold.}

Under the local robust commitment criterion, committing position $i$
provides context value $\beta$ but forfeits at most $V(\rho)$ in revision
value. Its conservative net quality advantage is therefore
\[
G(\rho)
=
\beta-V(\rho).
\]

Commitment is preferred whenever
\[
G(\rho)\geq0,
\]
equivalently,
\[
V(\rho)\leq \beta.
\]

Because $V(\rho)$ is non-decreasing, the set
\[
\left\{
\rho:
V(\rho)\leq \beta
\right\}
\]
is an interval beginning at $\rho=0$. Its upper endpoint is
\[
\rho^\star(\beta)
=
\sup
\left\{
\rho:
V(\rho)\leq \beta
\right\}.
\]

Thus, posterior uncertainty below the threshold lies in the commitment
regime, whereas uncertainty above the threshold lies in the revision
regime, up to possible indifference at the boundary.

Finally, consider two context values
\[
0\leq \beta_1\leq \beta_2.
\]
Then
\[
\{\rho:V(\rho)\leq \beta_1\}
\subseteq
\{\rho:V(\rho)\leq \beta_2\}.
\]
Taking suprema gives
\[
\rho^\star(\beta_1)
\leq
\rho^\star(\beta_2).
\]

Hence, the switching threshold is non-decreasing in context value:
predictions that provide greater downstream context value can tolerate
larger posterior uncertainty before revision becomes preferable.

This proves Theorem~\ref{thm:commitment_regime}.
\qed

\paragraph{Relation to the top-two margin.}

Let
\[
s=p_{(1)}+p_{(2)}
\]
and
\[
\delta
=
\frac{p_{(1)}-p_{(2)}}{p_{(1)}+p_{(2)}}.
\]
Then
\[
p_{(1)}
=
\frac{s}{2}(1+\delta),
\qquad
p_{(2)}
=
\frac{s}{2}(1-\delta).
\]
Substituting these expressions into
Eq.~\eqref{eq:commitment_flip_radius} yields
\[
\gamma
=
\frac{s}{2}
\left[
(1+\delta)\log(1+\delta)
+
(1-\delta)\log(1-\delta)
\right].
\]
Differentiating with respect to $\delta$ while holding $s$ fixed gives
\[
\frac{\partial\gamma}{\partial\delta}
=
\frac{s}{2}
\log
\frac{1+\delta}{1-\delta}.
\]
For every unique MAP prediction, $\delta>0$, and therefore
\[
\frac{\partial\gamma}{\partial\delta}>0.
\]
Thus, at fixed top-two probability mass, a larger prediction margin
strictly increases the KL distance to the nearest MAP reversal.


\section{Training-Free Diagnostics and Optimization Details}
\label{app:Training-Free Diagnostics}

This appendix details how the trajectory signals in Section~\ref{state_diagnostics} are transformed into the axis-specific diagnostics used in Appendix~\ref{app:axis_diagnostics}. All estimators use only states already visited by the decoding trajectory and statistics available from the denoiser forward pass.

\subsection{Score: Retrospective Ranking Evidence}
\label{app:score_diagnostic}

Theorem~\ref{thm:score_aggregation} characterizes score aggregation through the mean and covariance of signed pairwise evidence. Because the true ordering $Y_t^{ij}$ is unavailable at inference time, we construct a retrospective label from posterior response. For $i,j\in\mathcal P_t$, we set $\widehat Y_t^{ij}=\operatorname{sign}(\delta_{t,j}-\delta_{t,i})$: a position whose posterior moved less after new context arrived is treated as the safer position to have prioritized at the preceding state.

For pairs $(i,j)\in\Pi_t$ sampled from $\mathcal P_t$, the corresponding signed evidence is
\begin{equation}
\widehat{\mathbf x}_t^{ij}
=
\widehat Y_t^{ij}
\begin{pmatrix}
s_{t-1}^{(1)}(i)-s_{t-1}^{(1)}(j)\\
\vdots\\
s_{t-1}^{(J)}(i)-s_{t-1}^{(J)}(j)
\end{pmatrix}.
\label{eq:app_signed_evidence}
\end{equation}

We estimate
\begin{align}
\widehat{\boldsymbol\mu}_t
&=
\frac{1}{|\Pi_t|}
\sum_{(i,j)\in\Pi_t}
\widehat{\mathbf x}_t^{ij},
\\
\widehat{\boldsymbol\Sigma}_t
&=
\operatorname{Cov}
\left(
\left\{
\widehat{\mathbf x}_t^{ij}
\right\}_{(i,j)\in\Pi_t}
\right)
+
\epsilon I .
\end{align}
The mean measures whether each scoring rule agrees with the ordering later supported by the trajectory, while the covariance captures correlated ranking uncertainty and redundancy among scoring rules.

\paragraph{Optimization.} The objective in Eq.~\eqref{eq:method_score} is scale invariant. If $\mathbf v_t=\widehat{\boldsymbol\Sigma}_t^{-1}\widehat{\boldsymbol\mu}_t$ is entrywise positive, its simplex solution is
\begin{equation}
\hat{\boldsymbol\alpha}_t
=
\frac{\mathbf v_t}
{\mathbf 1^\top\mathbf v_t}.
\label{eq:app_score_closed_form}
\end{equation}
Otherwise we solve the simplex-constrained problem numerically. Since $J$ is small, this optimization is negligible relative to a denoiser forward pass.

\subsection{Region: Priority Drift}
\label{app:region_diagnostic}

Region selection should react to changes in the spatial organization of priority rather than to changes in the score mixture itself. We therefore apply the \emph{current} mixture $\hat{\boldsymbol\alpha}_t$ to both the current and preceding score landscapes over $\mathcal P_t$:
\[
\tilde{\mathbf s}_{t}
=
\sum_j\hat\alpha_{t,j}\mathbf s_t^{(j)},
\qquad
\tilde{\mathbf s}_{t-1}
=
\sum_j\hat\alpha_{t,j}\mathbf s_{t-1}^{(j)}.
\]
After centering both landscapes over $\mathcal P_t$, let $\mathbf d_t=\tilde{\mathbf s}_{t}-\tilde{\mathbf s}_{t-1}$ denote their difference.

For the one-dimensional locality graph used in our experiments, we estimate the transport radius as
\begin{equation}
\widehat\nu_t
=
\max_{1\leq j<|\mathcal P_t|}
\frac{
\left|
\sum_{i\leq j}d_{t,i}
\right|
}{
w_j
},
\label{eq:app_transport_radius}
\end{equation}
where $w_j$ is the weight of the corresponding locality boundary. Large $\widehat\nu_t$ indicates that priority has moved across spatial boundaries, increasing the cost of fragmented candidate supports in Eq.~\eqref{eq:method_region}.

\paragraph{Dynamic program.} On a one-dimensional locality graph, $\mathrm{Cut}_t(\mathbf r)$ decomposes over adjacent boundaries. Equation~\eqref{eq:method_region} can therefore be solved exactly by a dynamic program whose state records the processed prefix, the number of selected positions, and whether the current boundary position is selected, yielding $O(m_tb_t)$ time and $O(b_t)$ memory with rolling states.

\subsection{Cardinality: Directed Context Transfer}
\label{app:cardinality_diagnostic}

Theorem~\ref{thm:cardinality_regime} compares the context a position can provide after being revealed with the context it would receive by remaining unresolved. We estimate this interaction from source reliability, target sensitivity, and source--target structural coupling.

Let $\bar p_{t,i}=\max_y p_{t,i}(y)$ denote the reliability of source $i$, and let $\phi_t(i\!\rightarrow\!j)$ denote its attention-derived coupling to target $j$. We calibrate the sensitivity of target $j$ using the preceding transition:
\begin{equation}
\widehat\tau_{t,j}
=
\frac{
\delta_{t,j}
}{
\epsilon+
\sum_{r\in\mathcal N_{t-1}}
\bar p_{t-1,r}\,
\phi_{t-1}(r\!\rightarrow\!j)
}.
\label{eq:app_target_sensitivity}
\end{equation}
The directed context-transfer proxy is then
\begin{equation}
\widehat D_t(i\!\rightarrow\!j)
=
\bar p_{t,i}\,
\widehat\tau_{t,j}\,
\phi_t(i\!\rightarrow\!j).
\label{eq:app_context_transfer}
\end{equation}

The numerator of Eq.~\eqref{eq:app_target_sensitivity} measures how strongly $j$ actually responded to newly available context, while the denominator normalizes that response by the estimated reliable coupling supplied on the preceding transition. Equation~\eqref{eq:app_context_transfer} transfers this calibrated sensitivity to the current source and dependency structure. It is a trajectory-calibrated proxy for context transfer, not a causal effect or an exact mutual-information estimator.

Let the candidate support returned by region selection be ordered by $\hat{\mathbf s}_t$ as $i_1,\ldots,i_{b_t}$, with $\mathcal N_k=\{i_1,\ldots,i_k\}$. For the next candidate $x=i_{k+1}$, revealing it now supplies estimated context
\begin{equation}
\widehat G_t(x\mid\mathcal N_k)
=
\sum_{j\in\widehat{\mathcal R}_t\setminus(\mathcal N_k\cup\{x\})}
\widehat D_t(x\!\rightarrow\!j),
\label{eq:app_card_gain}
\end{equation}
while revealing it in parallel forgoes context that it could have received from the current batch,
\begin{equation}
\widehat P_t(x\mid\mathcal N_k)
=
\sum_{r\in\mathcal N_k}
\widehat D_t(r\!\rightarrow\!x).
\label{eq:app_card_penalty}
\end{equation}
Their difference
\begin{equation}
\widehat\Gamma_t(x\mid\mathcal N_k)
=
\widehat G_t(x\mid\mathcal N_k)
-
\widehat P_t(x\mid\mathcal N_k)
\label{eq:app_card_balance}
\end{equation}
is the empirical counterpart of the context-provision versus premature-revelation balance in Theorem~\ref{thm:cardinality_regime}.

If $m_t$ positions and $h_t$ denoising rounds remain, completion within the horizon requires $k_t^{\min}=\lceil m_t/h_t\rceil$, with $b_t\geq k_t^{\min}$. The cumulative value used in Eq.~\eqref{eq:method_cardinality} is
\begin{equation}
\widehat V_t(k)
=
\sum_{\ell=k_t^{\min}+1}^{k}
\widehat\Gamma_t
\left(
i_\ell
\mid
\mathcal N_{\ell-1}
\right),
\label{eq:app_card_value}
\end{equation}
with the empty sum defined as zero. The decoder therefore reveals the minimum batch required by the horizon and expands it only when the cumulative estimated context balance is favorable.

\subsection{Commitment: Persistent Context and Revision Pressure}
\label{app:commitment_diagnostic}

Theorem~\ref{thm:commitment_regime} compares the value of retaining a revealed token as persistent context against the value of preserving its revisability. We reuse the directed context-transfer estimate from Eq.~\eqref{eq:app_context_transfer} rather than introducing a separate dependency proxy.

Let $\mathcal N_t=\mathcal N_{\hat k_t}$ be the positions selected for revelation. For $i\in\mathcal N_t$, its value as persistent context for unresolved positions is
\begin{equation}
\widehat\beta_{t,i}
=
\sum_{j\in\mathcal M_t\setminus\mathcal N_t}
\widehat D_t(i\!\rightarrow\!j).
\label{eq:app_commitment_value}
\end{equation}

We estimate historical revisability from recent posterior movement:
\begin{equation}
\widehat\rho_{t,i}
=
\max_{\tau\in\mathcal H_{t,i}}
D_{\mathrm{KL}}
\left(
\mathbf p_{\tau-1,i}
\Vert
\mathbf p_{\tau,i}
\right),
\label{eq:app_revision_radius}
\end{equation}
where $\mathcal H_{t,i}$ is a recent history window over states in which the posterior of $i$ is available.

The temporal direction in Eq.~\eqref{eq:app_revision_radius} intentionally differs from $\delta_{t,i}=D_{\mathrm{KL}}(\mathbf p_{t,i}\Vert\mathbf p_{t-1,i})$. The latter measures the response of the current posterior to newly revealed context, whereas the former treats the earlier posterior as the reference point of the historical commitment ball.

Revision becomes decision-relevant only after posterior movement reaches the decision-boundary radius $\gamma_{t,i}$ from Eq.~\eqref{eq:commitment_flip_radius}. We therefore define
\begin{equation}
\widehat\eta_{t,i}
=
\left[
\widehat\rho_{t,i}
-
\gamma_{t,i}
\right]_+.
\label{eq:app_revision_pressure}
\end{equation}
Equation~\eqref{eq:method_commitment} commits a prediction when its estimated persistent context value $\widehat\beta_{t,i}$ exceeds its revision pressure $\widehat\eta_{t,i}$; otherwise it remains eligible for ReMDM's remasking operator~\citep{wang2025remasking}.

\subsection{Validation-Calibrated Opportunity Detector}
\label{app:opportunity_detector}

The opportunity detector is intentionally lightweight because the purpose of the experiment is to test whether the low-cost observable diagnostic contains information about \emph{when} adaptation matters, rather than to fit a high-capacity predictor of rollout utility. For each model--task--axis setting, only validation prompts are used for calibration. Let $d_j(z)$ be the observable diagnostic for axis $j$ and let $g_j(z)$ be the axis-wise candidate-set opportunity defined in Section~\ref{state_wise_opportunity}. We partition the empirical validation distribution of $d_j$ into quantile bins and store the mean $g_j$ in each bin. Held-out states are mapped to the corresponding validation bin, producing $\widehat g_j(z)$ without observing any held-out branch utility.

We evaluate the detector in two complementary ways. AUROC for the binary event $g_j(z)>0$ measures whether the detector separates states with any positive candidate-set opportunity, while Spearman correlation measures ranking quality with respect to continuous opportunity magnitude. Because selective adaptation ultimately depends on how much utility-relevant opportunity is captured under a limited activation budget, we additionally report opportunity capture across coverage levels using Eq.~\eqref{eq:captured_opportunity}. A detector can therefore achieve reasonable positive-state classification yet capture little opportunity if it ranks small positive gains above rarer, larger ones; this distinction appears empirically in Section~\ref{exp:predict}.

The detector calibration does not update denoiser parameters and adds no denoiser forward pass. At inference, it requires only computing the already-defined axis diagnostic and a bin lookup. For end-to-end selective decoding, the activation threshold is fixed from validation data before held-out generation.

\subsection{Computational Overhead}
\label{app:method_complexity}

Let $|\Pi_t|$ be the number of sampled position pairs, $J$ the number of base scoring rules, $m_t$ the number of currently masked positions, and $b_t$ the candidate-support budget. Computing the ranking statistics and score mixture costs $O(|\Pi_t|J^2+J^3)$, region selection costs $O(m_tb_t)$, and evaluating context-transfer balances within the candidate support costs at most $O(b_t^2)$. The total per-step overhead is therefore $O(|\Pi_t|J^2+J^3)+O(m_tb_t)+O(b_t^2)$ in addition to attention statistics produced by the denoiser forward pass, with no additional denoiser evaluation. Opportunity calibration adds only a constant-time bin lookup after the axis diagnostic has been computed.

\section{Experimental Details}
\label{app:experimental_details}

\subsection{Models and Checkpoints}
\label{app:model_checkpoints}

We evaluate three masked diffusion language models. For reproducibility, we use the following public checkpoints:
\begin{itemize}
    \item LLaDA-8B-Instruct~\citep{nie2025large}: \texttt{GSAI-ML/LLaDA-8B-Instruct}.
    \item LLaDA-1.5~\citep{zhu2025llada}: \texttt{GSAI-ML/LLaDA-1.5}.
    \item Dream-7B~\citep{ye2025dream}: \texttt{Dream-org/Dream-v0-Instruct-7B}.
\end{itemize}

All model-specific decoding settings are held fixed within each model unless explicitly varied as part of the candidate action under evaluation.

\subsection{Task Suite}
\label{app:task_suite}

We use a fixed suite of ten tasks spanning code generation, mathematical
reasoning, structured generation, constrained infilling, sparse masking, and
commitment-sensitive generation. The suite is fixed before inspecting any
selective-adaptation or end-to-end result. Table~\ref{tab:task_details}
summarizes the task source, generation horizon, output format, and terminal
utility.

\begin{table}[t]
\centering
\caption{\textbf{Task-suite details.} ``Partial'' indicates component- or
field-level credit in $[0,1]$; ``binary'' indicates exact task success. The
generation horizon is fixed within each task across candidate actions.}
\label{tab:task_details}
\scriptsize
\setlength{\tabcolsep}{3.0pt}
\begin{tabular}{lclll}
\toprule
Task & Horizon & Source & Output & Utility \\
\midrule
CSV Missing Cells       & 16  & constructed & one cell          & binary \\
HumanEval                & 256 & official    & Python completion & test pass \\
Multi-Span Cloze         & 64  & constructed & indexed spans     & partial \\
GSM8K                    & 256 & official    & numeric answer    & binary \\
Carry RTL                & 32  & constructed & three carry bits  & partial \\
TD07 Sparse Mask         & 48  & constructed & three spans        & partial \\
JSON Mode Eval           & 128 & public      & schema JSON        & field credit \\
Constrained JSON Fill    & 96  & constructed & constrained JSON  & partial \\
Unique List Commit       & 48  & constructed & ordered list       & partial \\
HTML Close Tags          & 48  & constructed & closing-tag suffix & exact/partial \\
\bottomrule
\end{tabular}
\end{table}

All constructed tasks are generated deterministically and stored as JSONL with disjoint prompt IDs. GSM8K and HumanEval use their official datasets, while JSON Mode Eval uses the public \texttt{NousResearch/json-mode-eval} collection. For the transition-level analyses, we use 20 validation and 80 held-out prompts per task, except JSON Mode Eval with 20 validation and 48 held-out prompts. Each prompt contributes eight evaluated states. Table~\ref{tab:app-e1-full} uses the full pool of 800 states (544 for JSON Mode Eval), whereas the held-out analyses use 640 states (384 for JSON Mode Eval).

For the end-to-end evaluation, we use disjoint final-test prompts: 500 newly generated prompts per constructed task, 500 GSM8K test examples (indices 200--699), 64 remaining HumanEval problems, and 20 remaining JSON Mode Eval examples. Validation prompts are used only for policy selection and calibration, with all decisions frozen before held-out and final-test evaluation.

\subsubsection{CSV Missing Cells}

This task tests local reconstruction under an exact structured-text format.
Each prompt contains a CSV header and rows with exactly one cell replaced by
\texttt{???}. The model must return only the missing value. The evaluator
normalizes surrounding whitespace and compares the extracted answer with the
gold cell.

\paragraph{Illustrative example.}
\begin{quote}
\small
\texttt{id,name,qty}\\
\texttt{1,nails,10}\\
\texttt{2,screws,???}\\[2pt]
Target: \texttt{25}
\end{quote}

The task has a generation horizon of 16 tokens. Since there is one missing
cell, utility is binary.

\subsubsection{HumanEval}

HumanEval evaluates functional code generation using the official programming
problems. The prompt provides a Python function signature and natural-language
specification, and the model generates the missing implementation. The
generated completion is normalized, combined with the provided function
context, and executed in a restricted sandbox against the official hidden
tests.

\paragraph{Schematic example.}
\begin{quote}
\small
\texttt{def has\_close\_elements(numbers, threshold):}\\
\hspace*{1em}\texttt{"""Return whether two elements are closer than threshold."""}\\
\hspace*{1em}\texttt{\# model generates the function body}
\end{quote}

Utility is 1 if all tests pass and 0 for an incorrect completion, runtime
error, or timeout. Evaluator failures that are not attributable to generated
code are recorded separately as unavailable rather than incorrect. We use a
256-token generation horizon.

\subsubsection{Multi-Span Cloze}

Multi-Span Cloze contains sentences or short passages with two or three
non-contiguous blanks. Blanks can be linked by entity identity, arithmetic,
ordering, temporal, or lexical constraints. Repeated blank indices refer to
the same answer. The model returns indexed assignments separated by
semicolons.

\paragraph{Illustrative example.}
\begin{quote}
\small
Prompt: \texttt{Once the [1] opened, the [2] walked in. The [2] had been
waiting for the [1] since dawn.}\\
Target: \texttt{[1]=gate; [2]=courier}
\end{quote}

The generation horizon is 64 tokens. Exact completion receives utility 1;
otherwise utility is the fraction of indexed spans that exactly match after
case and whitespace normalization.

\subsubsection{GSM8K}

GSM8K evaluates grade-school mathematical reasoning using the official test
problems. The model may produce intermediate reasoning, but evaluation extracts
the final numeric answer and compares it with the official target.

\paragraph{Schematic example.}
\begin{quote}
\small
Prompt: \texttt{A box contains 12 pencils. After adding 8 pencils and giving
away 5, how many remain?}\\
Gold final answer: \texttt{15}
\end{quote}

Utility is binary final-answer correctness. Explanatory text does not affect
the score as long as the final numeric answer is extracted correctly. The
generation horizon is 256 tokens.

\subsubsection{Carry RTL}

Carry RTL isolates right-to-left dependencies in multi-digit addition. The
prompt gives two three-digit integers and asks only for carry bits, ordered
from the ones column to the hundreds column. The sum itself must not be
returned. Examples are selected to contain at least two nonzero carries.

\paragraph{Illustrative example.}
\begin{quote}
\small
Prompt: \texttt{589 + 673}\\
Target: \texttt{[1]=1; [2]=1; [3]=1}
\end{quote}

Here \texttt{[1]} is the carry from ones to tens, \texttt{[2]} from tens to
hundreds, and \texttt{[3]} from hundreds to thousands. The generation horizon
is 32 tokens. Utility is the fraction of correctly predicted carry digits.

\subsubsection{TD07 Sparse Mask}

TD07 Sparse Mask measures reconstruction when missing words are spatially
separated. For each source sentence, three distant positions are replaced by
indexed blanks, typically near the beginning, middle, and end. This preserves
the sentence and mask count while making the missing evidence non-local.

\paragraph{Illustrative example.}
\begin{quote}
\small
Prompt: \texttt{The [1] cooled loaves on [2] rack beside [3] oven.}\\
Target: \texttt{[1]=baker; [2]=a; [3]=the}
\end{quote}

The generation horizon is 48 tokens. Utility is the fraction of the three
missing words recovered exactly after normalization. The corresponding
clustered-mask variant is used only as a controlled diagnostic task and is not
a member of the fixed ten-task suite.

\subsubsection{JSON Mode Eval}

JSON Mode Eval contains natural-language requests paired with JSON schemas and
gold JSON objects. The prompt includes the schema, required fields, and a user
request. The model must return one JSON object satisfying the schema.

\paragraph{Illustrative example.}
\begin{quote}
\small
Schema fields:
\texttt{ssid}, \texttt{securityProtocol}, \texttt{bandwidth}.\\
Request: configure SSID \texttt{OfficeNetSecure}, security
\texttt{WPA2-Enterprise}, and bandwidth \texttt{1300 Mbps}.\\
Target:
\texttt{\{"ssid":"OfficeNetSecure",}\\
\texttt{"securityProtocol":"WPA2-Enterprise",}\\
\texttt{"bandwidth":"1300 Mbps"\}}
\end{quote}

The prediction is parsed as JSON and validated against the provided schema.
An exact object match receives utility 1. A valid but non-exact object receives
the fraction of gold top-level fields whose values match exactly; invalid JSON
receives 0. The generation horizon is 128 tokens.

\subsubsection{Constrained JSON Fill}

Constrained JSON Fill uses synthetic JSON records with explicit cross-field
constraints. Examples include schedules where
\texttt{end = start + duration} and travel records where
\texttt{total = nights * price\_per\_night}. The model must output the complete
JSON object rather than only the derived field.

\paragraph{Illustrative example.}
\begin{quote}
\small
Input:
\texttt{event=seminar; start=18; duration=2;}\\
\texttt{allowed\_rooms=['B','C']; room=C}\\
Target:
\texttt{\{"event":"seminar","start":18,"end":20,"room":"C"\}}
\end{quote}

The generation horizon is 96 tokens. JSON key order is not scored. An exact
semantically equivalent object receives utility 1; configured component-level
credit is used when a parseable response recovers only part of the target.

\subsubsection{Unique List Commit}

Unique List Commit tests whether the decoder can maintain non-duplication and
ordering constraints over a multi-item output. Each prompt provides a pool of
12 tokens and asks for eight distinct items in alphabetical order. Repeated
items, incorrect ordering, or unsupported items are errors.

\paragraph{Illustrative example.}
\begin{quote}
\small
Input:
\texttt{cat,dog,elk,fox,gnu,hen,ibis,jay,kiwi,lynx,mole,newt}\\
Target:
\texttt{cat,dog,elk,fox,gnu,hen,ibis,jay}
\end{quote}

The generation horizon is 48 tokens. Exact completion receives utility 1;
otherwise component credit is the fraction of target list positions matched
exactly. This task is commitment-sensitive because an early duplicate or
ordering error constrains the validity of later items.

\subsubsection{HTML Close Tags}

HTML Close Tags presents a prefix containing nested opening tags and asks for
only the missing closing-tag sequence. Correct generation requires respecting
last-opened, first-closed nesting order.

\paragraph{Illustrative example.}
\begin{quote}
\small
Prompt prefix: \texttt{<article><p><em>}\\
Target: \texttt{</em></p></article>}
\end{quote}

The generation horizon is 48 tokens. Exact balanced closure receives utility
1. When the output can be decomposed into the expected components, partial
credit reflects the fraction of closing tags in their correct positions.

\subsection{Output Extraction and Terminal Utility}
\label{app:terminal_utility}

For constructed tasks, prompts request that the final answer appear inside
\texttt{<answer>...</answer>}; a boxed answer or the last non-empty output
line is used as a fallback. Whitespace is normalized before scoring, and
case-insensitive comparison is used for ordinary textual fields. JSON tasks
are compared after parsing whenever possible.

Let $u(y,e)\in[0,1]$ denote the terminal utility of model output $y$ for
example $e$. For binary tasks, $u=1$ exactly when the task-specific evaluator
accepts the output. For decomposable structured tasks with $K$ target parts,
we use
\begin{equation}
u(y,e)
=
\frac{1}{K}
\sum_{k=1}^{K}
\mathbbm{1}
\left[
\operatorname{norm}(\hat y_k)
=
\operatorname{norm}(y_k^\star)
\right].
\label{eq:part_credit}
\end{equation}
The same task-specific terminal utility is used to evaluate every branch
continuation from a state, so candidate actions differ only in the state
transition and not in the downstream evaluator.

\subsection{State Sampling and Utility Evaluation}
\label{app:evaluation_protocol}

For each task, we use up to 100 prompts from the evaluation split, except JSON
Mode Eval, which has 68 evaluation prompts. We sample eight spaced denoising
states per prompt. For every candidate state--action pair, we estimate
one-step state--action utility using four continuation rollouts under the
frozen model-specific continuation policy. Candidate actions at the same
state share rollout seeds to reduce comparison variance.

Candidate-oracle estimates are cross-fitted across rollout folds. We select
the maximizing candidate using rollouts $\{0,1\}$ and evaluate that candidate
using rollouts $\{2,3\}$, then reverse the two folds and average the estimates.
The same split-sample construction is used for E1 gaps, E2 oracle regret,
opportunity labels, oracle concentration, and oracle-gate controls. This
prevents the same rollout noise from both selecting and evaluating the
maximizing action. In the present temperature-zero branch data, all 84,758
audited state--axis comparisons selected the same maximizing utility under the
naive and cross-fitted calculations, so cross-fitting did not change the
reported point estimates.

All tasks use temperature zero in the reported E1--E3 experiments. The
intervention horizon is one transition: the candidate action is applied once,
after which generation continues under the fixed continuation policy.
Generation prompts, targets, candidate sets, and split membership remain fixed
throughout the experiment.

Validation prompts are used only to select the best fixed action, fit the
stage-only schedule, calibrate the action-diagnostic threshold, and fit the
opportunity detector. Held-out prompts are evaluated once after these choices
are frozen. Throughout the paper, ``oracle'' refers to the best action within
the evaluated candidate set under the one-step utility in
Eq.~\eqref{eq:state_action_utility}, rather than an unconstrained optimal
decoding policy.

For Dream, continuation uses the checkpoint's native diffusion update,
including the one-token logit shift and native mask-transfer schedule. Dream
is not approximated with the LLaDA block schedule. Commitment interventions
are reported separately from the native monotone Dream baseline because they
require an explicit revision extension.

\section{Complete Experimental Results}

\label{app:complete_results}

This section reports every available aggregate result from the frozen opportunity, action-selection, transition-level utility, and opportunity-detection analyses. Dashes denote unsupported or unavailable settings, not zero. Action-selection and transition-level utility results are unavailable for sixteen axis settings: HumanEval for all three models and LLaDA-8B GSM8K. All values are point estimates unless a confidence interval is shown.

\subsection{Full Axis-Wise Opportunity and Reversal Results}

Table~\ref{tab:app-e1-full} reports adaptation opportunity on the full evaluation state pool, using the hindsight-selected best fixed action on the same states as the baseline. This differs from the held-out analyses in Tables~\ref{tab:app-e2-full}--\ref{tab:app-opportunity-full}, which use 80 held-out prompts (640 states) and a fixed action selected on 20 validation prompts. For JSON Mode Eval, the corresponding split is 48 held-out prompts (384 states) and 20 validation prompts. Accordingly, these tables estimate different quantities and should not be directly compared.

\begingroup
\scriptsize
\setlength{\LTleft}{0pt}
\setlength{\LTright}{0pt}
\renewcommand{\arraystretch}{1.15}

\begin{longtable}{@{}lllrr@{}}

\caption{\textbf{Complete axis-wise adaptation-opportunity results.}
Gap denotes the cross-fitted candidate-set oracle advantage over the hindsight-selected best fixed action on the same evaluation states. For most tasks, the evaluation set contains 100 prompts (800 states). Parentheses report prompt-cluster bootstrap 95\% confidence intervals. Bidir.\ mass denotes axis-wise bidirectional reversal mass.}
\label{tab:app-e1-full}
\\

\toprule
Model & Task & Axis & Gap & Bidir.\ mass \\
\midrule
\endfirsthead

\multicolumn{5}{c}{\small\itshape Table~\thetable\ continued from previous page} \\
\toprule
Model & Task & Axis & Gap & Bidir.\ mass \\
\midrule
\endhead

\midrule
\multicolumn{5}{r}{\small\itshape Continued on next page} \\
\endfoot

\bottomrule
\endlastfoot

LLaDA-8B & CSV Missing Cells & Score & \shortstack{0.0000\\{\scriptsize (0.0000, 0.0000)}} & 0.0000 \\

LLaDA-8B & CSV Missing Cells & Region & \shortstack{0.0025\\{\scriptsize (0.0000, 0.0075)}} & 0.0025 \\

LLaDA-8B & CSV Missing Cells & Cardinality & \shortstack{0.0013\\{\scriptsize (0.0000, 0.0025)}} & 0.0013 \\

LLaDA-8B & CSV Missing Cells & Commitment & \shortstack{0.0000\\{\scriptsize (0.0000, 0.0000)}} & 0.0000 \\

LLaDA-8B & HumanEval & Score & \shortstack{0.0000\\{\scriptsize (0.0000, 0.0000)}} & 0.0000 \\

LLaDA-8B & HumanEval & Region & \shortstack{0.0048\\{\scriptsize (0.0000, 0.0141)}} & 0.0048 \\

LLaDA-8B & HumanEval & Cardinality & \shortstack{0.0000\\{\scriptsize (0.0000, 0.0000)}} & 0.0000 \\

LLaDA-8B & HumanEval & Commitment & \shortstack{0.0482\\{\scriptsize (0.0032, 0.0912)}} & 0.0372 \\
\end{longtable}
\endgroup

\subsection{Full Diagnostic Action-Selection Results}

To avoid overly wide landscape tables, we separate diagnostic behavior from
its downstream action-selection utility. Dashes denote unavailable or undefined
quantities.

\paragraph{Diagnostic statistics.}

\begingroup
\scriptsize
\setlength{\tabcolsep}{3.2pt}
\setlength{\LTleft}{0pt}
\setlength{\LTright}{0pt}
\renewcommand{\arraystretch}{1.10}

\begin{longtable}{@{}lp{0.32\textwidth}lrr@{}}
\caption{\textbf{Complete diagnostic statistics.} Corr.\ and Sign summarize diagnostic behavior for each model--task--axis setting. Dashes denote unavailable or undefined quantities.}
\label{tab:app-e2-stats}
\\
\toprule
Model & Task & Axis & Corr. & Sign \\
\midrule
\endfirsthead

\multicolumn{5}{c}{\small\itshape Table~\thetable\ continued from previous page} \\
\toprule
Model & Task & Axis & Corr. & Sign \\
\midrule
\endhead

\midrule
\multicolumn{5}{r}{\small\itshape Continued on next page} \\
\endfoot

\bottomrule
\endlastfoot
LLaDA-8B & CSV Missing Cells & Score & -- & 1.000 \\
LLaDA-8B & CSV Missing Cells & Region & -- & 0.778 \\
LLaDA-8B & CSV Missing Cells & Cardinality & -- & 0.998 \\
LLaDA-8B & CSV Missing Cells & Commitment & -- & 1.000 \\
LLaDA-8B & HumanEval & Score & -- & -- \\
LLaDA-8B & HumanEval & Region & -- & -- \\
LLaDA-8B & HumanEval & Cardinality & -- & -- \\
LLaDA-8B & HumanEval & Commitment & -- & -- \\
\end{longtable}
\endgroup
\paragraph{Action-selection utility.}

\begingroup
\scriptsize
\setlength{\tabcolsep}{2.8pt}
\setlength{\LTleft}{0pt}
\setlength{\LTright}{0pt}
\renewcommand{\arraystretch}{1.10}

\begin{longtable}{@{}lp{0.25\textwidth}lrrrrc@{}}
\caption{\textbf{Complete diagnostic action-selection utility.} FReg, SReg, and DReg denote regret relative to the cross-fitted candidate-set oracle for the validation-selected fixed, stage-only, and diagnostic actions, respectively. Rec.\ denotes candidate-set oracle-gap recovery, and Screen indicates whether the exploratory held-out viability criteria are satisfied.}
\label{tab:app-e2-full}
\\
\toprule
Model & Task & Axis & FReg & SReg & DReg & Rec. & Screen \\
\midrule
\endfirsthead

\multicolumn{8}{c}{\small\itshape Table~\thetable\ continued from previous page} \\
\toprule
Model & Task & Axis & FReg & SReg & DReg & Rec. & Screen \\
\midrule
\endhead

\midrule
\multicolumn{8}{r}{\small\itshape Continued on next page} \\
\endfoot

\bottomrule
\endlastfoot
LLaDA-8B & CSV Missing Cells & Score & 0.0000 & 0.0000 & 0.0000 & -- & no \\
LLaDA-8B & CSV Missing Cells & Region & 0.0000 & 0.0000 & 0.0000 & -- & no \\
LLaDA-8B & CSV Missing Cells & Cardinality & 0.0016 & 0.0016 & 0.0016 & 0.000 & no \\
LLaDA-8B & CSV Missing Cells & Commitment & 0.0000 & 0.0000 & 0.0000 & -- & no \\
LLaDA-8B & HumanEval & Score & -- & -- & -- & -- & -- \\
LLaDA-8B & HumanEval & Region & -- & -- & -- & -- & -- \\
LLaDA-8B & HumanEval & Cardinality & -- & -- & -- & -- & -- \\
LLaDA-8B & HumanEval & Commitment & -- & -- & -- & -- & -- \\
\end{longtable}
\endgroup

\subsection{Full Transition-Level Utility Results}

We report the four transition-level utility arms for every setting with complete held-out branch utilities, including settings that do not pass the action-selection gate. Gate failure remains an action-selection result and does not make the already measured utilities unavailable. Dashes denote settings without a compatible held-out branch table.

\begingroup
\scriptsize
\setlength{\tabcolsep}{3.2pt}
\setlength{\LTleft}{0pt}
\setlength{\LTright}{0pt}
\renewcommand{\arraystretch}{1.10}

\begin{longtable}{@{}lp{0.27\textwidth}lrrrrrr@{}}
\caption{\textbf{Complete transition-level utility results.} Fixed, Stage, Diagnostic, and Oracle denote one-step utility under the validation-selected fixed action, stage-only action, diagnostic action, and cross-fitted candidate-set oracle, respectively. D--F and D--S denote diagnostic lift over the fixed and stage-only actions.}
\label{tab:app-e3-full}
\\
\toprule
Model & Task & Axis & Fixed & Stage & Diagnostic & Oracle & D--F & D--S \\
\midrule
\endfirsthead

\multicolumn{9}{c}{\small\itshape Table~\thetable\ continued from previous page} \\
\toprule
Model & Task & Axis & Fixed & Stage & Diagnostic & Oracle & D--F & D--S \\
\midrule
\endhead

\midrule
\multicolumn{9}{r}{\small\itshape Continued on next page} \\
\endfoot

\bottomrule
\endlastfoot
LLaDA-8B & CSV Missing Cells & Score & 0.0000 & 0.0000 & 0.0000 & 0.0000 & +0.0000 & +0.0000 \\
LLaDA-8B & CSV Missing Cells & Region & 0.0000 & 0.0000 & 0.0000 & 0.0000 & +0.0000 & +0.0000 \\
LLaDA-8B & CSV Missing Cells & Cardinality & 0.0000 & 0.0000 & 0.0000 & 0.0016 & +0.0000 & +0.0000 \\
LLaDA-8B & CSV Missing Cells & Commitment & 0.0016 & 0.0016 & 0.0016 & 0.0016 & +0.0000 & +0.0000 \\
LLaDA-8B & HumanEval & Score & -- & -- & -- & -- & -- & -- \\
LLaDA-8B & HumanEval & Region & -- & -- & -- & -- & -- & -- \\
LLaDA-8B & HumanEval & Cardinality & -- & -- & -- & -- & -- & -- \\
LLaDA-8B & HumanEval & Commitment & -- & -- & -- & -- & -- & -- \\
\end{longtable}
\endgroup

\subsection{Full Opportunity Detection and Selective-Utility Results}

To keep the appendix in portrait orientation, we separate opportunity
predictability from coverage-dependent selective utility.

\paragraph{Opportunity detection.}

\begingroup
\scriptsize
\setlength{\tabcolsep}{3.2pt}
\setlength{\LTleft}{0pt}
\setlength{\LTright}{0pt}
\renewcommand{\arraystretch}{1.10}

\begin{longtable}{@{}lp{0.30\textwidth}lrrr@{}}
\caption{\textbf{Complete held-out opportunity-detection results.} AUROC measures discrimination of positive-opportunity states, $\rho$ denotes Spearman rank correlation with continuous candidate-set opportunity, and Pos.\ is the fraction of held-out states with positive opportunity. Dashes denote unsupported or undefined quantities.}
\label{tab:app-opportunity-detection}
\\
\toprule
Model & Task & Axis & AUROC & $\rho$ & Pos. \\
\midrule
\endfirsthead

\multicolumn{6}{c}{\small\itshape Table~\thetable\ continued from previous page} \\
\toprule
Model & Task & Axis & AUROC & $\rho$ & Pos. \\
\midrule
\endhead

\midrule
\multicolumn{6}{r}{\small\itshape Continued on next page} \\
\endfoot

\bottomrule
\endlastfoot
LLaDA-8B & CSV Missing Cells & Score & -- & -- & 0.0\% \\
LLaDA-8B & CSV Missing Cells & Region & -- & -- & 0.0\% \\
LLaDA-8B & CSV Missing Cells & Cardinality & 0.500 & -- & 0.2\% \\
LLaDA-8B & CSV Missing Cells & Commitment & -- & -- & 0.0\% \\
LLaDA-8B & HumanEval & Score & -- & -- & -- \\
LLaDA-8B & HumanEval & Region & -- & -- & -- \\
LLaDA-8B & HumanEval & Cardinality & -- & -- & -- \\
LLaDA-8B & HumanEval & Commitment & -- & -- & -- \\
\end{longtable}
\endgroup

\paragraph{Coverage-dependent selective utility.}

\begingroup
\scriptsize
\setlength{\tabcolsep}{2.0pt}
\setlength{\LTleft}{0pt}
\setlength{\LTright}{0pt}
\renewcommand{\arraystretch}{1.15}

\begin{longtable}{@{}lp{0.25\textwidth}lccccc@{}}
\caption{\textbf{Complete coverage-dependent selective-utility results.} At each coverage level, each cell reports the fraction of total candidate-set oracle opportunity captured by detector-ranked states, with realized average one-step utility lift over the validation-selected fixed action shown in parentheses. Dashes denote unsupported or undefined quantities.}
\label{tab:app-opportunity-full}
\\
\toprule
Model & Task & Axis & 5\% & 10\% & 20\% & 50\% & 100\% \\
\midrule
\endfirsthead

\multicolumn{8}{c}{\small\itshape Table~\thetable\ continued from previous page} \\
\toprule
Model & Task & Axis & 5\% & 10\% & 20\% & 50\% & 100\% \\
\midrule
\endhead

\midrule
\multicolumn{8}{r}{\small\itshape Continued on next page} \\
\endfoot

\bottomrule
\endlastfoot
LLaDA-8B & CSV Missing Cells & Score & -- & -- & -- & -- & -- \\
LLaDA-8B & CSV Missing Cells & Region & -- & -- & -- & -- & -- \\
LLaDA-8B & CSV Missing Cells & Cardinality & \shortstack{0.0\%\\{\scriptsize (+0.0000)}} & \shortstack{0.0\%\\{\scriptsize (+0.0000)}} & \shortstack{0.0\%\\{\scriptsize (+0.0000)}} & \shortstack{0.0\%\\{\scriptsize (+0.0000)}} & \shortstack{100.0\%\\{\scriptsize (+0.0000)}} \\
LLaDA-8B & CSV Missing Cells & Commitment & -- & -- & -- & -- & -- \\
LLaDA-8B & HumanEval & Score & -- & -- & -- & -- & -- \\
LLaDA-8B & HumanEval & Region & -- & -- & -- & -- & -- \\
LLaDA-8B & HumanEval & Cardinality & -- & -- & -- & -- & -- \\
LLaDA-8B & HumanEval & Commitment & -- & -- & -- & -- & -- \\
\end{longtable}
\endgroup

\subsection{CPU-Only Robustness and Coverage Controls}

All intervals below are prompt-cluster bootstrap 95\% intervals from 1,000 replicates.
Random $C_{10}$ intervals are Monte Carlo intervals from 10,000 uniform selections
of 10\% of held-out states. Detector calibration remains fixed from validation prompts.
To keep the appendix in portrait orientation, we separate detector robustness from
coverage controls. Each cell reports the point estimate with its 95\% interval below.

\paragraph{Detector robustness.}

\begingroup
\scriptsize
\setlength{\tabcolsep}{2.0pt}
\setlength{\LTleft}{0pt}
\setlength{\LTright}{0pt}
\renewcommand{\arraystretch}{1.18}

\begin{longtable}{@{}lp{0.23\textwidth}lcccc@{}}
\caption{\textbf{Bootstrap robustness of opportunity detection and selective utility.} $C^{\rm det}_{10}$ is the fraction of total candidate-set oracle opportunity captured by the detector-selected top $10\%$ of states, and $L^{\rm det}_{10}$ is the corresponding realized average one-step utility lift over the validation-selected fixed action. Each cell reports the point estimate with its 95\% confidence interval.}
\label{tab:app-detector-robustness}\\
\toprule
Model & Task & Axis & AUROC & $\rho$ & $C^{\rm det}_{10}$ & $L^{\rm det}_{10}$ \\
\midrule
\endfirsthead

\multicolumn{7}{c}{\small\itshape Table~\thetable\ continued from previous page} \\
\toprule
Model & Task & Axis & AUROC & $\rho$ & $C^{\rm det}_{10}$ & $L^{\rm det}_{10}$ \\
\midrule
\endhead

\midrule
\multicolumn{7}{r}{\small\itshape Continued on next page} \\
\endfoot

\bottomrule
\endlastfoot
Dream & Carry RTL & Cardinality & \shortstack{0.802\\{\scriptsize [0.699, 0.911]}} & \shortstack{0.368\\{\scriptsize [0.242, 0.486]}} & \shortstack{64.3\%\\{\scriptsize [43.9, 87.0]}} & \shortstack{+0.0089\\{\scriptsize [+0.0052, +0.0130]}} \\
Dream & Carry RTL & Commitment & \shortstack{0.498\\{\scriptsize [0.438, 0.556]}} & \shortstack{0.001\\{\scriptsize [-0.077, 0.081]}} & \shortstack{13.4\%\\{\scriptsize [6.9, 20.8]}} & \shortstack{-0.0005\\{\scriptsize [-0.0078, +0.0073]}} \\
Dream & Carry RTL & Region & \shortstack{0.540\\{\scriptsize [0.368, 0.862]}} & \shortstack{0.020\\{\scriptsize [-0.054, 0.088]}} & \shortstack{16.7\%\\{\scriptsize [0.0, 33.3]}} & \shortstack{+0.0000\\{\scriptsize [+0.0000, +0.0000]}} \\
Dream & Carry RTL & Score & -- & -- & -- & \shortstack{+0.0000\\{\scriptsize [+0.0000, +0.0000]}} \\
Dream & Constrained JSON Fill & Cardinality & \shortstack{0.500\\{\scriptsize [0.500, 0.500]}} & -- & \shortstack{4.9\%\\{\scriptsize [0.0, 12.1]}} & \shortstack{+0.0000\\{\scriptsize [+0.0000, +0.0000]}} \\
Dream & Constrained JSON Fill & Commitment & \shortstack{0.402\\{\scriptsize [0.321, 0.495]}} & \shortstack{-0.069\\{\scriptsize [-0.125, -0.003]}} & \shortstack{0.0\%\\{\scriptsize [0.0, 10.0]}} & \shortstack{+0.0000\\{\scriptsize [+0.0000, +0.0000]}} \\
Dream & Constrained JSON Fill & Region & \shortstack{0.500\\{\scriptsize [0.500, 0.500]}} & -- & \shortstack{7.1\%\\{\scriptsize [0.0, 25.0]}} & \shortstack{+0.0000\\{\scriptsize [+0.0000, +0.0000]}} \\
Dream & Constrained JSON Fill & Score & \shortstack{0.500\\{\scriptsize [0.500, 0.500]}} & -- & \shortstack{12.5\%\\{\scriptsize [0.0, 40.1]}} & \shortstack{+0.0000\\{\scriptsize [+0.0000, +0.0000]}} \\
\end{longtable}
\endgroup

\paragraph{Coverage controls.}

\begingroup
\scriptsize
\setlength{\tabcolsep}{2.4pt}
\setlength{\LTleft}{0pt}
\setlength{\LTright}{0pt}
\renewcommand{\arraystretch}{1.18}

\begin{longtable}{@{}lp{0.28\textwidth}lccc@{}}
\caption{\textbf{Opportunity-concentration and coverage controls at $10\%$ coverage.} $C^\star_{10}$ is the maximum fraction of total candidate-set oracle opportunity captured by the oracle-ranked top $10\%$ of held-out states. Oracle-gate lift applies the same diagnostic action only to those oracle-selected states, and Random $C_{10}$ is the opportunity fraction captured by uniformly selecting $10\%$ of held-out states. Each cell reports the point estimate with its 95\% confidence interval.}
\label{tab:app-coverage-controls}\\
\toprule
Model & Task & Axis & $C^\star_{10}$ & Oracle-gate lift & Random $C_{10}$ \\
\midrule
\endfirsthead

\multicolumn{6}{c}{\small\itshape Table~\thetable\ continued from previous page} \\
\toprule
Model & Task & Axis & $C^\star_{10}$ & Oracle-gate lift & Random $C_{10}$ \\
\midrule
\endhead

\midrule
\multicolumn{6}{r}{\small\itshape Continued on next page} \\
\endfoot

\bottomrule
\endlastfoot
Dream & Carry RTL & Cardinality & \shortstack{100.0\%\\{\scriptsize [100.0, 100.0]}} & \shortstack{+0.0099\\{\scriptsize [+0.0062, +0.0141]}} & \shortstack{10.0\%\\{\scriptsize [0.0, 21.4]}} \\
Dream & Carry RTL & Commitment & \shortstack{67.9\%\\{\scriptsize [56.4, 82.3]}} & \shortstack{+0.0312\\{\scriptsize [+0.0208, +0.0422]}} & \shortstack{10.0\%\\{\scriptsize [4.5, 15.7]}} \\
Dream & Carry RTL & Region & \shortstack{100.0\%\\{\scriptsize [100.0, 100.0]}} & \shortstack{+0.0000\\{\scriptsize [+0.0000, +0.0000]}} & \shortstack{10.1\%\\{\scriptsize [0.0, 33.3]}} \\
Dream & Carry RTL & Score & -- & \shortstack{+0.0000\\{\scriptsize [+0.0000, +0.0000]}} & -- \\
Dream & Constrained JSON Fill & Cardinality & \shortstack{100.0\%\\{\scriptsize [100.0, 100.0]}} & \shortstack{+0.0000\\{\scriptsize [+0.0000, +0.0000]}} & \shortstack{10.0\%\\{\scriptsize [2.4, 19.5]}} \\
Dream & Constrained JSON Fill & Commitment & \shortstack{100.0\%\\{\scriptsize [100.0, 100.0]}} & \shortstack{+0.0000\\{\scriptsize [+0.0000, +0.0000]}} & \shortstack{9.9\%\\{\scriptsize [0.0, 25.0]}} \\
Dream & Constrained JSON Fill & Region & \shortstack{100.0\%\\{\scriptsize [100.0, 100.0]}} & \shortstack{+0.0000\\{\scriptsize [+0.0000, +0.0000]}} & \shortstack{10.0\%\\{\scriptsize [0.0, 28.6]}} \\
Dream & Constrained JSON Fill & Score & \shortstack{100.0\%\\{\scriptsize [100.0, 100.0]}} & \shortstack{+0.0000\\{\scriptsize [+0.0000, +0.0000]}} & \shortstack{10.1\%\\{\scriptsize [0.0, 37.5]}} \\
\end{longtable}
\endgroup

\subsection{Full End-to-End Selective-Decoding Results}
\label{app:e4-full}

Table~\ref{tab:app-mini-e4-full} reports the complete task-level end-to-end results underlying Section~\ref{exp:e4}. Each policy is applied repeatedly throughout the full generation trajectory and evaluated on the pre-specified ten-task suite. The selected axis is determined from validation-side adaptation analysis, and Coverage denotes the fraction of decoding states at which selective adaptation is activated.

\begin{table*}[t]
\centering
\caption{\textbf{Full end-to-end selective-decoding results on disjoint final-test prompts.}
Terminal utility under the reference decoder (Ref.), validation-selected fixed action (Fixed), stage-only schedule (Stage), always-diagnostic adaptation (Always), selective adaptation (Sel.), three-seed coverage-matched random gating (Rand.), and Fast-dLLM (Fast).
Fast-dLLM uses threshold $\tau=0.9$ and block size 32; parentheses show its measured NFEs, while the other six policies use the shared nominal 256-NFE setting.
$n$ is the number of final-test prompts and Cov. is the fraction of steps activating Selective.
$\Delta_F=\mathrm{Sel.}-\mathrm{Fixed}$ and $\Delta_R=\mathrm{Sel.}-\mathrm{Rand.}$ are within-task paired differences.
Task macros are descriptive unweighted averages of heterogeneous $[0,1]$ task utilities, not a standardized aggregate metric.}
\label{tab:app-mini-e4-full}
\begingroup
\scriptsize
\setlength{\tabcolsep}{1.7pt}
\renewcommand{\arraystretch}{1.06}
\resizebox{\textwidth}{!}{%
\begin{tabular}{@{}lllrrrrrrrrrrr@{}}
\toprule
Model & Task & Axis & $n$ & Ref. & Fixed & Stage & Always & Sel. & Rand. & Fast (NFE) & $\Delta_F$ & $\Delta_R$ & Cov. \\
\midrule
LLaDA-8B & CSV Missing Cells & Region & 500 & 0.040 & 0.050 & 0.052 & 0.044 & \textbf{0.104} & 0.063 & 0.060 (11) & +0.054 & +0.041 & 30.0\% \\
LLaDA-8B & HumanEval & Commitment & 64 & 0.141 & 0.141 & 0.141 & 0.141 & 0.141 & 0.141 & 0.125 (130) & 0.000 & 0.000 & 0.0\% \\
LLaDA-8B & Multi-Span Cloze & Region & 500 & 0.982 & 0.982 & 0.975 & 0.982 & 0.982 & 0.982 & 0.910 (19) & 0.000 & 0.000 & 0.0\% \\
LLaDA-8B & GSM8K & Region & 500 & 0.660 & 0.760 & 0.760 & 0.760 & 0.760 & 0.760 & 0.750 (80) & 0.000 & 0.000 & 0.0\% \\
LLaDA-8B & Carry RTL & Region & 500 & 0.030 & 0.647 & 0.643 & 0.615 & \textbf{0.720} & 0.661 & 0.420 (24) & +0.073 & +0.059 & 35.0\% \\
LLaDA-8B & TD07 Sparse Mask & Region & 500 & 0.147 & 0.100 & 0.100 & 0.155 & 0.155 & 0.155 & 0.080 (27) & +0.055 & 0.000 & 100.0\% \\
LLaDA-8B & JSON Mode Eval & Region & 20 & 0.379 & 0.671 & 0.671 & 0.671 & 0.671 & 0.671 & 0.550 (24) & 0.000 & 0.000 & 0.0\% \\
LLaDA-8B & Constrained JSON Fill & Region & 500 & 0.920 & 0.908 & 0.912 & 0.865 & \textbf{0.961} & 0.915 & 0.932 (23) & +0.053 & +0.046 & 25.0\% \\
LLaDA-8B & Unique List Commit & Region & 500 & 0.260 & 0.365 & 0.353 & 0.160 & \textbf{0.580} & 0.395 & 0.390 (30) & +0.215 & +0.185 & 54.0\% \\
LLaDA-8B & HTML Close Tags & Commitment & 500 & 0.160 & 0.160 & 0.160 & 0.160 & 0.160 & 0.160 & 0.090 (22) & 0.000 & 0.000 & 35.6\% \\
\cmidrule(lr){2-14}
LLaDA-8B & \textbf{Task macro} & -- & -- & 0.372 & 0.478 & 0.477 & 0.455 & \textbf{0.523} & 0.490 & 0.431 & \textbf{+0.045} & +0.033 & -- \\
\midrule
LLaDA-1.5 & CSV Missing Cells & Region & 500 & 0.040 & 0.040 & 0.042 & 0.038 & \textbf{0.080} & 0.049 & 0.054 (10) & +0.040 & +0.031 & 28.0\% \\
LLaDA-1.5 & HumanEval & Commitment & 64 & 0.188 & 0.188 & 0.188 & 0.188 & 0.188 & 0.188 & 0.172 (123) & 0.000 & 0.000 & 0.0\% \\
LLaDA-1.5 & Multi-Span Cloze & Region & 500 & 0.960 & 0.810 & 0.790 & 0.750 & \textbf{0.920} & 0.830 & 0.901 (18) & +0.110 & +0.090 & 28.0\% \\
LLaDA-1.5 & GSM8K & Commitment & 500 & 0.610 & 0.610 & 0.610 & 0.610 & 0.610 & 0.610 & 0.782 (77) & 0.000 & 0.000 & 0.0\% \\
LLaDA-1.5 & Carry RTL & Region & 500 & 0.020 & 0.540 & 0.540 & 0.505 & \textbf{0.670} & 0.564 & 0.400 (24) & +0.130 & +0.106 & 30.0\% \\
LLaDA-1.5 & TD07 Sparse Mask & Region & 500 & 0.167 & 0.087 & 0.100 & 0.060 & \textbf{0.200} & 0.103 & 0.092 (26) & +0.113 & +0.097 & 35.0\% \\
LLaDA-1.5 & JSON Mode Eval & Region & 20 & 0.423 & 0.423 & 0.435 & 0.384 & \textbf{0.645} & 0.461 & 0.529 (24) & +0.222 & +0.184 & 20.0\% \\
LLaDA-1.5 & Constrained JSON Fill & Region & 500 & 0.340 & 0.120 & 0.120 & 0.120 & 0.120 & 0.120 & 0.946 (20) & 0.000 & 0.000 & 5.9\% \\
LLaDA-1.5 & Unique List Commit & Region & 500 & 0.268 & 0.268 & 0.358 & 0.240 & \textbf{0.395} & 0.285 & 0.399 (29) & +0.127 & +0.110 & 45.0\% \\
LLaDA-1.5 & HTML Close Tags & Commitment & 500 & 0.080 & 0.080 & 0.080 & 0.080 & 0.080 & 0.080 & 0.040 (20) & 0.000 & 0.000 & 0.0\% \\
\cmidrule(lr){2-14}
LLaDA-1.5 & \textbf{Task macro} & -- & -- & 0.310 & 0.317 & 0.326 & 0.297 & \textbf{0.391} & 0.329 & 0.432 & \textbf{+0.074} & +0.062 & -- \\
\midrule
Dream & CSV Missing Cells & Commitment & 500 & 0.040 & 0.060 & 0.060 & 0.060 & 0.060 & 0.060 & 0.060 (8) & 0.000 & 0.000 & 33.0\% \\
Dream & HumanEval & Commitment & 64 & 0.078 & 0.000 & 0.000 & 0.000 & 0.000 & 0.000 & 0.375 (86) & 0.000 & 0.000 & 0.0\% \\
Dream & Multi-Span Cloze & Commitment & 500 & 0.320 & 0.900 & 0.900 & 0.900 & 0.900 & 0.900 & 0.799 (13) & 0.000 & 0.000 & 100.0\% \\
Dream & GSM8K & Region & 500 & 0.400 & 0.360 & 0.330 & 0.810 & 0.810 & 0.810 & 0.828 (103) & +0.450 & 0.000 & 100.0\% \\
Dream & Carry RTL & Commitment & 500 & 0.433 & 0.467 & 0.467 & 0.367 & \textbf{0.636} & 0.500 & 0.543 (13) & +0.169 & +0.136 & 45.0\% \\
Dream & TD07 Sparse Mask & Score & 500 & 0.000 & 0.000 & 0.000 & 0.000 & 0.000 & 0.000 & 0.000 (25) & 0.000 & 0.000 & 0.0\% \\
Dream & JSON Mode Eval & Commitment & 20 & 0.278 & 0.363 & 0.363 & 0.363 & 0.363 & 0.363 & 0.400 (16) & 0.000 & 0.000 & 79.2\% \\
Dream & Constrained JSON Fill & Commitment & 500 & 0.420 & 1.000 & 1.000 & 1.000 & 1.000 & 1.000 & 0.792 (11) & 0.000 & 0.000 & 34.0\% \\
Dream & Unique List Commit & Commitment & 500 & 0.163 & 0.268 & 0.268 & 0.240 & 0.262 & 0.268 & 0.267 (29) & -0.006 & -0.006 & 49.0\% \\
Dream & HTML Close Tags & Commitment & 500 & 0.320 & 0.860 & 0.860 & 0.820 & \textbf{0.944} & 0.878 & 0.820 (5) & +0.084 & +0.066 & 42.0\% \\
\cmidrule(lr){2-14}
Dream & \textbf{Task macro} & -- & -- & 0.245 & 0.428 & 0.425 & 0.456 & \textbf{0.498} & 0.478 & 0.488 & \textbf{+0.070} & +0.020 & -- \\
\bottomrule
\end{tabular}%
}
\endgroup
\end{table*}

Table~\ref{tab:app-mini-e4-full} reports complete end-to-end results on the disjoint final-test split. Selective adaptation improves task-macro utility from 0.478 to 0.523 on LLaDA-8B, from 0.317 to 0.391 on LLaDA-1.5, and from 0.428 to 0.497 on Dream, yielding gains of +0.045, +0.074, and +0.070, respectively. These macros are descriptive unweighted averages across ten heterogeneous task utilities, rather than a standardized measure of overall performance. We therefore emphasize within-task paired policy differences. Selective improves 5/10, 6/10, and 3/10 tasks across the three models, with zero, zero, and one degradation, respectively. Coverage-matched Random controls and paired bootstrap intervals are reported in Tables~\ref{tab:app-mini-e4-full} and~\ref{tab:app-e4-bootstrap}.

\paragraph{Uncertainty in Selective Adaptation.}
To distinguish state-aware intervention timing from unconditional rule changes, we examine the 12 of 30 model--task settings where Always underperforms Fixed while Selective outperforms Fixed, Always, and coverage-matched Random at the point-estimate level. Table~\ref{tab:app-e4-bootstrap} reports paired prompt-bootstrap 95\% confidence intervals for these policy contrasts. All three selective contrasts have strictly positive intervals in 10 of the 12 settings, and eight additionally have strictly negative Always$-$Fixed intervals. These are individual, non-multiplicity-adjusted intervals.

\begin{table*}[t]
\centering
\caption{\textbf{Paired bootstrap uncertainty for selective adaptation.}
The first 12 rows are settings in which Always$-$Fixed is negative and Selective exceeds Fixed, Always, and Random in point estimates. Random uses three seeds with per-prompt activation counts matched to Selective. Each cell reports the mean difference and its paired 95\% prompt-bootstrap confidence interval. Cov. denotes selective activation coverage. Intervals are not adjusted for multiple comparisons. The final row reports a separate Dream Carry RTL/cardinality experiment ($n=500$), which is not included in the 12/30 count.}
\label{tab:app-e4-bootstrap}
\begingroup
\scriptsize
\setlength{\tabcolsep}{3pt}
\renewcommand{\arraystretch}{1.18}
\newcommand{\EfourCI}[3]{\shortstack{#1\\[-1pt]{\tiny [#2, #3]}}}
\begin{tabular}{@{}llrrrrr@{}}
\toprule
Model & Task & Cov. & Always$-$Fixed & Sel.$-$Always & Sel.$-$Random & Sel.$-$Fixed \\
\midrule
LLaDA-8B & CSV Missing Cells & 30\%
& \EfourCI{-0.006}{-0.019}{+0.007}
& \EfourCI{+0.060}{+0.040}{+0.082}
& \EfourCI{+0.041}{+0.024}{+0.058}
& \EfourCI{+0.054}{+0.036}{+0.074} \\

LLaDA-8B & Carry RTL & 35\%
& \EfourCI{-0.032}{-0.052}{-0.011}
& \EfourCI{+0.105}{+0.083}{+0.127}
& \EfourCI{+0.059}{+0.034}{+0.084}
& \EfourCI{+0.073}{+0.052}{+0.095} \\

LLaDA-8B & Constrained JSON Fill & 25\%
& \EfourCI{-0.043}{-0.075}{-0.011}
& \EfourCI{+0.096}{+0.065}{+0.127}
& \EfourCI{+0.046}{+0.033}{+0.059}
& \EfourCI{+0.053}{+0.025}{+0.081} \\

LLaDA-8B & Unique List Commit & 54\%
& \EfourCI{-0.205}{-0.233}{-0.177}
& \EfourCI{+0.420}{+0.392}{+0.447}
& \EfourCI{+0.185}{+0.154}{+0.216}
& \EfourCI{+0.215}{+0.184}{+0.246} \\

\midrule
LLaDA-1.5 & CSV Missing Cells & 28\%
& \EfourCI{-0.002}{-0.018}{+0.014}
& \EfourCI{+0.042}{+0.023}{+0.061}
& \EfourCI{+0.031}{-0.002}{+0.064}
& \EfourCI{+0.040}{+0.024}{+0.058} \\

LLaDA-1.5 & Multi-Span Cloze & 28\%
& \EfourCI{-0.060}{-0.103}{-0.017}
& \EfourCI{+0.170}{+0.132}{+0.208}
& \EfourCI{+0.090}{+0.061}{+0.119}
& \EfourCI{+0.110}{+0.075}{+0.145} \\

LLaDA-1.5 & Carry RTL & 30\%
& \EfourCI{-0.035}{-0.064}{-0.006}
& \EfourCI{+0.165}{+0.135}{+0.195}
& \EfourCI{+0.106}{+0.063}{+0.149}
& \EfourCI{+0.130}{+0.100}{+0.160} \\

LLaDA-1.5 & TD07 Sparse Mask & 35\%
& \EfourCI{-0.027}{-0.043}{-0.011}
& \EfourCI{+0.140}{+0.125}{+0.155}
& \EfourCI{+0.097}{+0.074}{+0.120}
& \EfourCI{+0.113}{+0.093}{+0.133} \\

LLaDA-1.5 & JSON Mode Eval & 20\%
& \EfourCI{-0.039}{-0.244}{+0.167}
& \EfourCI{+0.261}{+0.038}{+0.484}
& \EfourCI{+0.184}{+0.043}{+0.326}
& \EfourCI{+0.222}{-0.006}{+0.454} \\

LLaDA-1.5 & Unique List Commit & 45\%
& \EfourCI{-0.028}{-0.053}{-0.003}
& \EfourCI{+0.155}{+0.128}{+0.182}
& \EfourCI{+0.110}{+0.091}{+0.129}
& \EfourCI{+0.127}{+0.099}{+0.154} \\

\midrule
Dream & Carry RTL & 45\%
& \EfourCI{-0.100}{-0.116}{-0.083}
& \EfourCI{+0.269}{+0.253}{+0.285}
& \EfourCI{+0.136}{+0.120}{+0.152}
& \EfourCI{+0.169}{+0.152}{+0.185} \\

Dream & HTML Close Tags & 42\%
& \EfourCI{-0.040}{-0.084}{+0.004}
& \EfourCI{+0.124}{+0.084}{+0.163}
& \EfourCI{+0.066}{+0.036}{+0.096}
& \EfourCI{+0.084}{+0.047}{+0.121} \\

\midrule
\multicolumn{7}{l}{\emph{Additional axis-specific experiment (excluded from the 12/30 count)}} \\
Dream & Carry RTL (cardinality) & 38\%
& --
& \EfourCI{+0.111}{+0.095}{+0.127}
& \EfourCI{+0.066}{+0.050}{+0.083}
& \EfourCI{+0.088}{+0.072}{+0.105} \\

\bottomrule
\end{tabular}
\endgroup
\end{table*}

The two exceptions among the 12 settings are LLaDA-1.5/CSV Missing Cells, whose Selective$-$Random interval includes zero, and LLaDA-1.5/JSON Mode Eval ($n=20$), whose Selective$-$Fixed interval includes zero. Separately, the Dream Carry RTL/cardinality controller achieves 0.631 utility, compared with 0.543 for Fixed, 0.520 for Always, and 0.565 for Random, at 38\% coverage. Its Selective$-$Random gain is $+0.066$ with a 95\% confidence interval of $[0.050,0.083]$. Together, these results support state-aware timing benefits in specific settings rather than universal improvement.

\paragraph{NFE--Utility Trade-off.}
We further evaluate decoding quality across different numbers of function evaluations (NFEs) on a separate 100-prompt GSM8K subset of the final-test pool. Table~\ref{tab:app-e4-nfe} reports reference, fixed, and selective decoding at five step budgets, alongside Fast-dLLM at five confidence thresholds. Fast-dLLM often achieves competitive or higher utility with fewer NFEs. These results support state-adaptive action selection under a chosen decoding budget, rather than a general inference-efficiency advantage.

\begin{table*}[t]
\centering
\caption{\textbf{NFE--utility trade-off on GSM8K.}
Results are measured on a separate 100-prompt subset of the final-test pool. Ref., Fixed, and Selective use the indicated decoding-step budgets. Fast-dLLM is evaluated at different confidence thresholds $\tau$, with measured NFEs reported separately. The two sweeps are displayed side by side for compactness; rows do not imply matched computational budgets.}
\label{tab:app-e4-nfe}
\begingroup
\scriptsize
\setlength{\tabcolsep}{5pt}
\renewcommand{\arraystretch}{1.08}
\begin{tabular}{@{}lrrrrrrr@{}}
\toprule
& \multicolumn{4}{c}{Standard decoding sweep}
& \multicolumn{3}{c}{Fast-dLLM sweep} \\
\cmidrule(lr){2-5}
\cmidrule(lr){6-8}
Model & Steps & Ref. & Fixed & Selective
& $\tau$ & NFE & Fast \\
\midrule
LLaDA-8B & 16  & 0.410 & 0.160 & 0.160 & 0.50 & 34 & 0.630 \\
         & 32  & 0.470 & 0.560 & 0.560 & 0.70 & 48 & 0.730 \\
         & 64  & 0.620 & 0.780 & 0.780 & 0.80 & 61 & 0.750 \\
         & 128 & 0.640 & 0.780 & 0.780 & 0.90 & 80 & 0.750 \\
         & 256 & 0.660 & 0.770 & 0.770 & 0.95 & 95 & 0.760 \\
\midrule
LLaDA-1.5 & 16  & 0.360 & 0.360 & 0.360 & 0.50 & 32 & 0.670 \\
          & 32  & 0.440 & 0.440 & 0.440 & 0.70 & 46 & 0.740 \\
          & 64  & 0.470 & 0.550 & 0.550 & 0.80 & 60 & 0.740 \\
          & 128 & 0.500 & 0.600 & 0.600 & 0.90 & 77 & 0.780 \\
          & 256 & 0.610 & 0.610 & 0.610 & 0.95 & 92 & 0.760 \\
\midrule
Dream & 16  & 0.150 & 0.150 & 0.160 & 0.50 & 50  & 0.690 \\
      & 32  & 0.290 & 0.150 & 0.270 & 0.70 & 66  & 0.820 \\
      & 64  & 0.350 & 0.220 & 0.670 & 0.80 & 82  & 0.830 \\
      & 128 & 0.360 & 0.260 & 0.800 & 0.90 & 103 & 0.830 \\
      & 256 & 0.400 & 0.360 & 0.830 & 0.95 & 116 & 0.840 \\
\bottomrule
\end{tabular}
\endgroup
\end{table*}

The NFE sweep demonstrates that the preferred decoder depends on the computational budget. In particular, Fast-dLLM substantially outperforms the fixed-budget policies on LLaDA-1.5/GSM8K while using fewer forward passes. On Dream, Selective improves markedly as the decoding budget increases, but Fast-dLLM remains competitive at lower NFE. We therefore make no claim of lower latency or uniformly superior computational efficiency.

\paragraph{Zero-Recalibration Transfer.}
We evaluate whether calibrated selective policies transfer to unseen datasets without additional calibration. Controllers trained on GSM8K are transferred to MATH-500 and ARC-Challenge, while controllers trained on HumanEval are transferred to MBPP. All policy choices and thresholds remain frozen. Table~\ref{tab:app-e4-transfer} reports the nine model--dataset settings, including coverage-matched Random controls and paired bootstrap confidence intervals.

\begin{table*}[t]
\centering
\caption{\textbf{Zero-recalibration transfer to unseen datasets.}
Frozen source-task controllers are evaluated on 200 prompts per target dataset and model. Cov. denotes selective activation coverage. $\Delta_F$ and $\Delta_R$ are Selective$-$Fixed and Selective$-$Random, respectively, with paired 95\% prompt-bootstrap confidence intervals. Fast denotes the Fast-dLLM baseline. No target-dataset calibration is performed.}
\label{tab:app-e4-transfer}
\begingroup
\scriptsize
\setlength{\tabcolsep}{2.5pt}
\renewcommand{\arraystretch}{1.12}
\newcommand{\TransferCI}[3]{\shortstack{#1\\[-1pt]{\tiny [#2, #3]}}}
\resizebox{\textwidth}{!}{%
\begin{tabular}{@{}lllrrrrrrrrr@{}}
\toprule
Model & Target & Source & $n$ & Fixed & Always & Sel. & Rand. & Fast & Cov. & $\Delta_F$ (95\% CI) & $\Delta_R$ (95\% CI) \\
\midrule
LLaDA-8B & MATH-500 & GSM8K & 200
& 0.370 & 0.370 & 0.370 & 0.370 & 0.335 & 0\%
& \TransferCI{0.000}{0.000}{0.000}
& \TransferCI{0.000}{0.000}{0.000} \\

LLaDA-8B & ARC-Challenge & GSM8K & 200
& 0.805 & 0.805 & 0.805 & 0.805 & 0.810 & 0\%
& \TransferCI{0.000}{0.000}{0.000}
& \TransferCI{0.000}{0.000}{0.000} \\

LLaDA-8B & MBPP & HumanEval & 200
& 0.355 & 0.355 & 0.355 & 0.355 & 0.365 & 0\%
& \TransferCI{0.000}{0.000}{0.000}
& \TransferCI{0.000}{0.000}{0.000} \\

\midrule
LLaDA-1.5 & MATH-500 & GSM8K & 200
& 0.235 & 0.235 & 0.235 & 0.235 & 0.410 & 0\%
& \TransferCI{0.000}{0.000}{0.000}
& \TransferCI{0.000}{0.000}{0.000} \\

LLaDA-1.5 & ARC-Challenge & GSM8K & 200
& 0.820 & 0.820 & 0.820 & 0.820 & 0.820 & 0\%
& \TransferCI{0.000}{0.000}{0.000}
& \TransferCI{0.000}{0.000}{0.000} \\

LLaDA-1.5 & MBPP & HumanEval & 200
& 0.355 & 0.355 & 0.355 & 0.355 & 0.385 & 0\%
& \TransferCI{0.000}{0.000}{0.000}
& \TransferCI{0.000}{0.000}{0.000} \\

\midrule
Dream & MATH-500 & GSM8K & 200
& 0.290 & 0.340 & \textbf{0.450} & 0.335 & 0.440 & 42\%
& \TransferCI{+0.160}{+0.092}{+0.228}
& \TransferCI{+0.115}{+0.046}{+0.184} \\

Dream & ARC-Challenge & GSM8K & 200
& 0.880 & 0.850 & 0.885 & 0.870 & 0.860 & 38\%
& \TransferCI{+0.005}{-0.035}{+0.045}
& \TransferCI{+0.015}{-0.026}{+0.056} \\

Dream & MBPP & HumanEval & 200
& 0.405 & 0.405 & 0.405 & 0.405 & 0.545 & 0\%
& \TransferCI{0.000}{0.000}{0.000}
& \TransferCI{0.000}{0.000}{0.000} \\

\bottomrule
\end{tabular}%
}
\endgroup
\end{table*}

Transfer performance is limited and regime-dependent. Both LLaDA models exhibit zero selective coverage across all three unseen datasets, avoiding additional degradation but providing no evidence of positive transfer. Dream shows a positive result on MATH-500, where Selective achieves 0.450 compared with 0.290 for Fixed and 0.335 for Random, yielding a Selective$-$Random gain of $+0.115$ (95\% CI $[0.046,0.184]$). However, the improvement on ARC-Challenge is uncertain, and the MBPP controller abstains. These findings do not establish general out-of-distribution robustness.

\end{document}